\documentclass{article} 
\usepackage{colm2024_conference}

\usepackage{booktabs}
\usepackage{graphicx}
\usepackage{enumitem}
\usepackage{wrapfig}
\usepackage{algorithm}
\usepackage{algpseudocode}
\usepackage{natbib}
\usepackage{makecell}
\usepackage{booktabs}
\usepackage{bbm}
\usepackage{array}
\usepackage{amsmath} 
\usepackage{amssymb}
\usepackage{amsfonts}
\usepackage{multirow}
\usepackage{verbatim}
\usepackage{caption}
\usepackage{longtable}
\usepackage{supertabular}
\usepackage{hyperref}
\usepackage{CJKutf8}
\usepackage[utf8]{inputenc} 
\usepackage[T1]{fontenc} 
\usepackage[french,vietnamese,mongolian,greek,english]{babel}
\usepackage{pifont}
\usepackage{afterpage}
\usepackage{enumitem}
\usepackage{tablefootnote}
\usepackage{colortbl}
\definecolor{Gray}{gray}{0.9}
\definecolor{QwenPurple}{HTML}{4F46E5}
\usepackage{xspace}
\usepackage{textcomp}
\usepackage{makecell}
\usepackage{lscape} 
\usepackage{siunitx}
\usepackage{listings}
\usepackage{xcolor}
\usepackage{adjustbox}
\definecolor{dt}{gray}{0.7}
\definecolor{tongyi-purple}{RGB}{97,92,237}
\colorlet{tongyi-purple-alpha}{tongyi-purple!38}

\newcommand{\qwenvl}{{Qwen3-VL}\xspace}

\usepackage{pifont}       
\usepackage{bbding}       
\usepackage{fontawesome}

\usepackage{scrextend}

\usepackage{tgpagella}
\usepackage{latexsym}
\usepackage[T1]{fontenc}
\usepackage[utf8]{inputenc}
\usepackage{microtype}
\definecolor{mydarkblue}{rgb}{0,0.08,0.45}
\definecolor{citecolor}{HTML}{0071BC}
\usepackage{url}            
\usepackage{nicefrac}       
\usepackage{changepage}
\usepackage{xargs}          
\usepackage{wrapfig,lipsum,booktabs}
\usepackage{longtable}
\usepackage{subcaption}
\usepackage{endnotes}

\usepackage{pgfplots}
\usetikzlibrary{pgfplots.groupplots}
\pgfplotsset{compat=1.3}
\usepackage{tikz}
\usetikzlibrary{patterns}

\usepackage[most]{tcolorbox}
\usepackage{fvextra}
\usepackage{graphicx}
\usepackage[capitalize,noabbrev]{cleveref}
\crefname{section}{Section}{\S\S}
\Crefname{section}{Section}{\S\S}
\crefname{table}{Table}{Tables}
\crefname{figure}{Figure}{Figures}
\crefname{algorithm}{Algorithm}{}
\crefname{equation}{eq.}{}
\crefname{appendix}{Appendix}{}
\crefformat{section}{Section #2#1#3}
\usepackage{multicol}
\usepackage{fancyvrb} 
\usepackage{tcolorbox}
\newsavebox{\myverbcontent}
\usepackage{titlesec}
\titleformat*{\section}{\large\bfseries}

\usepackage{nicematrix} 
\usepackage{arydshln}

\makeatletter
\DeclareRobustCommand\onedot{\futurelet\@let@token\@onedot}
\def\@onedot{\ifx\@let@token.\else.\null\fi\xspace}

\def\etc{\emph{etc}\onedot} 

\def\eg{\emph{e.g}\onedot} 

\title{Qwen-RobotNav Technical Report: A Scalable Navigation Model Designed for an Agentic Navigation System}

\author{
\bf Qwen Team}

\newcommand{\qwennav}{{Qwen-RobotNav}\xspace}

\begin{document}

\maketitle

\begin{abstract} 
Agentic navigation systems require a base navigation model whose observation strategy can be externally reconfigured at inference time, because instruction following, object search, target tracking, and autonomous driving share the same perception-planning backbone yet demand fundamentally different strategies for consuming the visual stream.
We present \qwennav, a scalable navigation model built on \qwenvl that addresses it through a parameterised interface with two complementary dimensions: multiple task modes that select the navigation behaviour, and controllable observation parameters (e.g., token budget, per-camera weights) that govern how visual history is encoded.
With training-time randomization over all parameters, \qwennav is robust to any inference-time configuration requiring zero architectural modification to the \qwenvl backbone.
We train \qwennav on 15.6M samples;
co-training with vision-language data 
prevents the collapse into reactive action-sequence mappers observed in trajectory-only training.
The parameterised interface also makes \qwennav a natural building block for agentic systems: for long-horizon scenarios, an upper-level planner decomposes goals into sub-tasks and dynamically switches \qwennav's task mode and context strategy mid-episode, composing complex behaviours from repeated calls to the same model.
Extensive experiments show that \qwennav sets new state-of-the-art results across major navigation benchmarks, achieving 76.5\% success rate on VLN-CE RxR, 90.0\% tracking rate on EVT-Bench, and 91.4 PDMS on NAVSIM.
Beyond these standalone results, an agentic navigation system built with \textbf{\qwennav set a new state of the art on Embodied Question Answering, improving over the best prior method by 10.8\% on HM-EQA and 15.4\% on EXPRESS-Bench while requiring 77\% fewer navigation steps}.
The model exhibits favourable scaling from 2B to 8B parameters, with joint multi-task training developing a shared spatial-planning substrate that transfers across task families, and demonstrates strong zero-shot generalisation to real-world robots across diverse environments.

\end{abstract}
\begin{figure*}[ht]
\centering
\includegraphics[width=1\linewidth]{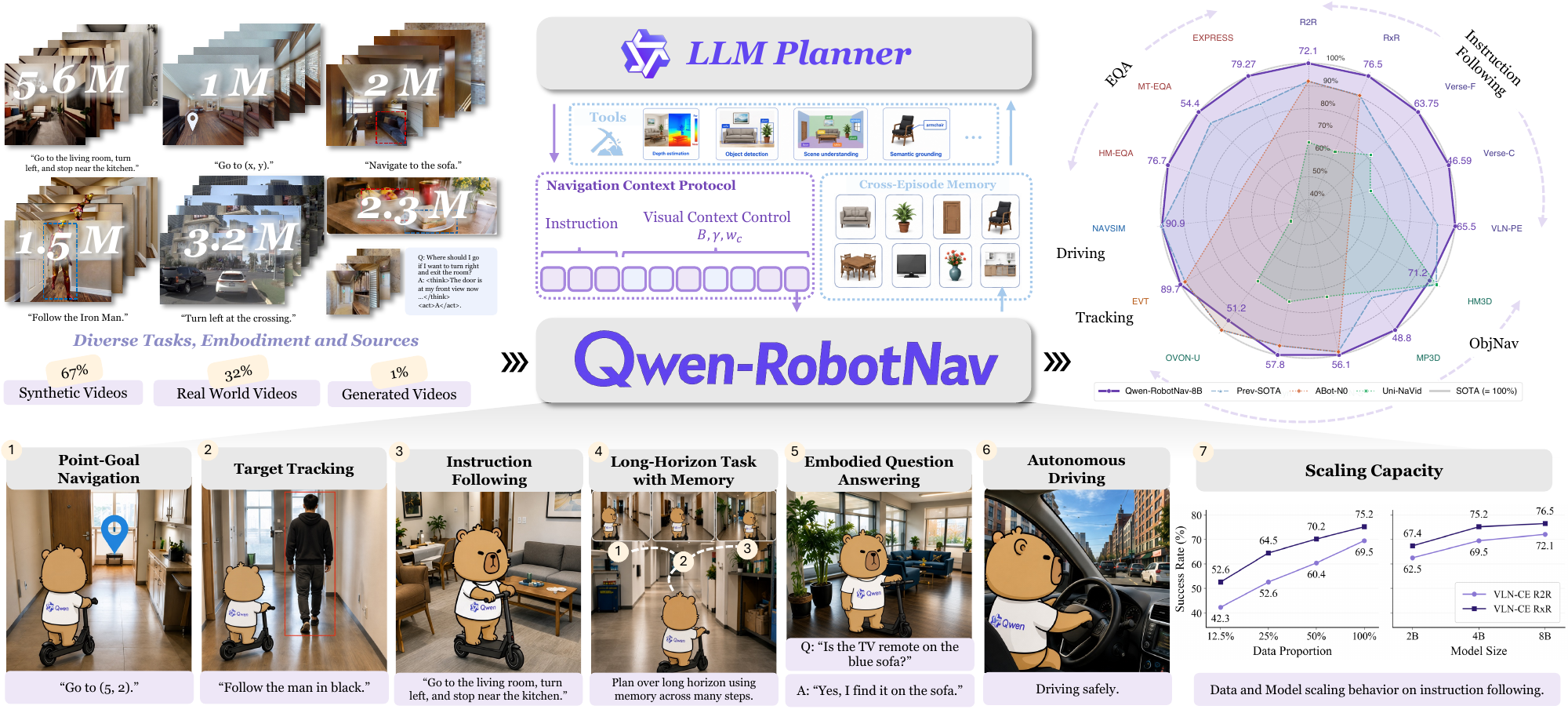}
\end{figure*}

\begin{figure*}[ht]
\centering
\includegraphics[width= 0.9\linewidth]{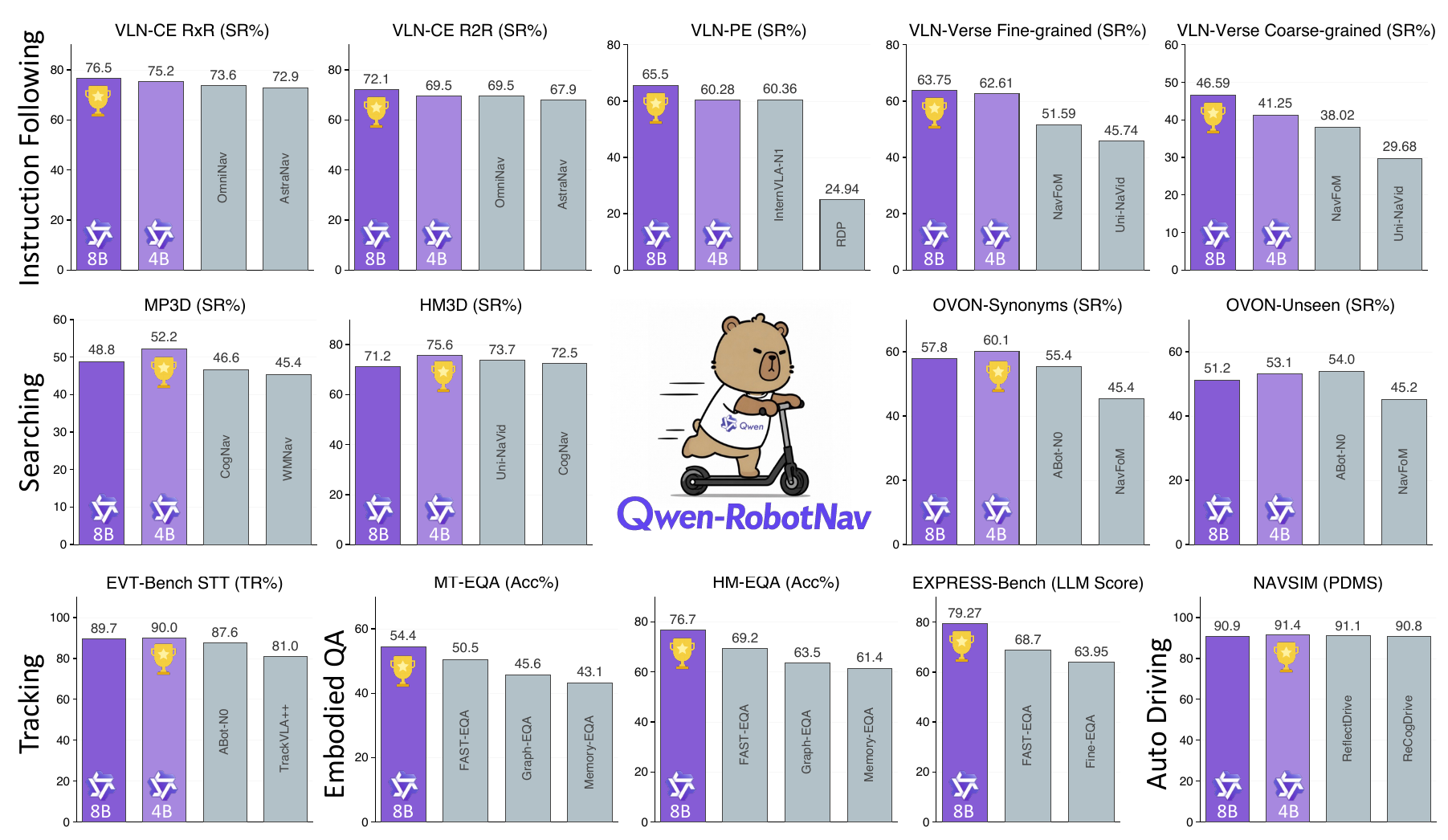}
\caption{\textbf{Benchmark summary.} Across instruction following, object search, target tracking, embodied question answering, and autonomous driving, \qwennav-4B and \qwennav-8B achieve state-of-the-art or competitive performance against specialist and navigation foundation model baselines. Trophy icons mark the best result in each benchmark group.}
\end{figure*}

\newpage

\section{Introduction}

Embodied navigation spans a remarkably diverse family of tasks, including instruction following~\citep{anderson2018vision, krantz2020beyond, anderson2020rxr, yin2025gcvln}, point-goal navigation~\citep{savva2019habitat, wijmans2020ddppo}, object searching~\citep{batra2020objectnav, yokoyama2024hm3d, yokoyama2024vlfm, yin2025unigoal}, target tracking~\citep{zhong2024empowering, wang2025trackvla, liu2025trackvlapp}, and autonomous driving~\citep{caesar2020nuscenes, sun2020scalability}, each coupling perception with purposeful movement through open-world environments.
Recent unified navigation models~\citep{zhang2024navid, zhang2024uni, zhang2025embodied, abotN0, cheng2024navila, li2026gn0, internvlan1, zhou2025same, zhong2025robotrom, yin2026alldaynav} and navigation-oriented evaluation suites~\citep{vlnverse, zhao2025vlnmme} have demonstrated that a single architecture can handle multiple task families.

However, complex real-world scenarios, such as Embodied Question Answering~\citep{das2018embodied, openeqa, exploreeqa, 3dmem, fasteqa}, demand more: the navigation model must serve as a core module within larger agentic systems, where an outer planner dynamically orchestrates navigation capabilities to accomplish long-horizon, multi-step goals.
This requires not only multi-task capability, but also a controllable interface whose observation strategy can be reconfigured by an external agent at inference time. Currently, no existing model meets this requirement: NavFoM~\citep{zhang2025embodied} uniformly sub-samples frames; ABot-N0~\citep{abotN0} retains only a sliding window, each embedding a single assumption about which observations matter that cannot be adjusted at deployment time.
The root cause is that different navigation tasks demand fundamentally different temporal context and memory strategies.
Instruction following requires retaining observations spanning dozens of prior steps to re-reference distant landmarks~\citep{an2024etpnav, wang2024lookahead, wei2025streamvln, wei2025ground}, whereas target tracking is governed by the most recent few frames and treats stale history as noise~\citep{zhong2024empowering, wang2025trackvla, liu2025trackvlapp}.
Object searching further complicates the picture by shifting its own requirements within a single episode, from broad global memory during exploration to tight recency-focused attention during approach~\citep{yokoyama2024vlfm, yin2024sgnav, kuang2024openfmnav, cao2024cognav}.
No single fixed context strategy can serve this full spectrum, let alone adapt mid-episode.

We introduce \qwennav, a scalable navigation model built on \qwenvl~\citep{qwen3} that reframes the central challenge of multi-task navigation as observation context modelling rather than architecture design.
\qwennav formulates all tasks as unified waypoint trajectory prediction and exposes a parameterised interface with two complementary configuration dimensions.
First, multiple task modes (VLN, PointNav, ObjNav, Tracking) allow an upper-level planner to select the navigation behaviour appropriate for each sub-goal.
Second, controllable observation parameters, including visual token budget, temporal decay, and per-camera importance weighting, govern how the model consumes the observation stream, enabling dynamic adjustment of the context strategy at inference time without task-specific retraining.
Training-time randomisation over all parameters ensures that \qwennav generalises to any inference-time configuration without task-specific tuning: the model never trains at a fixed setting, so it cannot overfit to one regime.
Camera identity, temporal order, and embodiment type are communicated entirely through natural-language tags and prompt preambles, requiring zero architectural modification to the \qwenvl backbone; supporting a new platform requires only a new prompt template, not new parameters.

The parameterised interface makes \qwennav a natural building block for agentic navigation systems.
For long-horizon scenarios such as Embodied Question Answering, where prior systems rely on explicit reasoning, exploration, and memory modules~\citep{zhou2024navgpt, zhou2024navgpt2, long2024instructnav, exploreeqa, grapheqa, 3dmem, fasteqa}, we deploy \qwennav within a two-tier hierarchical system.
An upper-tier planner (Qwen3.6-Plus) decomposes complex goals into sub-goals and dispatches configurable navigation calls, each specifying a task mode, a sub-goal instruction, and an observation configuration; \qwennav serves as the reactive executor, predicting waypoint trajectories at high frequency.
The planner can dynamically reconfigure \qwennav's task mode and context strategy mid-episode, and the two tiers communicate exclusively through natural language, keeping the system modular and extensible.
To support long-horizon reasoning, the system maintains a two-level memory: compact single-episode memory summarising each navigation rollout, and a persistent cross-episode memory that accumulates durable conclusions such as searched regions, candidate object locations, and rejected hypotheses, enabling effective context compression over extended episodes.

To train \qwennav, we curate \textbf{15.6M} samples comprising navigation trajectory planning data (85\%) spanning five task families (instruction following, point-goal navigation, object searching, target tracking, and autonomous driving) across diverse embodiments, and navigation-related vision-language reasoning data (15\%).
Co-training with vision-language data preserves the language understanding that underpins \qwennav's natural-language camera and temporal tags and mitigates the degradation of open-world perception that can occur when vision-language models are adapted only to action prediction~\citep{zhou2024navgpt2, zhang2024navid, zhang2024uni, cheng2024navila}.

Without any task-specific fine-tuning, \qwennav achieves state-of-the-art results across diverse navigation benchmarks.
On VLN-CE, \qwennav-8B attains \textbf{72.1\%} SR on R2R and \textbf{76.5\%} SR on RxR Val-Unseen, surpassing NavFoM by 10.4\% and 12.1\% SR respectively.
On EVT-Bench, \qwennav achieves the highest tracking rate (\textbf{90.0\%} TR) among all evaluated methods.
Strong cross-embodiment generalisation is also observed on HM3Dv2 object-goal navigation (\textbf{75.6\%} SR) and NAVSIM autonomous driving (\textbf{91.4} PDMS).
The model exhibits favourable scaling properties: performance improves from 2B to 8B parameters, with particularly pronounced gains on long-horizon reasoning tasks.
Equipped with the agentic system, \qwennav sets new state-of-the-art results on three EQA benchmarks (HM-EQA, MT-EQA, and EXPRESS-Bench), and further demonstrates zero-shot transfer to real-world robots across diverse environments.

\section{Navigation Model}
\label{sec:navigation_model}

\begin{figure}[t]
    \centering
    \includegraphics[width=\linewidth]{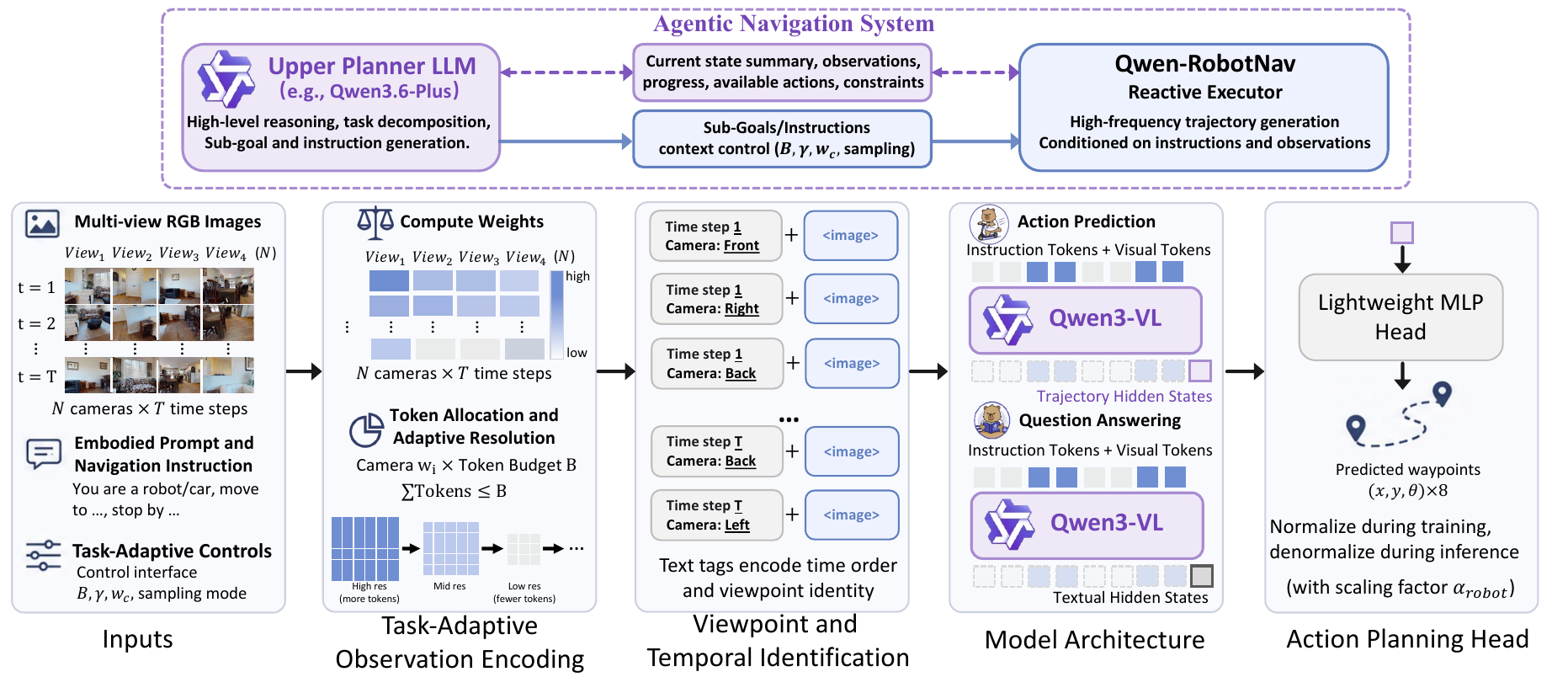}
    \caption{\textbf{\qwennav architecture.} \textit{Top:} In the agentic navigation system, an upper planner LLM decomposes long-horizon goals into sub-goals and controls \qwennav through task-adaptive context parameters such as token budget $B$, temporal decay $\gamma$, camera weights $w_c$, and frame sampling mode. \textit{Bottom:} \qwennav receives multi-view RGB observations, an embodied prompt, and a navigation instruction; allocates visual tokens across cameras and timesteps; inserts natural-language temporal and viewpoint tags; and feeds the resulting visual-language sequence into the \qwenvl backbone. A lightweight MLP action head maps trajectory hidden states to eight waypoint predictions $(x,y,\theta)$.}
    \label{fig:model_arch}
\end{figure}

\textbf{Task Formulation.}
We consider a general mobile navigation setting~\citep{zhang2025embodied,abotN0,zhou2025same} in which an embodied agent receives a textual instruction $\mathcal{L}$ and a sequence of multi-view observations $\mathbf{I}_{1:T}^{1:N} \in \mathbb{R}^{H \times W \times 3}$ captured from $N$ cameras at $T$ timesteps.
Given these inputs, the navigation policy $\pi$ must predict a waypoint trajectory
\begin{equation}
    \mathcal{W} = \{(x_k,\, y_k,\, \theta_k)\}_{k=1}^{K},
    \label{eq:task}
\end{equation}
where $K{=}8$ waypoints each encode a 2D position $(x_k, y_k)$ and heading $\theta_k$.
A central challenge is that $T$ grows online as navigation proceeds, $N$ varies across robot platforms, and the total visual token count scales as $\mathcal{O}(T \cdot N)$, rapidly exceeding the context budget of any LLM backbone without a principled compression strategy.
Moreover, different tasks impose fundamentally different demands on how observations should be modelled: target tracking requires only a short recency window of high-resolution frames to maintain lock on a moving object, whereas object-goal navigation must retain long-horizon episode history to recall previously explored regions and avoid redundant revisits.
A single, task-agnostic observation encoding therefore cannot serve all navigation tasks well, motivating the task-adaptive strategy described in Section~\ref{sec:token_alloc}.
 
\subsection{Model Architecture}
\label{sec:model_arch}

\qwennav inherits the \qwenvl architecture~\citep{qwen3} and augments it with a lightweight action head for trajectory regression.
The overall system consists of three components:

\begin{itemize}[leftmargin=1.5em]
    \item \textbf{Vision Encoder.} The vision encoder, inherited from \qwenvl, is built on a SigLIP-2~\citep{tschannen2025siglip} Vision Transformer with native dynamic-resolution support via 2D-RoPE~\citep{chen2025comp}, enabling the token allocation strategy (Section~\ref{sec:token_alloc}) to freely rescale each camera view to its allocated pixel budget.
    A two-layer MLP patch merger compresses $s_m^2$ adjacent spatial tokens into a single vector aligned to the LLM hidden dimension, and the DeepStack mechanism~\citep{meng2024deepstack} injects visual tokens from multiple ViT layers into early LLM layers, preserving multi-level visual representations critical for spatial reasoning.

    \item \textbf{Language Backbone.} The LLM, also inherited from \qwenvl, processes the concatenation of visual tokens (organised by the strategy described in Sections~\ref{sec:token_alloc} and \ref{sec:tvi}) and language tokens from $\mathcal{L}$.
    Through large-scale vision-language pretraining, the backbone has already acquired rich world knowledge, spatial understanding, and cross-modal reasoning capabilities that transfer directly to navigation.

    \item \textbf{Action Head.} A lightweight 4-layer MLP maps the LLM's final hidden state to $K{=}8$ waypoints, each with 3 degrees of freedom $(x_k, y_k, \theta_k)$, yielding a 24-dimensional output (Section~\ref{sec:action_planning}).
    By keeping the action head minimal, the bulk of spatial reasoning remains within the LLM, allowing the model to benefit from the pretrained language backbone's world knowledge when planning trajectories in unseen environments.
\end{itemize}

\begin{algorithm}[t]
    \caption{Task-Adaptive Observation Encoding}
    \label{alg:token_alloc}
    \begin{algorithmic}[1]
    \Require Observations $\mathbf{I}_{1:T}^{1:N}$, instruction $\mathcal{L}$, config $\Phi{=}(B,\, \gamma,\, \{w_c\},\, m,\, b_{\min},\, b_{\max})$
    \Ensure Token sequence fed to the LLM backbone
    \State $\mathbf{I}' \gets \Call{SampleFrames}{\mathbf{I}_{1:T}^{1:N},\; m,\; T'}$ \Comment{$T'{\leq}T$ frames}
    \State $\omega_t \gets \exp\!\bigl(\gamma \cdot t\,/\,(T'{-}1)\bigr),\; t{=}0,\ldots,T'{-}1$
    \State $\mathbf{W}[t,c] \gets \omega_t \cdot w_c$ for all $(t,c)$ \Comment{Eq.~\ref{eq:weight_matrix}}
    \State $\mathbf{A} \gets \Call{ConstrainedAlloc}{\mathbf{W},\; B,\; b_{\min},\; b_{\max}}$
    \For{each frame $(t, c)$}
        \State $\hat{\mathbf{I}}'_{t,c} \gets \Call{Resize}{\mathbf{I}'_{t,c},\;\; \mathbf{A}[t,c] \times (p \cdot s_m)^2}$
    \EndFor
    \State $\mathbf{V} \gets \Call{VisionEncode}{\hat{\mathbf{I}}'}$ \Comment{ViT + patch merger}
    \State $\mathbf{S} \gets \Call{Interleave}{\mathbf{V},\; \text{viewpoint/temporal tags}}$ \Comment{Sec.~\ref{sec:tvi}}
    \State \Return $\mathrm{Concat}(\mathbf{S},\; \mathrm{Tokenize}(\mathcal{L}))$
    \Statex
    \Function{ConstrainedAlloc}{$\mathbf{W},\, B,\, b_{\min},\, b_{\max}$}
        \State $\mathbf{A} \gets b_{\min} \cdot \mathbf{1}_{T' \times N}$;\quad $\mathcal{U} \gets$ all cells
        \Repeat
            \State Distribute $B {-} \sum\!\mathbf{A}$ to $\mathcal{U}$ proportionally to $\mathbf{W}$
            \State Clamp cells exceeding $b_{\max}$; remove from $\mathcal{U}$
        \Until{no cell exceeds $b_{\max}$}
        \State \Return $\mathbf{A}$
    \EndFunction
    \end{algorithmic}
    \end{algorithm}

\subsection{Task-Adaptive Observation Encoding}
\label{sec:token_alloc}

Navigation is inherently a sequential decision-making process under partial observability: the agent cannot see the full environment at once and must accumulate spatial knowledge from a stream of egocentric observations.
How much history to retain, and at what fidelity, depends fundamentally on the task's decision structure.
Tasks that require \emph{plan verification}, such as instruction following, demand global episode memory so the agent can re-reference previously observed landmarks against the instruction; tasks driven by \emph{reactive pursuit}, such as target tracking, depend almost entirely on the most recent frames to maintain a tight perception loop around the moving target.
Fixed context strategies, such as uniform sampling~\citep{zhang2024navid,zhang2024uni,zhang2025embodied} or sliding windows~\citep{abotN0}, commit to a single regime and cannot adapt across tasks without retraining.
\qwennav instead provides a unified, parameterised interface with four orthogonal control axes:

\begin{center}
\begin{tabular}{llll}
\toprule
\textbf{Parameter} & \textbf{Symbol} & \textbf{Range} & \textbf{Effect} \\
\midrule
Visual token budget & $B$ &  2048--4096 & Total tokens across all cameras and timesteps \\
Temporal decay & $\gamma$ &  $[1,3]$ & Recency bias toward the latest frames \\
Camera weights & $w_c$ & per-robot defaults & Per-camera importance multiplier \\
Frame sample mode & {-} & \texttt{random} / \texttt{latest} & History coverage vs.\ recency window \\
\bottomrule
\end{tabular}
\end{center}

\begin{figure*}[t]
    \centering
    \includegraphics[width=\linewidth]{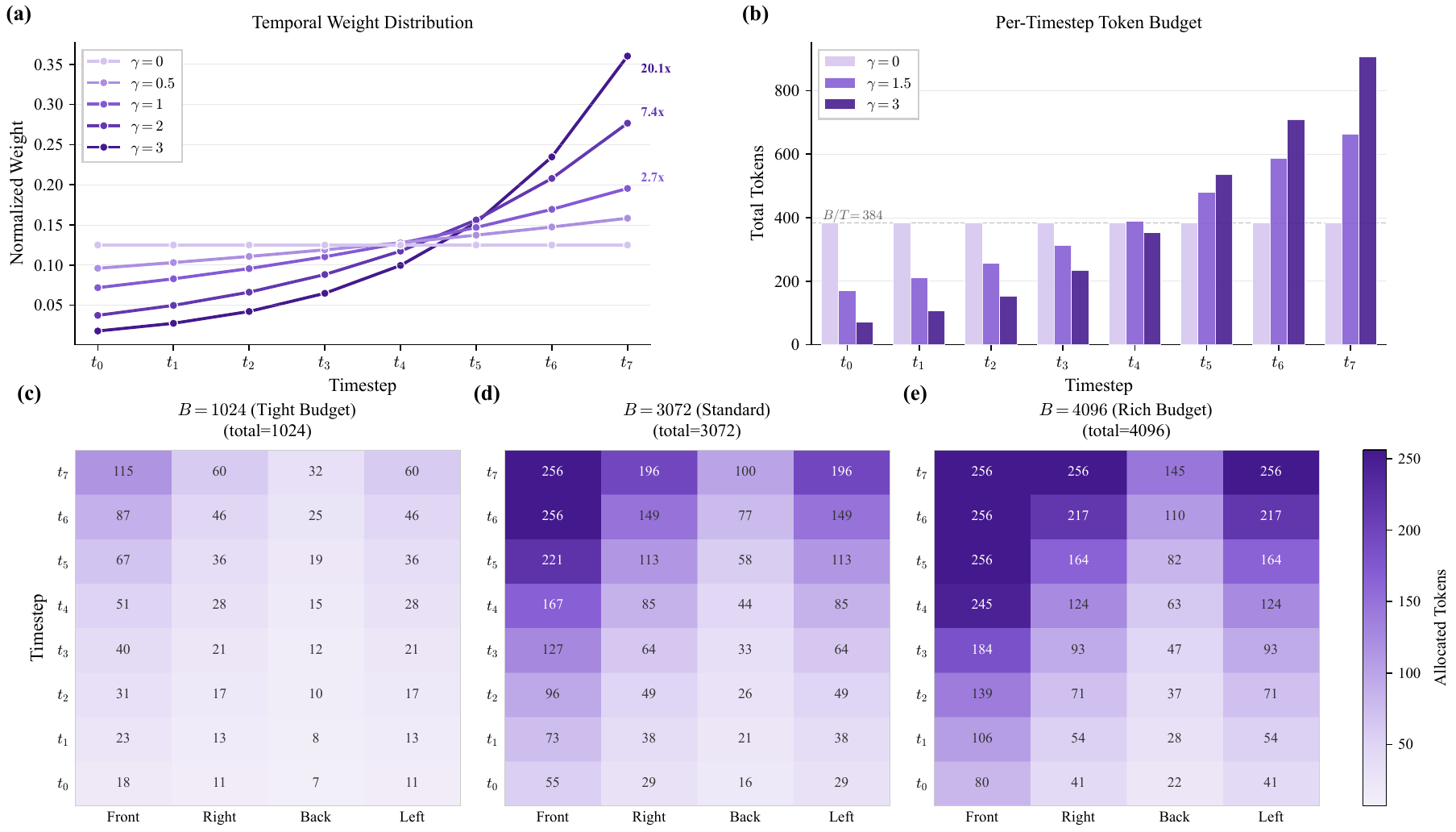}
    \caption{%
    \textbf{Visualization of task-adaptive observation encoding.}
    \textbf{(a)}~Normalized temporal weights $\omega_t = \exp(\gamma \cdot t / (T'{-}1))$ for varying decay factors $\gamma$ when $T'{>}1$; annotations show the newest-to-oldest weight ratio.
    \textbf{(b)}~Resulting per-timestep token budget (summed across all cameras) under a fixed total budget $B{=}3072$ with camera weights $w_c{=}[2.0, 1.0, 0.5, 1.0]$ for front, right, back, and left views. The dashed line marks the uniform baseline $B/T'$.
    \textbf{(c, d, e)}~Token allocation matrices across timesteps and cameras for three budget levels ($B{=}1024$, $3072$, $4096$) at fixed $\gamma{=}2.0$, with per-image constraints $b_{\min}{=}4$ and $b_{\max}{=}256$.
    Higher $\gamma$ concentrates tokens on the most recent frames, while larger $B$ increases overall fidelity until the per-image ceiling $b_{\max}$ saturates the highest-weight cells.
    }
    \label{fig:token_alloc_vis}
\end{figure*}

\textbf{Token Allocation Algorithm.}
Given $T$ timesteps and $N$ cameras, we first sub-sample or select $T'{\leq}T$ frames according to the frame sample mode, and then compute a temporal weight for each retained frame:
\begin{equation}
    \omega_t =
    \begin{cases}
    1, & T'=1,\\
    \exp\!\left(\gamma \cdot \frac{t}{T' - 1}\right), & T'>1,
    \end{cases}
    \quad t = 0, \ldots, T'{-}1,
    \label{eq:temporal_weight}
\end{equation}
so that $\gamma{=}0$ recovers uniform weighting and larger $\gamma$ allocates disproportionately more tokens to recent frames (at $\gamma{=}2$ the most recent frame receives $\approx 7.4\times$ the budget of the oldest), as visualised in \Cref{fig:token_alloc_vis}(a).
A joint weight matrix is then formed as:
\begin{equation}
    \mathbf{W}[t, c] = \omega_t \cdot w_c, \quad t \in \{0,\ldots,T'{-}1\},\; c \in \{1,\ldots,N\},
    \label{eq:weight_matrix}
\end{equation}
where $w_c$ is a per-camera importance weight that reflects the asymmetric information density across viewpoints: the forward-facing camera, which captures the richest actionable cues such as obstacles, navigable paths, and goal landmarks, is assigned the highest weight, while the rear camera receives the lowest as it primarily provides contextual redundancy (e.g. $w_c{=}[2.0, 1.0, 0.5, 1.0]$ for front, right, back, left). During training, $w_c$ is randomly sampled from its own range (e.g.\ the front camera from $\mathcal{U}[1.5, 2.5]$); see Section~\ref{sec:training} for full randomisation details.

For a given token budget $B$ and the frame $(t, c)$ at time step $t$ and camera $c$, token allocation proceeds in three stages: (1) each cell $(t, c)$ receives a minimum floor of $b_{\min}$ tokens; (2) the remaining budget $B - T' N b_{\min}$ is distributed proportionally to $\mathbf{W}$; (3) any cell exceeding the per-image ceiling $b_{\max}$ releases its surplus, which is redistributed iteratively until stable.
We require $T'N b_{\min} \le B \le T'N b_{\max}$; if a sampled configuration violates this constraint, $B$ is clipped to the feasible interval before allocation.
The allocated token count for each image then determines its pixel resolution through the patch size and spatial merge size of the vision encoder, and the image is rescaled to this pixel budget while preserving the original aspect ratio. Note that this token allocation is an empirically proven heuristic to expose a unified interface to control the token context of the model, which is also used in our agentic system (Section~\ref{sec:agentic}). We believe this strategy could be further improved by a more principled token allocation algorithm.
The full procedure is summarised in \Cref{alg:token_alloc}.
\Cref{fig:token_alloc_vis}(c, d, e) illustrates how the constrained allocation distributes tokens across timesteps and cameras under three representative budget levels.

\textbf{Training-Time Randomisation.}
At training time, all observation hyperparameters are independently randomised per sample: $\gamma \sim \mathcal{U}[1, 3]$, $B \sim \mathcal{U}[2048, 4096]$, each camera weight $w_c$ sampled from its own range (e.g.\ the front camera from $\mathcal{U}[1.5, 2.5]$), and the per-image token floor and ceiling $b_{\min} \sim \mathcal{U}_{\mathbb{Z}}[1, 8]$, $b_{\max} \sim \mathcal{U}_{\mathbb{Z}}[128, 256]$.
This ensures the model is never trained at a fixed configuration and supports robust generalisation across configurations within the randomised training range, with moderate extrapolation beyond it, without task-specific tuning.

\subsection{Viewpoint and Temporal Identification}
\label{sec:tvi}

After encoding, visual tokens from different cameras and timesteps are indistinguishable to the LLM: the backbone has no built-in mechanism to associate a token with ``front camera, step 12'' versus ``left camera, step 3''.
Without explicit identity signals, the model cannot learn camera-specific or temporally-ordered representations, limiting its ability to reason about spatial layout or episode history.

\qwennav resolves this by interleaving natural-language viewpoint and timestep tags with the visual tokens before they enter the LLM.
Concretely, each timestep group is introduced by a \texttt{Time step $t$} header, followed by each camera's name and its corresponding \texttt{<image>} token.
For example, a two-step, six-camera input is serialised as:

\begin{quote}
\small\ttfamily
Time step 0 Front View \textlangle image\textrangle\space
Front Right View \textlangle image\textrangle\space \ldots\space
Front Left View \textlangle image\textrangle\space
Time step 1 Front View \textlangle image\textrangle\space \ldots
\end{quote}

Camera identity and temporal order are thus communicated entirely through ordinary vocabulary tokens already present in \qwenvl, requiring zero architectural modification.
The model's existing open-world language grounding handles viewpoint semantics naturally: words like ``Front'', ``Left'', and ``Back Right'' carry spatial meaning that the pretrained LLM has already internalised from its pretraining corpus.
An alternative representation is to specify each camera by its numeric azimuth (e.g.\ ``right 90 degrees''); however, we find that descriptive names yield slightly better performance, likely because they carry richer semantic associations that the language model can leverage for spatial reasoning. 

\subsection{Embodiment-Aware Prompt Design}
\label{sec:embodiment_prompt}

Since \qwennav operates across fundamentally different physical platforms, it must distinguish between embodiments such as indoor mobile robots and autonomous vehicles.
Rather than encoding embodiment information through learned embeddings or separate model heads, \qwennav communicates the agent's physical identity through a natural-language preamble in the system prompt.
For example, an indoor robot sample begins with ``\texttt{Imagine you are a \textbf{robot} programmed for \textbf{navigation tasks}}'', whereas an autonomous driving sample begins with ``\texttt{Imagine you are a \textbf{car} programmed for \textbf{autonomous driving}}''.

This embodiment preamble acts as a \emph{task prior}: by identifying itself as a ``robot'' or a ``car'', the model recruits different subsets of its pretrained world knowledge (e.g., indoor spatial layouts versus traffic rules and road geometry) to inform trajectory prediction.
The text-based approach is also inherently extensible: supporting a new platform, such as a drone, a wheeled robot, or a quadruped robot, requires only defining a new prompt template, with no architectural changes or additional parameters.

\subsection{Action Planning}
\label{sec:action_planning}

The action planning module maps the LLM's final hidden state $\mathbf{E}^{\mathcal{A}} \in \mathbb{R}^{d}$ to a waypoint trajectory.
We use a lightweight 4-layer MLP head with hidden dimension 512 and GELU activations.
The 24-dimensional output encodes $K{=}8$ waypoints each with 3 DOF $(x_k, y_k, \theta_k)$.
Waypoints are normalised to $[-1,1]$ during training using per-dataset scale factors computed as the 99th percentile of each coordinate across all training trajectories, and training minimises the MSE loss between predicted and ground-truth trajectories.
At inference, the predicted waypoints are de-normalised using the same scale factors. 

\subsection{Training Strategy}
\label{sec:training}

\textbf{Training Objective.}
\qwennav is trained with a composite loss that jointly optimises trajectory regression and vision-language alignment:
\begin{equation}
    \mathcal{L} = \mathcal{L}_{\mathrm{traj}} + \lambda\,\mathcal{L}_{\mathrm{VL}},
    \label{eq:total_loss}
\end{equation}
where $\mathcal{L}_{\mathrm{traj}} = \|\hat{\mathcal{W}} - \mathcal{W}^*\|_2^2$ is the MSE loss between predicted and ground-truth waypoints (active only on navigation trajectory samples), and $\mathcal{L}_{\mathrm{VL}}$ is the standard next-token prediction loss on navigation-related vision-language reasoning samples.
The two objectives share the same forward pass; $\lambda$ is set to 1.0 in all experiments.

\textbf{Configuration Randomisation.}
A central principle of \qwennav's training is that \emph{no observation configuration is fixed}.
At each training step, all continuous control parameters of the task-adaptive interface ($B$, $\gamma$, $w_c$, $b_{\min}$, $b_{\max}$; ranges detailed in Section~\ref{sec:token_alloc}) are independently sampled per sample, and the frame sample mode alternates between \texttt{random} and \texttt{latest} with equal probability.


By exposing the model to a broad combinatorial space of these parameters during training, \qwennav generalises across configurations within the randomised training range without task-specific tuning.
In particular, training with both \texttt{random} (global coverage) and \texttt{latest} (recency window) frame sampling in equal proportion supports zero-shot context-strategy switching at deployment time, a property exploited by the agentic system described in Section~\ref{sec:agentic}.

\textbf{Co-training.}
The training corpus mixes 85\% navigation trajectory planning data with 15\% navigation-related vision-language reasoning data (Section~\ref{sec:data}).
Datasets are sampled at the batch level using per-dataset rates defined in a dataset registry, ensuring balanced exposure across all navigation task types (instruction following, point-goal navigation, object searching, target tracking, and autonomous driving) within every training epoch.
The vision-language component prevents catastrophic forgetting of \qwenvl's open-world perception capabilities: models trained on navigation trajectory data alone tend to collapse toward reactive action sequences and lose general-purpose spatial reasoning, while co-training preserves the rich language grounding that underpins \qwennav's zero-shot generalisation to unseen environments and instruction styles.

\textbf{Optimisation Details.}
\qwennav is initialised from the pretrained \qwenvl checkpoint and fine-tuned end-to-end.
We use the AdamW optimiser~\citep{loshchilov2019decoupled} with $\beta_1{=}0.9$, $\beta_2{=}0.95$, and weight decay $10^{-2}$.
A cosine learning rate schedule is applied with a linear warm-up over the first 3\% of training steps; the peak learning rate is $2{\times}10^{-5}$ for the vision encoder and LLM backbone and $1{\times}10^{-4}$ for the action head.
Gradient norms are clipped to 1.0.
The 8B model is trained with a global batch size of 256 for a total of 2{,}816 H100 GPU hours.

\section{\qwennav for Agentic Navigation}
\label{sec:agentic}

\subsection{Overview}
\label{sec:agent_overview}

The navigation model described in \Cref{sec:navigation_model} is designed not only for standalone benchmark evaluation, but also for deployment as a navigation module inside general-purpose embodied agents~\citep{zhou2024navgpt, long2024instructnav}.
In this setting, the upper-level agent should be able to decide not only \emph{what} navigation sub-goal to execute, but also \emph{which navigation task mode}, \emph{which observation configuration}, and \emph{when to query auxiliary visual evidence} should be used for the current phase of the task.
For example, a long-horizon search episode may alternate between instruction following, object-goal search, point-to-point movement, and target tracking, while also changing how much visual history should be retained.

We therefore expose \qwennav as an agent-ready navigation model through a lightweight tool interface.
At each navigation step, the upper-level planner provides a sub-goal instruction $\mathcal{L}_i$, selects a task mode $\tau_i$, and specifies an observation configuration $\Phi_i$.
The task mode $\tau_i$ describes the navigation behavior to be invoked, such as \texttt{VLN}, \texttt{PointNav}, \texttt{ObjNav}, or \texttt{Tracking}.
The configuration $\Phi_i=(B,\gamma,\{w_c\},m,b_{\min},b_{\max})$ follows the notation in \Cref{sec:token_alloc}: $B$ controls the total visual-token budget, $\gamma$ controls the recency bias, $\{w_c\}$ specifies camera weights, $m$ specifies the frame sample mode, and $b_{\min}, b_{\max}$ define the per-image allocation constraints.
Given the selected task mode and configuration, \qwennav predicts a waypoint trajectory $\mathcal{W}_i$ in the same form as \Cref{eq:task}.

Besides navigation calls, the planner may issue auxiliary vision-tool calls over the current observations or stored key frames.
In our interface, these tools include object detection, scene understanding, and semantic grounding.
They provide additional evidence for deciding the next sub-goal or verifying candidate observations, but they do not predict waypoints and do not replace \qwennav as the core navigation executor.

The resulting system uses \qwennav as the core navigation executor.
The upper-level planner is responsible for decomposing long-horizon goals, selecting task modes and context configurations, querying auxiliary visual evidence when needed, and reasoning over the evidence returned by previous navigation calls.
A lightweight harness connects the two sides: it converts planner decisions into \qwennav calls, and converts completed navigation rollouts into compact trajectory evidence for subsequent planning.
This keeps the agentic layer simple while allowing it to access the controllable interfaces already built into \qwennav.

\begin{figure*}[t]
    \centering
    \includegraphics[width=\linewidth]{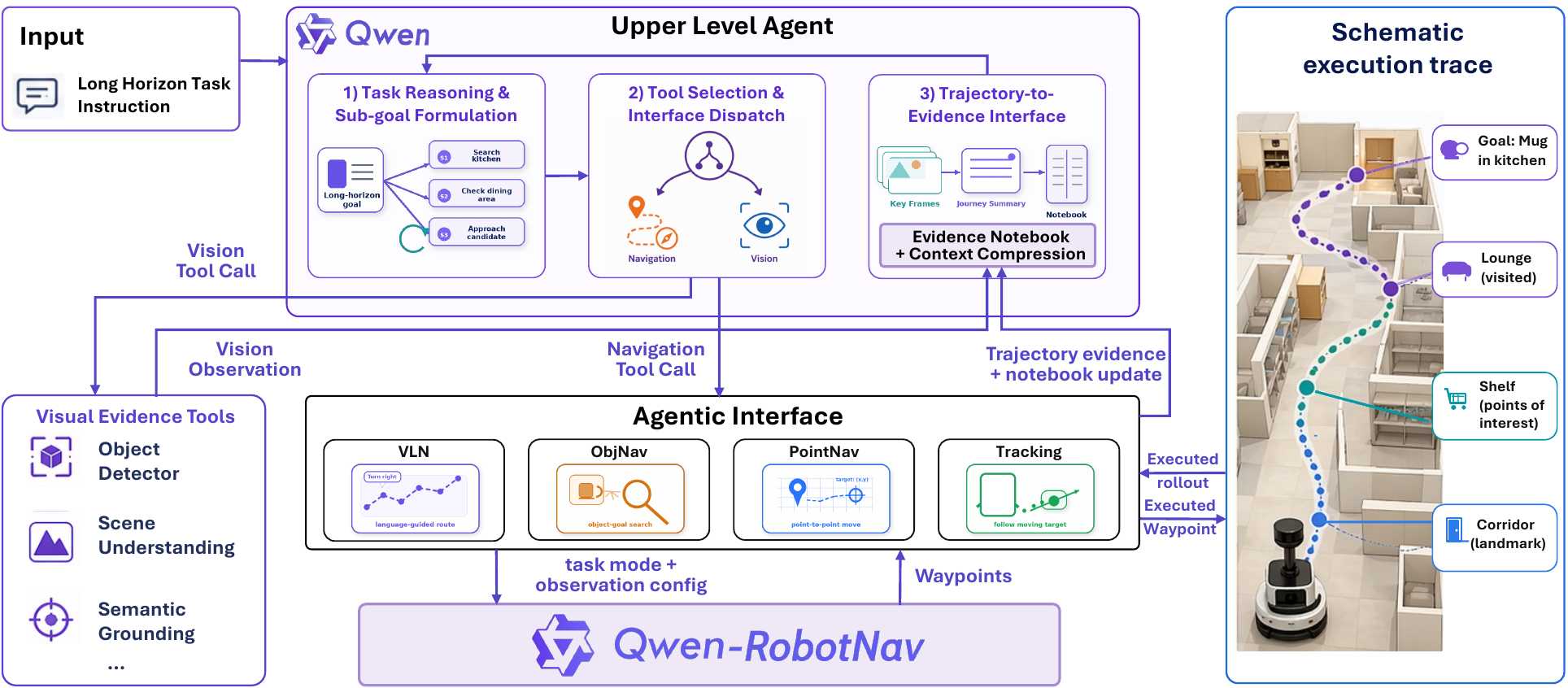}
    \caption{\textbf{\qwennav for agentic navigation.}
    An upper-level planner decomposes a long-horizon task into sub-goals and dispatches either auxiliary vision-tool calls or \qwennav navigation calls.
    Each navigation call is parameterized by a sub-goal instruction $\mathcal{L}_i$, a task mode $\tau_i$, and an observation configuration $\Phi_i$.
    \qwennav uses the selected task mode and configuration to predict waypoints $\mathcal{W}_i$, which are executed in the environment.
    The harness converts the resulting rollout into source-indexed key frames, trajectory summaries, and evidence-notebook updates, providing compact context for the next planning step.
    }
    \label{fig:agent_overview}
\end{figure*}

\subsection{Agent-Facing \qwennav Interface}
\label{sec:agent_interface}

The key agent-facing property of \qwennav is that its navigation behavior is externally configurable at inference time.
This configurability comes from two complementary navigation interfaces.

First, \qwennav supports multiple navigation task modes.
In \texttt{VLN} mode, the model follows natural-language route instructions and grounds them in visual observations.
In \texttt{PointNav} mode, it moves toward a specified spatial target or waypoint-like goal.
In \texttt{ObjNav} mode, it searches for an object category or instance using accumulated visual evidence.
In \texttt{Tracking} mode, it prioritizes recent observations to maintain lock on a moving or recently observed target.
These modes are not separate navigation policies; they are different task interfaces to the same \qwennav model.

Second, each navigation call can specify an observation configuration $\Phi$.
The model section has already described the full task-adaptive observation encoding algorithm in \Cref{sec:token_alloc}, so here we only emphasize how it is used by the planner.
The planner can increase $B$ and use a weaker recency bias when the current sub-goal needs broader episode history, such as object search or revisiting a previously explored region.
It can instead use a smaller $B$, larger $\gamma$, and \texttt{latest} frame sampling when the sub-goal is local and reactive, such as approaching a visible object or tracking a moving target.
In practice, $w_c$, $b_{\min}$, and $b_{\max}$ are usually kept at platform-level defaults, while $B$, $\gamma$, and $m$ are the main controls exposed to the planner.

A navigation call can therefore be written abstractly as:
\begin{equation}
    \mathcal{W}_i
    =
    \texttt{nav\_qwennav}
    \bigl(
        \mathcal{L}_i,\,
        \tau_i,\,
        \Phi_i
    \bigr),
    \label{eq:agent_qwennav_call}
\end{equation}
where $\mathcal{L}_i$ is the current sub-goal, $\tau_i$ is the selected task mode, and $\Phi_i$ is the observation configuration.
This interface gives the planner access to \qwennav's internal context controls without requiring architectural changes, task-specific fine-tuning, or a separate routing model.
The same model weights can be used across different task phases; only the tool-call arguments change.

For example, when searching for an object in a large indoor scene, the planner may call \qwennav in \texttt{ObjNav} mode with a larger token budget and history-covering frame sampling.
After a candidate object is observed, the planner can switch to \texttt{Tracking} or local \texttt{PointNav} behavior with a more recency-focused configuration.
This illustrates the main role of the agent-facing interface: it allows the upper-level planner to compose long-horizon behavior from repeated, configurable calls to the same navigation foundation model.

\subsection{Navigation Harness: Trajectory Evidence and Context Compression}
\label{sec:agent_harness}

The interface above controls how \qwennav is called.
The remaining question is how the result of each navigation call is returned to the planner.
A raw \qwennav rollout contains dense egocentric observations, sampled history frames, low-level control traces, and a predicted waypoint trajectory.
Passing this full stream back into the planner dialogue would quickly exhaust the context budget.
On the other hand, returning only a success flag would discard the evidence needed for future planning.
We therefore use a lightweight trajectory-to-evidence interface.

For each navigation call, the harness converts the executed rollout into a compact evidence record:
\begin{quote}\small\sloppy
\texttt{\{}\\
\hspace*{1em}\texttt{subgoal:} ``Search the kitchen area for a mug'',\\
\hspace*{1em}\texttt{task\_mode:} \texttt{ObjNav},\\
\hspace*{1em}\texttt{config:} $\Phi_i$ \texttt{(main controls: } $B,\gamma,m$ \texttt{)},\\
\hspace*{1em}\texttt{progress:} ``entered kitchen, checked countertop and dining table'',\\
\hspace*{1em}\texttt{salient:} [``sink'', ``countertop'', ``round table'', ``no mug observed''],\\
\hspace*{1em}\texttt{outcome:} ``target not found'',\\
\hspace*{1em}\texttt{key\_frames:} [\#18, \#31]\\
\texttt{\}}
\end{quote}

This record is not intended to replace visual evidence.
Instead, it acts as a source-indexed summary of what happened during the navigation segment.
The textual fields tell the planner what was attempted, where the agent moved, what was observed, and how the sub-goal ended.
The key-frame identifiers point back to stored visual observations that can be retrieved later if the planner needs to re-inspect a scene or verify a candidate target.

The harness maintains a compact evidence notebook for long-horizon reasoning.
The notebook stores durable conclusions such as searched regions, candidate object locations, rejected hypotheses, landmark cues, and layout assumptions.
Unlike raw dialogue history, notebook entries are designed to survive context compression.
Later entries may revise earlier beliefs, but the update history remains auditable.
A typical entry might be:
\begin{quote}\small\sloppy
\texttt{[step 47]} Kitchen entered and searched; countertop and dining table checked. No mug observed. Corridor shelf remains a possible candidate region from key frame \#12.
\end{quote}

Together, trajectory evidence and the evidence notebook form a two-level memory.
By default, the planner reasons over compact textual records and notebook entries.
When text is insufficient, it can retrieve source images through visual recall using the stored key-frame identifiers.
This keeps the planner context concise while preserving access to detailed visual evidence.

The harness can also expose auxiliary vision-tool calls, such as object detection, scene understanding, and semantic grounding.
These tools support the planner's situational awareness, but they do not replace \qwennav as the core navigation model.
Their role is to provide additional evidence channels around the same planner--\qwennav loop: the planner selects a task mode and configuration, \qwennav executes the navigation segment, and the harness returns compact trajectory evidence for the next decision.

In summary, the agentic layer does not introduce a separate navigation policy.
Its main function is to make \qwennav's existing task modes and observation controls accessible to an upper-level planner, and to make completed rollouts usable for long-horizon reasoning through trajectory evidence, notebook memory, and context compression.
\section{Data}
\label{sec:data}


\begin{figure}[ht]
  \centering
  \includegraphics[width=0.85\linewidth, trim=120 150 200 0, clip]{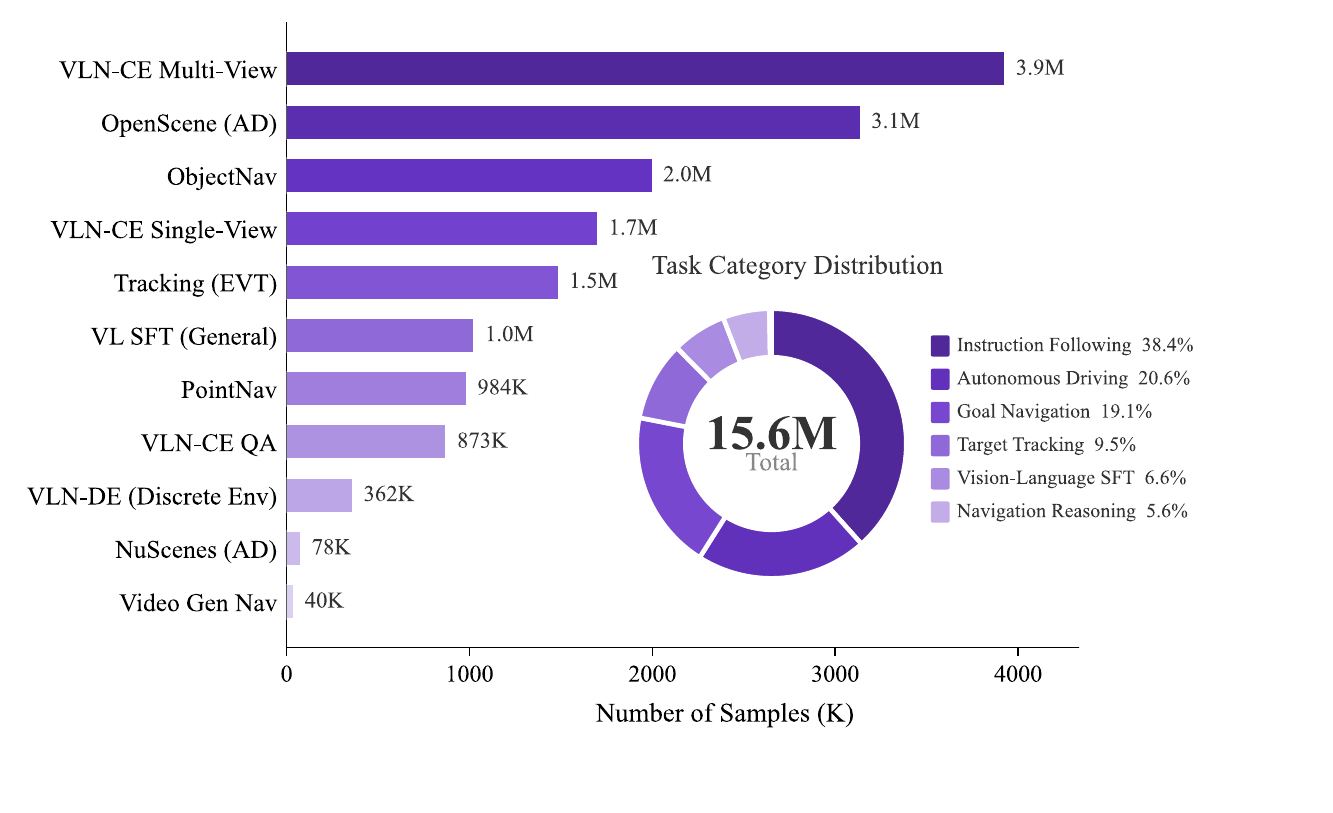}
  \caption{\textbf{Training data distribution.} \textit{Left:} Per-dataset sample counts across all navigation trajectory and vision-language sources. \textit{Right:} Aggregated distribution over task categories, totalling 15.6M training samples.}
  \label{fig:data_overview}
\end{figure}

\subsection{Navigation Trajectory Planning}

A key design principle of \qwennav is to train on a deliberately broad spectrum of navigation tasks rather than specialising in any single paradigm.
We structure the trajectory planning corpus around four capability dimensions that broaden the operational demands placed on the model: grounding language or geometric guidance into executable motion, exploring partially observed spaces to locate targets~\citep{zheng2025densegrounding,zheng2024denseg,peng2025proxytransformation}, interacting with dynamic agents whose future states must be anticipated in real time, and planning under high-speed, multi-agent, safety-critical dynamics in open-world environments.
These dimensions roughly span increasing environmental uncertainty, temporal dependence, and embodiment diversity, but they are not mutually exclusive in practice.
Rather, more complex navigation regimes often reuse simpler competencies while introducing additional perception, planning, and control requirements.
Jointly training across this spectrum encourages the model to develop a shared spatial-planning substrate that generalises beyond any single task family.
The full training corpus comprises 15.6M samples; \cref{fig:data_overview} summarises the per-dataset and per-category distributions.

Guided by these dimensions, we select five concrete task families as the instantiation of our training corpus:
\begin{itemize}[leftmargin=1.5em]
    \item \textbf{Instruction following} and \textbf{point-goal navigation} jointly ground guided navigation by pairing rich procedural language with compact directional or coordinate-based commands, thereby spanning the full granularity spectrum of task specification.
    \item \textbf{Object-goal navigation} realises the exploration dimension, requiring the agent to build implicit spatial maps and hypothesise target locations without step-by-step instructions.
    \item \textbf{Active target tracking} introduces dynamic interaction, demanding real-time motion anticipation and re-identification of moving persons in crowded scenes.
    \item \textbf{Autonomous driving} realises cross-embodiment planning under high-speed, multi-agent traffic dynamics with stringent safety constraints, an operational regime that shares the same underlying spatial-planning principles as indoor navigation yet demands substantially different perception range, action granularity, and decision latency.
\end{itemize}
These five tasks are chosen because they are \emph{mutually non-redundant}: each exercises a distinct combination of perception, planning, and control that the others leave undertrained, while together they provide sufficient coverage for these capability dimensions to reinforce one another during joint optimisation.

\subsubsection{Instruction Following}

Instruction following represents the language-rich form of guided navigation.
The agent must interpret natural language instructions alongside egocentric visual observations and plan waypoint trajectories to reach the described destination while adhering to the procedural milestones specified in the instruction.
This supervision trains fine-grained language-to-control grounding: spatial relations, landmark references, and temporal ordering in the instruction must be translated into executable motion.
We construct \textbf{5.63M} instruction-following training samples from two VLN-CE benchmarks~\citep{krantz2020beyond,xia2026habitat}, building on Matterport3D~\citep{chang2017matterport3d} scenes.

\textbf{VLN-CE R2R} (1,491K).
The Room-to-Room benchmark~\citep{krantz2020beyond} provides approximately 10K continuous navigation clips in indoor environments, each paired with a concise goal-oriented instruction.
We unroll each ground-truth trajectory with teacher forcing and record RGB observations in both single-camera (front only) and multi-camera (front, left, right, and rear) configurations at every discrete step, yielding 1,491K training samples after data balancing across view configurations, instruction refinement, and image quality enhancement.

\textbf{VLN-CE RxR} (4,140K).
The Room-Across-Room benchmark~\citep{anderson2020rxr} significantly expands R2R in both scale and complexity, featuring longer paths, richer spatial relationships, and multilingual instructions with dense landmark references.
Following the same teacher-forcing protocol, we extract 4,140K training samples from approximately 20K continuous clips across view configurations and augmentation variants.

All instruction-following samples from both sources undergo the image quality enhancement and instruction refinement pipelines described in Section~\ref{sec:data_augmentation}: simulator renders are transformed into photorealistic images via Qwen-Image-Edit~\citep{wu2025qwenimage} style transfer, and each instruction is paraphrased into multiple linguistically diverse variants using an LLM~\citep{qwen3}; the sample counts above include all resulting augmentation variants.

\subsubsection{Point Goal Navigation}

\begin{figure*}[t]
  \centering
  \includegraphics[width=\linewidth]{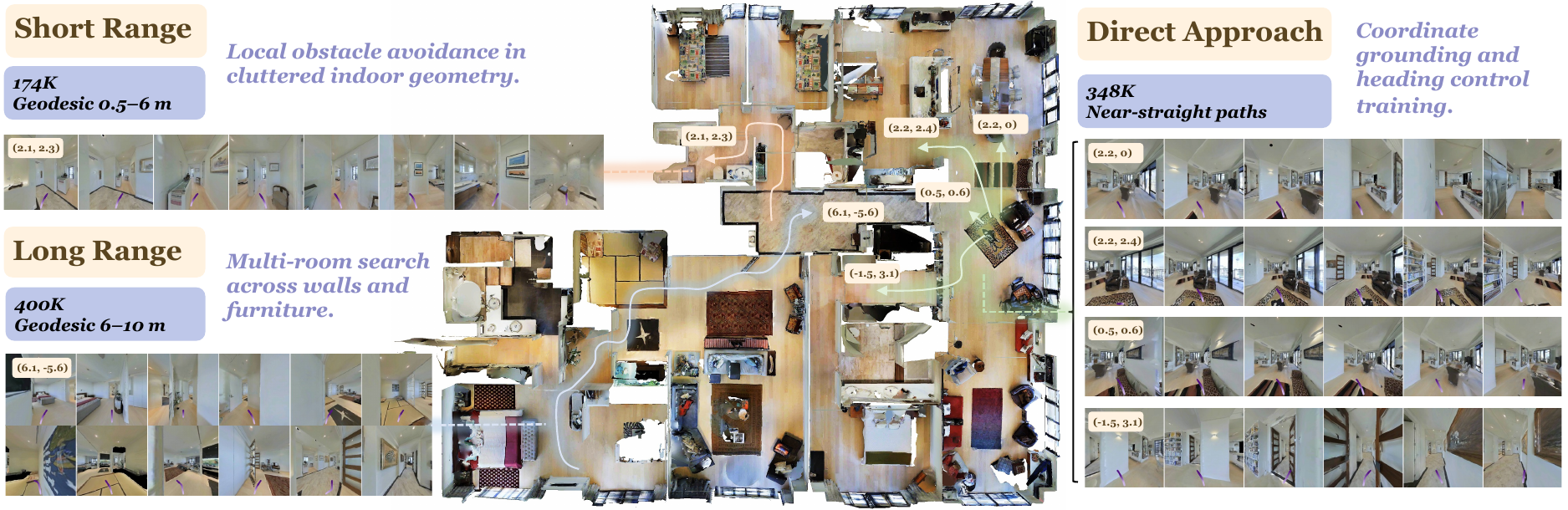}
  \caption{Visualization of the three coordinate-based point-goal navigation categories. \textbf{Direct Approach} (348K) targets near-straight paths along canonical egocentric directions. \textbf{Short Range} (174K) introduces local obstacle avoidance within cluttered indoor geometry. \textbf{Long Range} (400K) requires multi-room path search over extended horizons, navigating around walls and furniture.}
  \label{fig:pointnav_vis}
\end{figure*}

Point-goal navigation provides the language-sparse counterpart to instruction following.
The agent must reach a specified target given only its current visual observation and a compact task specification, without detailed natural language guidance.
Whereas instruction-following data emphasise procedural language grounding, point-goal data isolate geometric path planning, local obstacle avoidance, and smooth goal approach from minimal coordinate or command inputs.
We generate \textbf{984K} point-goal navigation training samples from Matterport3D~\citep{chang2017matterport3d} and HM3D~\citep{ramakrishnan2021hm3d} scenes using the Habitat simulator~\citep{savva2019habitat}, and organise them into a curriculum from primitive motion grounding to long-horizon path search.

\textbf{Coordinate-based point goals} (922K).
The agent receives the target position as a numeric coordinate in its egocentric frame together with its current pose, distance, and bearing angle to the goal.
We allocate this subset according to difficulty rather than sampling uniformly.
\begin{itemize}[leftmargin=1.5em]
    \item \textbf{Direct-approach point goals} (348K).
    Targets are placed along canonical egocentric directions or on an integer grid within $[-5,\,5] \times [-5,\,5]$\,m, and paths whose geodesic-to-euclidean ratio exceeds $1.15$ are rejected.
    This group corresponds to simplified point-of-interest navigation in which the target can be reached by a near-straight trajectory, focusing supervision on basic coordinate grounding, heading control, and smooth approach behaviour.
    \item \textbf{Short-range point goals} (174K; geodesic distance $0.5$--$6$\,m).
    These episodes introduce cluttered indoor geometry and local obstacle avoidance while keeping the destination within a limited spatial neighbourhood.
    Because they are relatively local and the visual evidence is often sufficient to infer a feasible detour, a moderate sample budget is sufficient.
    \item \textbf{Long-range point goals} (400K; geodesic distance $6$--$10$\,m).
    These episodes receive the largest allocation because they frequently lack direct line of sight to the target and require the agent to search over alternative routes, navigate around walls and furniture, and plan across rooms over extended horizons.
\end{itemize}
This distribution implements a curriculum in which the model first learns reliable coordinate-to-motion grounding, then local collision avoidance, and finally the more difficult path-searching behaviour needed for non-myopic navigation.

\textbf{Command-based motion primitives} (62K).
Instead of explicit coordinates, the agent receives a textual motion primitive.
\emph{Parameterised commands} specify both action and magnitude (\eg, ``\textit{Move forward 2.0 meters}'', ``\textit{Turn left 90 degrees}''), while \emph{bare commands} provide only the direction without a numeric target (\eg, ``\textit{Move forward}'', ``\textit{Turn left}''), requiring the model to infer an appropriate displacement from visual context.
This subset is intentionally small because such commands primarily anchor basic action semantics, such as moving forward by a specified distance or rotating by a specified angle; extensive repetition yields limited additional path-planning diversity compared with coordinate-based long-range episodes.

Point-goal navigation (PointNav) also serves as a controlled baseline for studying navigation competence because the goal specification is simple and the required behaviour can be evaluated without the confounding effect of complex language.
This simplicity makes PointNav particularly suitable for controlled observation-configuration and motion-dynamics augmentation; the corresponding front-view-only training protocol and simulator perturbations are described in Section~\ref{sec:data_augmentation}.

Each training sample contains a uniformly sampled set of history frames, the task specification (coordinates or command), and an 8-waypoint future trajectory.
Within $1.5$\,m of the goal, deceleration trajectories with linearly decreasing step sizes are generated to teach smooth stopping behaviour.
Forward steps are subsampled at a $45\%$ inclusion rate to rebalance the action distribution, while turns and stop actions are always retained. 

\subsubsection{Object-Goal Navigation}

\begin{figure*}[t]
  \centering
  \includegraphics[width=\linewidth]{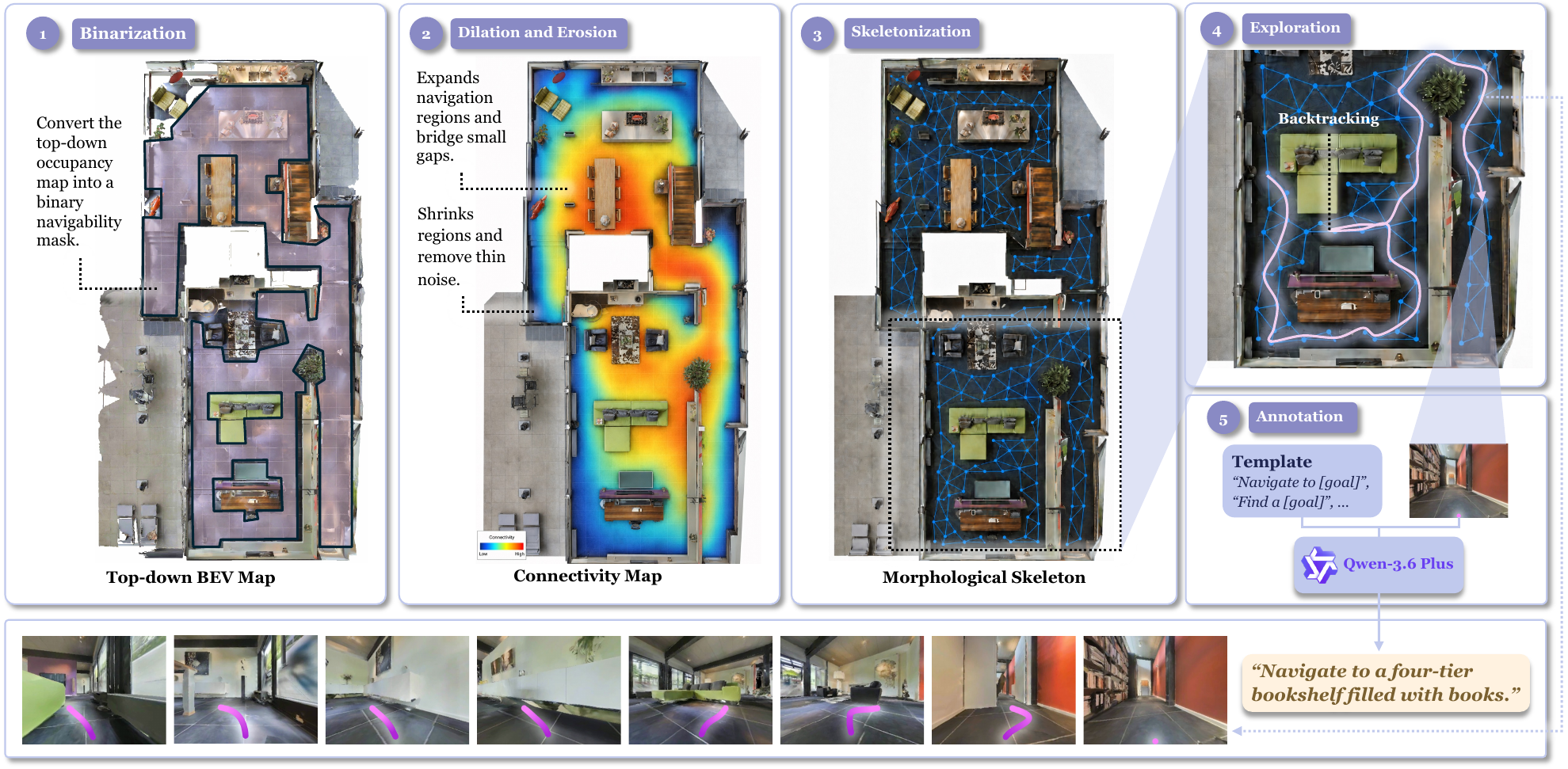}
  \caption{Object-goal navigation data generation pipeline. \textbf{(1)} The top-down occupancy map is binarised into a navigability mask. \textbf{(2)} Morphological dilation and erosion expand navigable regions and remove thin noise. \textbf{(3)} Skeletonisation extracts the medial-axis graph of the navigable space. \textbf{(4)} An exploration trajectory is generated by traversing the skeleton with backtracking at dead ends. \textbf{(5)} A VLM annotates the goal object at the terminal viewpoint, producing open-vocabulary goal specifications embedded in diverse instruction templates.}
  \label{fig:objnav_pipeline}
\end{figure*}

Object-goal navigation addresses goal-directed exploration under partial observability.
The agent receives a semantic object category or open-vocabulary object description, rather than a complete route, and must search the environment until an instance of the target is reached.
This setting exercises capabilities that are difficult to induce from route-following data alone: maintaining an implicit memory of explored regions, using scene priors to hypothesise likely object locations, and revising the route when the target is absent from the current field of view.

We construct \textbf{2,000K} object-goal navigation training samples from Matterport3D~\citep{chang2017matterport3d} and HM3D~\citep{ramakrishnan2021hm3d} reconstructed scenes, including open-vocabulary object annotations from HM3D-OVON~\citep{yokoyama2024hm3d}.

\textbf{Skeleton-based exploration trajectories.}
A key challenge is that standard shortest-path followers produce straight-to-goal trajectories that do not reflect realistic search behaviour: in practice, an agent exploring for an unseen object must traverse corridors, inspect rooms, and backtrack from dead ends before locating the target.
We therefore generate trajectories using a skeleton-based exploration strategy over the scene's navigable space.
Given the simulator's top-down occupancy map, we first extract the largest connected component of the navigable area and then apply morphological skeletonisation~\citep{zhang1984fast} to reduce it to a one-pixel-wide medial-axis graph.
Short spurious branches are pruned to suppress redundant bifurcation points.
On this skeleton graph, a goal location is sampled at random, and the agent starts from a random navigable position.
The path is planned by traversing the skeleton: at each junction the agent randomly selects a branch; upon reaching a dead end it backtracks and explores an alternative branch, continuing until the branch containing the goal is reached.
Because the skeleton lies along the medial axis of navigable regions, the resulting paths maintain a safe distance from walls and obstacles.
The raw skeleton path is then smoothed via cubic-spline interpolation: waypoints sampled at regular intervals along the skeleton are fitted with a parametric cubic spline, and the final trajectory points are sampled from this spline at a fixed step size of $0.25$\,m, producing smooth, physically plausible motion.

\textbf{VLM-based goal annotation.}
Each trajectory requires an associated object goal.
Rather than relying on a fixed category taxonomy, we adopt a VLM-in-the-loop annotation strategy that produces open-vocabulary goals by construction.
At the terminal point of a generated trajectory, the agent is oriented towards a random direction and the resulting egocentric image is submitted to a vision-language model, which is asked to identify a salient, reachable object in the scene.
If the VLM confirms the presence of a suitable target, the trajectory is retained and the object name returned by the VLM is used as the goal specification; otherwise the trajectory is discarded.
Because the VLM is free to name any visible object without a predefined label set, the resulting goal vocabulary is inherently open and covers a long tail of everyday objects beyond the fixed categories of standard ObjectNav benchmarks.
The goal is then embedded into diverse natural-language instruction templates (\eg, ``\textit{navigate to the \{goal\_object\}}'', ``\textit{find and reach the \{goal\_object\}}'') to provide linguistic variation during training.
Each training sample contains uniformly sampled history frames, the task specification, and an 8-waypoint future trajectory, following the same format used by the other navigation task families.
The same observation-configuration and image-quality augmentations described in Section~\ref{sec:data_augmentation} are applied, including camera height and field-of-view randomisation.

\subsubsection{Target Tracking}

Active visual tracking extends the trajectory corpus from static goal reaching to dynamic interaction.
The agent must identify a person described by a natural language query (\eg, ``\textit{Follow the man in the blue t-shirt}''), maintain pursuit in crowded indoor environments, and keep an appropriate following distance while avoiding obstacles.
Compared with object-goal navigation, this setting introduces time-critical requirements including motion anticipation, occlusion handling, target re-identification, and balancing pursuit with safety constraints.
We collect \textbf{1,486K} training samples from the EVT-Bench dataset~\citep{wang2025trackvla}, which covers diverse indoor scenes populated with hundreds of digital avatars whose appearance and motion are controlled independently.
Following the benchmark protocol~\citep{wang2025trackvla}, we focus on the \emph{Single Target Tracking} (STT) split, where a unique target is specified per episode and no distractors are introduced.

Each training sample contains the current egocentric observation, a short history of past frames, the textual target description, and the ground-truth future trajectory.
We additionally apply the same image-quality and observation-configuration augmentations used in the instruction-following data to improve cross-scene and cross-sensor generalisation.

\subsubsection{Autonomous Driving}
Autonomous driving is incorporated as a cross-embodiment source of trajectory supervision rather than as a separate top-level capability class~\citep{hu2023planning,li2025recogdrive,Peng_2026_CVPR,liang2026planning}.
It requires an agent to reason over complex, dynamic, and safety-critical traffic environments and to predict feasible future trajectories under diverse road geometries, traffic rules, and agent interactions.
Unlike indoor embodied navigation, driving scenarios involve higher-speed motion, structured lane topology, long-range perception, and dense multi-agent dynamics.
These differences stress the same spatial-planning substrate under a substantially different operating regime, exposing the model to long-horizon planning, interaction-aware motion prediction, and safety-constrained control.
To equip the model with transferable planning capabilities across both embodied and vehicle-centric settings, we incorporate large-scale driving data into training and cast autonomous driving as a unified waypoint prediction problem conditioned on multimodal observations.
We construct the autonomous driving training set from two complementary sources: nuScenes (78K)~\citep{caesar2020nuscenes} and OpenScene (3,138K)~\citep{openscene2023}.
Together, these datasets provide diverse urban scenes, rich sensor observations, map-aware traffic contexts, and a broad spectrum of driving behaviours ranging from routine lane following to complex maneuvers such as turning, merging, yielding, and obstacle avoidance.
By integrating them into a common trajectory-planning format, we obtain approximately \textbf{3.2M} autonomous-driving trajectory-supervision instances.
Here, the reported number refers to the total number of annotated supervision variants rather than the number of distinct raw driving trajectories.
Specifically, a single underlying trajectory may be instantiated into multiple conditioning variants, depending on whether navigation instructions, ego-state information, and historical ground-truth trajectory priors are provided.
This design allows the same driving behavior to be observed under different levels of prior information, thereby encouraging the model to learn trajectory planning that is robust to heterogeneous inputs and transferable across different settings.

Each supervision instance is centered on predicting the ground-truth future trajectory at the current frame, conditioned on multimodal observations and a configurable set of auxiliary priors.
All instances include multi-view camera observations, while different annotation variants may additionally provide navigation instructions, the current ego-vehicle state, and/or a short history of past ground-truth trajectories.
Thus, the autonomous-driving data is not treated as a single fixed-input prediction task, but as a family of trajectory-planning tasks with different available context.
This unified yet flexible representation enables the model to reason over visual scene context, vehicle motion, temporal dynamics, and high-level route intent when available, while also learning to remain effective when some priors are absent.

\subsection{Autogenerated Navigation Data with a Video Generator}
\label{sec:autogen_data}

\begin{figure*}[t]
    \centering
    \includegraphics[width=0.9\linewidth]{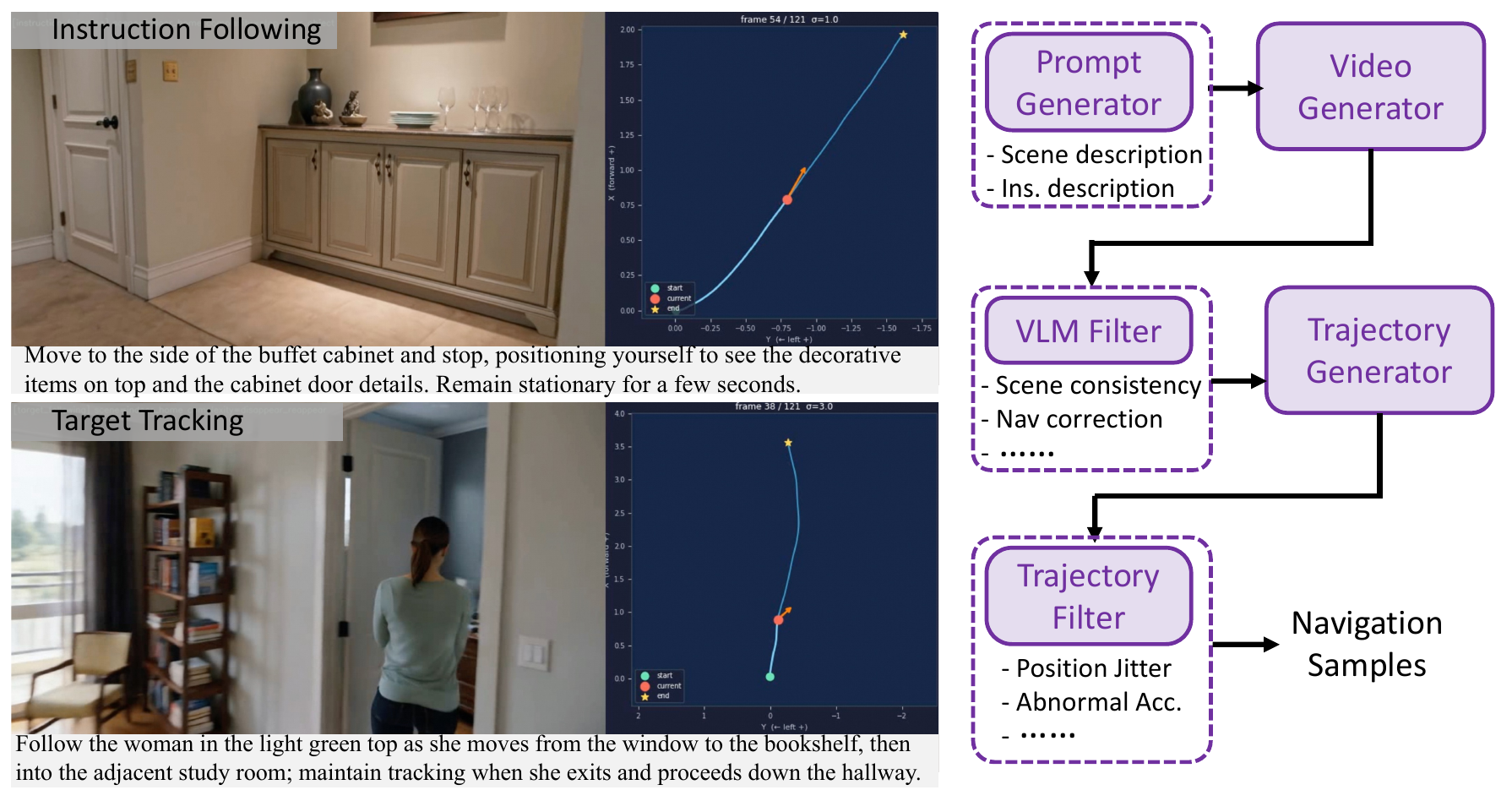}
    \caption{\textbf{Autogenerated navigation data pipeline.} \textit{Right:} A large language model first generates paired video prompts and navigation instructions; a text-to-video model then synthesises first-person egocentric videos, which are filtered by a vision-language model for quality before a monocular depth-and-pose estimator extracts 2-D trajectories; a final kinematic filter removes physically implausible samples. \textit{Left:} Two example outputs covering instruction following (top) and target tracking (bottom), each showing a generated video frame alongside the recovered bird's-eye-view trajectory.}
    \label{fig:video_2_traj}
\end{figure*}

While simulator-derived trajectories provide precise ground-truth supervision, they are constrained by the availability of reconstructed 3-D assets and exhibit a persistent visual domain gap relative to real-world observations.
To complement the simulator corpus with diverse, photorealistic training data that does not require 3-D scene reconstruction, we introduce a fully automated pipeline that converts text-to-video (T2V) generations into navigation trajectory samples.
The pipeline covers two core capabilities, \emph{instruction following} and \emph{target tracking}, and produces \textbf{40K} training samples through five sequential stages illustrated in \cref{fig:video_2_traj}.

\begin{itemize}[leftmargin=1.5em]
    \item \textbf{Stage 1: Prompt and instruction generation.} A large language model generates paired outputs for each sample: a detailed video prompt describing a first-person navigation scenario, and a corresponding natural-language navigation instruction.
    Prompts are drawn from a scene-complexity matrix spanning 12 scene categories (\eg, home, office, shopping mall, hospital, park, street) and 7 interaction complexities (\eg, obstacle avoidance, target occlusion and reappearance, crowd navigation, sudden direction changes), yielding 29 distinct combinations that ensure broad environmental and behavioural coverage.
    \item \textbf{Stage 2: Text-to-video synthesis.} Each video prompt is fed to a T2V model, which renders a first-person egocentric navigation video of approximately five seconds.
    The resulting clips depict the agent moving through the described scene with realistic lighting, textures, and dynamic elements such as pedestrians, without requiring any 3-D assets or physics simulation.
    \item \textbf{Stage 3: VLM-based video quality filtering.} A vision-language model evaluates every generated video against capability-specific quality dimensions.
    For instruction-following videos the assessment covers scene consistency, navigation correctness, goal arrival, stopping behaviour, motion continuity, and collision avoidance; target-tracking videos are additionally evaluated on target visibility, pursuit behaviour, and identity consistency across frames.
    Only videos that pass all applicable criteria are retained.
    \item \textbf{Stage 4: Trajectory extraction.} A monocular depth-and-pose estimation model recovers per-frame camera-to-world poses from each retained video.
    The estimated poses are converted from camera coordinates into the robot-centric ground-plane frame, producing a 2-D trajectory $[\,x,\,y,\,\mathrm{yaw}\,]$ per frame, where $x$ and $y$ denote forward and lateral displacement in metres and $\mathrm{yaw}$ is the heading angle in radians, all relative to the first frame.
    \item \textbf{Stage 5: Trajectory quality filtering.} A rule-based kinematic filter removes trajectories that exhibit physically implausible characteristics, including near-zero total displacement, excessive positional or heading jitter, single-step teleportation, abnormal acceleration, and high-frequency noise detected via spectral analysis.
    Each check uses a tunable threshold, and a trajectory is discarded if it triggers any single criterion.
\end{itemize}

The resulting dataset provides trajectory supervision grounded in photorealistic, T2V-generated imagery across a wide variety of real-world-like scenes, complementing the simulator-derived corpus with greater visual diversity and reduced domain gap.

\subsubsection{Navigation Data Augmentation}
\label{sec:data_augmentation}

\begin{figure*}[t]
  \centering
  \includegraphics[width=\linewidth]{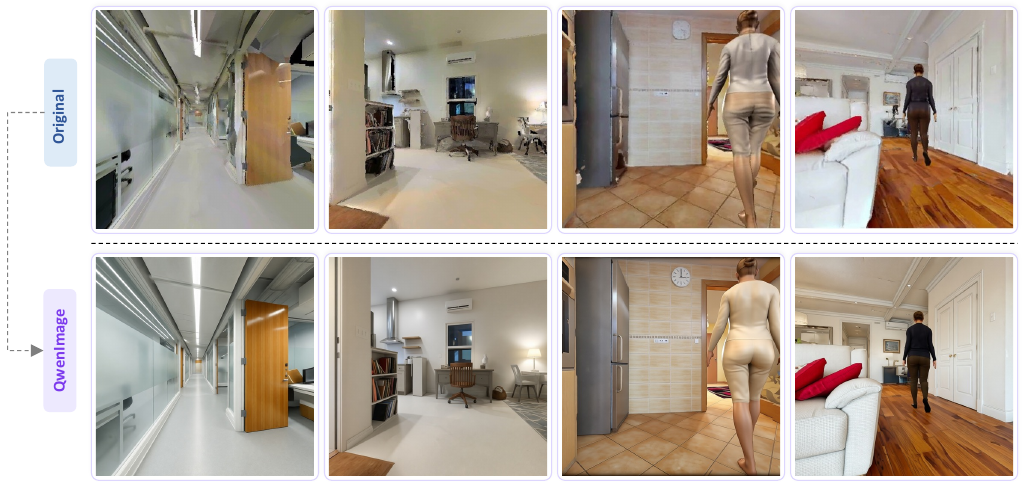}
  \caption{Visual comparison between original Habitat simulator renders (top) and Qwen-Image-Edit~\citep{wu2025qwenimage} (bottom).}
  \label{fig:qwenim_vis}
\end{figure*}

To enhance robustness and data diversity, we apply task-dependent augmentation strategies to simulator-derived navigation data.
Instruction refinement is restricted to language-conditioned trajectories, whereas visual refinement, observation-configuration perturbation, and speed variation are applied to simulator-derived trajectory samples when supported by the source data.
\begin{itemize}[leftmargin=1.5em]
    \item \textbf{Instruction refinement.} We employ an LLM-based paraphrasing pipeline~\citep{qwen3} that first deduplicates instructions by trajectory identifier and then generates multiple paraphrased variants per unique instruction.
    The generation prompt instructs the model to preserve all spatial directions and relative landmarks, vary landmark nouns with synonyms or hypernyms, diversify sentence structure and action verbs, and correct any grammatical noise, so that each variant reads as if written by a different annotator.
    Three variants are generated per unique instruction, effectively tripling the linguistic diversity of the corpus.
    \item \textbf{Image quality enhancement.} The sim-to-real visual gap remains a major bottleneck for transferring simulation-trained policies to real environments. We build a scalable refinement pipeline based on Qwen-Image-Edit~\citep{wu2025qwenimage}, which applies prompt-guided diffusion-based style transfer to every rendered observation.
    Given a source image and a natural-language editing prompt (\eg, ``\textit{Convert the rendered image into a photorealistic photograph while preserving spatial layout}''), the model generates a visually refined version that retains the original geometric layout while exhibiting more realistic textures, lighting, and material appearance.
    \item \textbf{Observation and camera augmentation.} Camera height is uniformly sampled in $[0.5,\,1.5]$\,m, horizontal field of view (HFoV) in $[90^\circ,\,120^\circ]$, and image aspect ratio between 2:1 and 4:3, so the model learns to navigate under diverse sensor configurations without task-specific retraining.
    For PointNav, we additionally perturb the robot's initial heading and egocentric viewpoint, and explicitly construct front-view-only variants in which the model receives only the front camera observation rather than panoramic inputs.
    This setting matches deployment scenarios where only a forward-facing camera is available.
    \item \textbf{Speed augmentation.} We generate trajectory data at multiple speed regimes to expose the model to diverse motion dynamics. The \emph{standard-speed} variant uses the default Habitat action discretisation ($0.25$\,m forward steps, $15^\circ$ turns). The \emph{low-speed} variant replays the same planned trajectories at finer temporal granularity with randomised sub-step sizes ($0.05$ to $0.25$\,m), while PointNav trajectories are further replayed under varied motion scales to improve robustness across robot actuation speeds.
\end{itemize}

\subsection{Vision Language Data}
\label{sec:vlm_to_vla}

\begin{figure*}[h]
  \centering
  \includegraphics[width=0.91\linewidth]{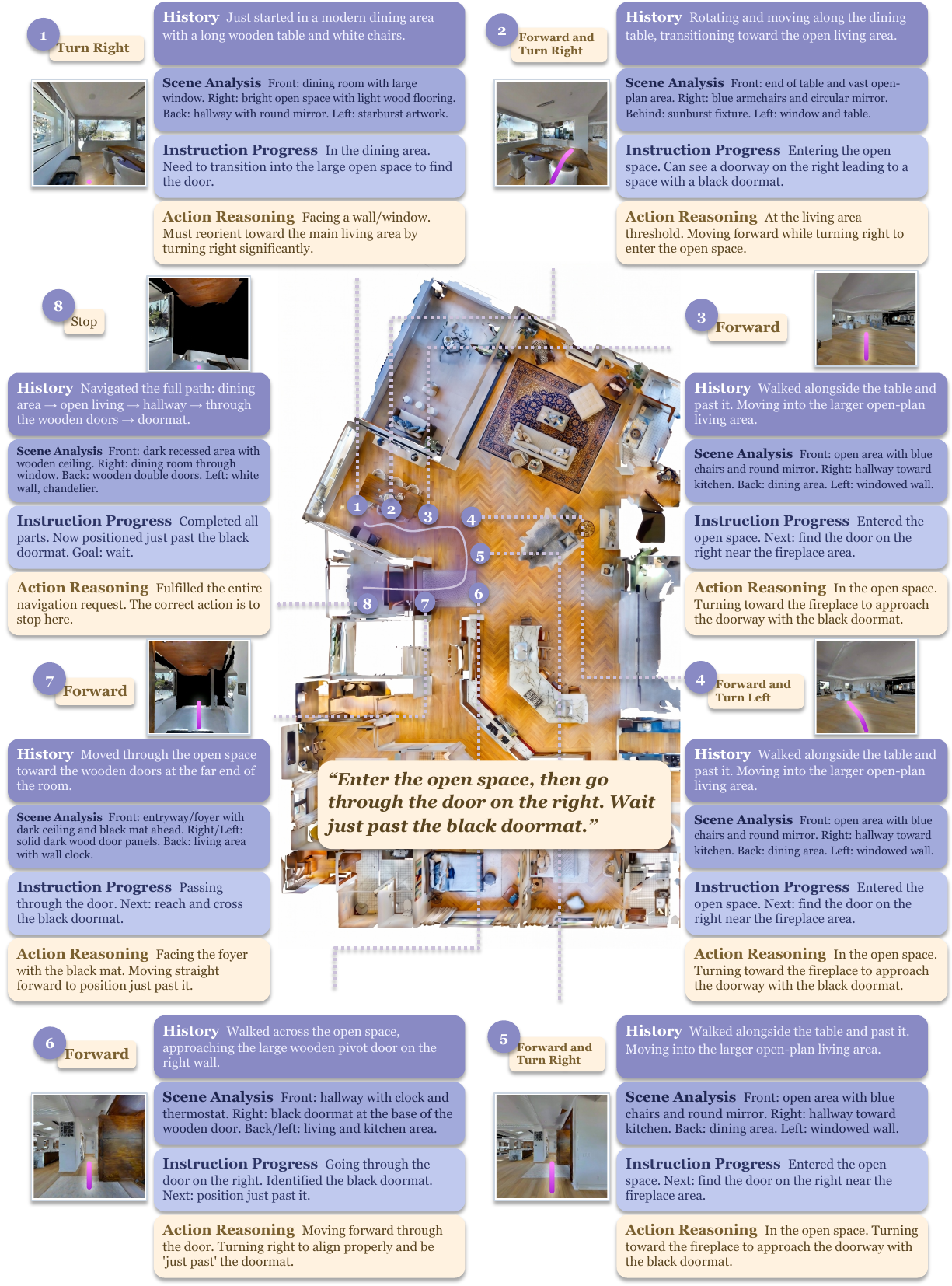}
  \caption{Visualization of structured multi-perspective reasoning along a complete navigation trajectory. Given the instruction ``\textit{Enter the open space, then go through the door on the right. Wait just past the black doormat},'' the agent executes eight sequential steps (numbered on the floor plan). At each step, a four-component reasoning chain is produced: \textbf{History} summarises the journey so far, \textbf{Scene Analysis} describes the current multi-view observations, \textbf{Instruction Progress} tracks completed and remaining sub-goals, and \textbf{Action Reasoning} derives the next action. This structured decomposition is used to generate navigation-specific reasoning supervision for VLM-to-VLA transfer.}
  \label{fig:reasoning_vis}
\end{figure*}

Robust action prediction rests on perceptual and reasoning capabilities that cut across all navigation task families and cannot be acquired reliably from trajectory supervision alone.
Recognising objects in cluttered scenes, reading text on signs, reasoning about spatial layouts, comparing multiple views, and interpreting novel visual contexts are prerequisites for navigation in unseen environments.
We co-train navigation trajectories with approximately \textbf{1.0M} general vision-language samples and \textbf{873K} navigation-specific reasoning samples, organised into two complementary groups that collectively maintain and strengthen the perceptual substrate required for the VLM-to-VLA transfer.

\textbf{General vision-language data} (${\sim}$1.0M).
We incorporate training samples spanning eight complementary categories:
visual question answering (${\sim}$669K; VQAv2~\citep{goyal2017making}, DVQA~\citep{dvqa}, FigureQA~\citep{figureqa}, CLEVR~\citep{johnson2017clevr}, \etc),
image captioning (${\sim}$6K),
visual grounding (${\sim}$178K; RefCOCO~\citep{kazemzadeh2014referitgame}, COCO~\citep{lin2014microsoft}, Objects365~\citep{shao2019objects365}),
instruction following (${\sim}$30K),
multi-image reasoning and comparison (${\sim}$38K),
general visual question answering (${\sim}$76K),
object and landmark recognition (${\sim}$16K),
and STEM problem solving (${\sim}$14K).
Rather than serving merely as a regulariser against catastrophic forgetting, these samples preserve and extend the in-the-wild visual understanding that underpins downstream action prediction---object and landmark recognition, spatial reasoning, text parsing, and multi-image analysis.
All samples follow a unified conversation format and are co-trained with navigation data, so that this perceptual substrate remains intact as the model acquires embodied competence.

\textbf{Navigation-specific reasoning} (873K).
Navigation reasoning is inherently domain-specific: deciding whether to turn left at a hallway intersection demands spatial understanding that differs qualitatively from general visual question answering.
We extend NavGPT-2~\citep{zhou2024navgpt2} and construct two complementary reasoning formats from existing VLN trajectory data.
\emph{Free-form QA} targets key decision points within each trajectory, the start step, the final step, and transition points where the dominant action category switches between forward motion and turning.
Each sample's 8-step future trajectory is classified into one of twelve fine-grained action classes (six pure turns: slight, medium, and large in both left and right directions: four forward-with-turn combinations, straight-ahead movement, and stop) and cast as a question--answer pair, training the model to articulate spatial reasoning in natural language before committing to actions.
\emph{Structured multi-perspective reasoning} extends this idea to multi-view contextual analysis.
For each selected sample, the agent's historical front-view observations (up to eight uniformly sampled frames) and current four-view panoramic images (front, right, back, left) are combined with the navigation instruction, ground-truth action label, and trajectory statistics into a structured prompt.
A vision-language model then generates a structured reasoning chain comprising four components:
\begin{itemize}[leftmargin=1.5em]
    \item \textbf{History reasoning.} A first-person narrative summarising the journey so far based on historical views.
    \item \textbf{Scene analysis.} A description of what is visible in each of the four current views.
    \item \textbf{Instruction progress.} An assessment of completed and remaining sub-goals relative to the original instruction.
    \item \textbf{Action reasoning.} A concise reasoning chain concluding with the predicted action and a confidence score.
\end{itemize}
Each sample is subsequently decomposed into four independent question--answer pairs, yielding separate fine-tuning signals for history comprehension, scene understanding, progress tracking, and action prediction.
By requiring the model to reconstruct its trajectory context, analyse the current scene, and assess instruction progress before deriving an action, both formats enforce a systematic reasoning process that serves as an inductive bias for the VLM-to-VLA transfer: the language-mediated reasoning, once internalised, acts as a transferable scaffold that improves both QA accuracy and the quality of predicted action trajectories.

\textbf{Discrete multi-round navigation data} (362K).
To complement the continuous-environment trajectory data, we incorporate approximately 362K training samples derived from discrete VLN datasets by reformulating graph-based navigation trajectories into a multi-round, multi-image conversation format.
Following the VLN-MME evaluation framework~\citep{zhao2025vlnmme}, each navigation step is formulated as a single-step action prediction task cast in a multiple-choice question format: the agent receives four perspective-view images (front, right, back, left) extracted from the panoramic observation rendered by the Matterport3D Simulator~\citep{chang2017matterport3d}, together with a set of candidate next-viewpoints annotated with visual markers as answer options, and selects the correct action among them.
Rather than rendering observations in the Habitat simulator, we use the Matterport3D Simulator, which produces higher-definition photorealistic panoramic views directly from the original 3-D scans.
Unlike the continuous-environment action data, which are decomposed into independent single-step samples, this discrete formulation allows each trajectory to be maintained as a complete multi-turn dialogue, preserving the full decision history within the conversation context and enabling the model to leverage prior turns for spatial reasoning without requiring an explicit history encoding module.
Through this pipeline, we incorporate additional VLN datasets beyond R2R and RxR: CVDN~\citep{thomason2020cvdn}, which introduces dialog-conditioned navigation requiring multi-turn linguistic grounding; SOON~\citep{zhu2021soon}, which targets situated object-oriented navigation with fine-grained spatial goal specifications; REVERIE~\citep{qi2020reverie}, which pairs remote referring expressions with object localisation; and SRDF~\citep{wang2024srdf}, which bootstraps large-scale augmented trajectories via a self-refining data flywheel across diverse indoor environments.
This approach enables us to significantly scale the number of training trajectories and their linguistic diversity while preserving the multi-turn reasoning capabilities that single-step formulations sacrifice.

\section{Experiments}
\label{sec:experiment}

\subsection{Deployment}

\begin{figure*}[t]
  \centering
  \includegraphics[width=0.95\linewidth]{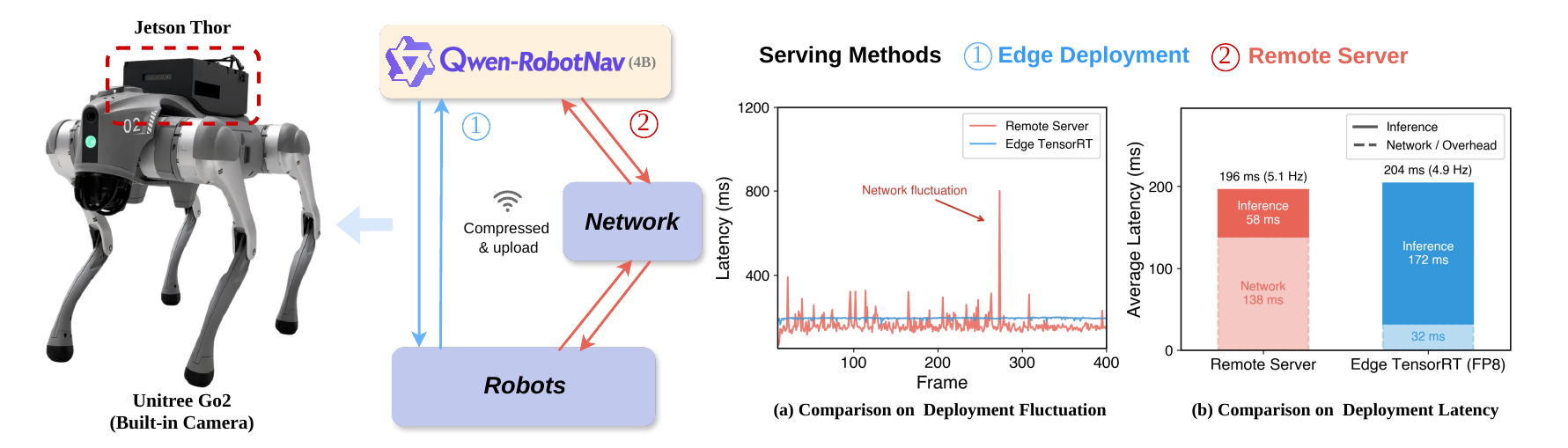}
  \caption{Deployment architecture and latency comparison of \qwennav-4B on Unitree Go2. Left: remote-server and edge deployment pipelines, where edge inference runs on Jetson Thor with FP8 quantization and TensorRT acceleration. Right: latency evaluation, including (a) per-frame latency fluctuations and (b) average latency breakdown for both deployments.}
  \label{fig:deployment_latency}
\end{figure*}

For real-world deployment, we explore both cloud-based and on-device inference for \qwennav. In the cloud setting, the robot uploads compressed observations and text instructions over the network to a remote server for trajectory prediction, and receives the planned actions for execution. This setup supports larger models with strong inference capability and faster server-side inference, but introduces communication overhead and relies on network stability. In the on-device setting, inference runs directly on the robot via an NVIDIA Jetson Thor, eliminating transmission delay and improving robustness, at the cost of being constrained by onboard compute and bandwidth resources.

Taking network transmission into account, the remote-server deployment still achieves lower average end-to-end latency than on-device inference, with 196\,ms (5.1\,Hz) versus 204\,ms (4.9\,Hz). The trade-off is higher sensitivity to network conditions, leading to larger latency variance and occasional spikes. By eliminating network transmission, on-device inference provides more consistent latency and is therefore more robust for latency-sensitive tasks such as  tracking. Overall, remote-server inference is faster on average, whereas on-device inference is more stable and reliable in real-world deployment.

\subsection{Evaluation}
\label{sec:experiment_main}

We evaluate \qwennav on three standard embodied navigation benchmarks: vision-and-language navigation in continuous environments (VLN-CE), open-vocabulary object-goal navigation (OVON), and active visual tracking (EVT-Bench). Results are compared against two recent navigation foundation model baselines (NavFoM~\citep{zhang2025embodied} and ABot-N0~\citep{abotN0}) as well as established specialist methods.

\subsubsection{Vision-and-Language Navigation}

\begin{table}[t]
  \centering
  \small
  \caption{\textbf{VLN-CE Val-Unseen results} under monocular and panoramic observation settings. \textbf{Bold}: best within each setting; \underline{underline}: second best.}
  \label{tab:vlnce}
  \setlength{\tabcolsep}{3.5pt}
  \begin{tabular}{lcccccccc}
    \toprule
    \multirow{2}{*}{\textbf{Method}} &
      \multicolumn{4}{c}{\textbf{R2R Val-Unseen}} &
      \multicolumn{4}{c}{\textbf{RxR Val-Unseen}} \\
    \cmidrule(lr){2-5} \cmidrule(lr){6-9}
    & NE$\downarrow$ & OS$\uparrow$ & SR$\uparrow$ & SPL$\uparrow$
    & NE$\downarrow$ & nDTW$\uparrow$ & SR$\uparrow$ & SPL$\uparrow$ \\
    \midrule
    \multicolumn{9}{l}{\textit{Monocular}} \\
    NaVid~\citep{zhang2024navid}
      & 5.72 & 49.2 & 41.9 & 36.5
      & 5.72 & -- & 45.7 & 38.2 \\
    Uni-NaVid~\citep{zhang2024uni}
      & 5.58 & 53.3 & 47.0 & 42.7
      & 6.24 & -- & 48.7 & 40.9 \\
    NaVILA~\citep{cheng2024navila}
      & 5.22 & 62.5 & 54.0 & 49.0
      & 6.77 & 58.8 & 49.3 & 44.0 \\
    StreamVLN~\citep{wei2025streamvln}
      & 4.98 & 64.2 & 56.9 & 51.9
      & 6.22 & 61.9 & 52.9 & 46.0 \\
    DualVLN~\citep{wei2025ground}
      & \textbf{4.05} & 70.7 & 64.3 & 58.5
      & 4.58 & \textbf{70.0} & 61.4 & 51.8 \\
    InternVLA-N1~\citep{internvlan1}
      & 4.83 & 63.3 & 58.2 & 54.0
      & 5.91 & 65.3 & 53.5 & 46.1 \\
    \textbf{\qwennav-4B}
      & \underline{4.22} & \textbf{73.6} & \textbf{66.9} & \textbf{60.5}
      & \textbf{4.15} & 68.6 & \underline{71.3} & \underline{61.5} \\
    \textbf{\qwennav-8B}
      & 4.36 & \underline{72.7} & \underline{65.7} & \underline{59.6}
      & \underline{4.16} & \underline{69.9} & \textbf{73.4} & \textbf{63.5} \\
    \midrule
    \multicolumn{9}{l}{\textit{Panoramic}} \\
    NavFoM~\citep{zhang2025embodied}
      & 4.61 & 72.1 & 61.7 & 55.3
      & 4.74 & 65.8 & 64.4 & 56.2 \\
    ABot-N0~\citep{abotN0}
      & 3.78 & 70.8 & 66.4 & 63.9
      & 3.83 & -- & 69.3 & 60.0 \\
    OmniNav~\citep{omninav}
      & \underline{3.74} & 74.6 & \underline{69.5} & \underline{66.1}
      & \underline{3.77} & -- & 73.6 & 62.0 \\
    AstraNav-World~\citep{astranavworld}
      & 3.86 & 73.9 & 67.9 & 65.4
      & 3.82 & -- & 72.9 & 61.5 \\
    \textbf{\qwennav-4B}
      & 3.80 & \underline{77.2} & \underline{69.5} & 63.6
      & 3.80 & \underline{71.9} & \underline{75.2} & \underline{65.0} \\
    \textbf{\qwennav-8B}
      & \textbf{3.53} & \textbf{78.5} & \textbf{72.1} & \textbf{66.6}
      & \textbf{3.58} & \textbf{72.5} & \textbf{76.5} & \textbf{65.7} \\
    \bottomrule
  \end{tabular}
\end{table}



\noindent \textbf{VLN-CE.}
Table~\ref{tab:vlnce} reports VLN-CE R2R and RxR Val-Unseen results under both monocular and panoramic observation settings.
Under the panoramic setting, \qwennav-8B achieves \textbf{72.1\%} SR and \textbf{66.6\%} SPL on R2R, surpassing NavFoM by {10.4\%} SR and ABot-N0 by {5.7\%} SR. On the longer-horizon RxR benchmark, \qwennav-8B reaches \textbf{76.5\%} SR and \textbf{72.5} nDTW, outperforming NavFoM by {12.1\%} SR and ABot-N0 by {7.2\%} SR.
Under the monocular setting, \qwennav remains competitive despite the significantly reduced field of view. \qwennav-4B achieves {66.9\%} SR and {60.5\%} SPL on R2R, surpassing the strongest monocular baseline DualVLN by {2.6\%} SR and {2.0\%} SPL. On RxR, \qwennav-8B reaches {73.4\%} SR and {63.5\%} SPL, outperforming DualVLN by {12.0\%} SR and {11.7\%} SPL, demonstrating particularly strong gains on long-horizon instructions.
The consistent improvements across both observation settings validate the effectiveness of task-adaptive token allocation and the generality of \qwennav: it delivers state-of-the-art performance regardless of the input modality, whether operating with a single forward-facing camera or a full panoramic observation.


\begin{table}[t]
  \centering
  \small
  \caption{\textbf{Navigation foundation model comparison on VLNVerse~\citep{vlnverse} test split.} TL: Trajectory Length; NE: Navigation Error; SR: Success Rate; OSR: Oracle Success Rate; SPL: Success weighted by Path Length. \textbf{Bold}: best; \underline{underline}: second best.}
  \label{tab:vlnverse}
  \resizebox{\columnwidth}{!}{%
  \begin{tabular}{lcccccccccc}
    \toprule
    \multirow{2}{*}{\textbf{Method}} &
      \multicolumn{5}{c}{\textbf{Fine-grained}} &
      \multicolumn{5}{c}{\textbf{Coarse-grained}} \\
    \cmidrule(lr){2-6} \cmidrule(lr){7-11}
    & TL & NE$\downarrow$ & OSR$\uparrow$ & SR$\uparrow$ & SPL$\uparrow$
    & TL & NE$\downarrow$ & OSR$\uparrow$ & SR$\uparrow$ & SPL$\uparrow$ \\
    \midrule
    InternVLA-N1~\citep{internvlan1}
      & 9.23  & 5.68 & 38.69 & 28.95 & 25.00
      & 4.00  & 5.71 & 23.32 & 17.51 & 16.54 \\
    Uni-NaVid~\citep{zhang2024uni}
      & 12.35 & 4.11 & 67.02 & 45.74 & 26.91
      & 10.33 & 5.05 & 42.70 & 29.68 & 13.47 \\
    NavFoM~\citep{zhang2025embodied}
      & 8.58  & 4.13 & 69.68 & 51.59 & 32.40
      & 6.52  & 4.93 & 45.93 & 38.02 & 23.15 \\
    \midrule
    \textbf{\qwennav-4B}
      & 7.99 & \underline{3.05} & \underline{70.36} & \underline{62.61} & \underline{56.22}
      & 7.39 & \underline{4.40} & \underline{51.03} & \underline{41.25} & \underline{37.37} \\
    \textbf{\qwennav-8B}
      & 8.16 & \textbf{3.00} & \textbf{72.81} & \textbf{63.75} & \textbf{57.93}
      & 7.68 & \textbf{4.08} & \textbf{55.97} & \textbf{46.59} & \textbf{41.54} \\
    \bottomrule
  \end{tabular}}
\end{table}

\noindent \textbf{VLNVerse.} Table~\ref{tab:vlnverse} reports results on VLNVerse~\citep{vlnverse}, a large-scale benchmark that unifies previously fragmented VLN tasks into a single evaluation framework and replaces teleporting ``ghost'' agents with full-kinematics embodied locomotion driven by a physics engine. We evaluate under both fine-grained and coarse-grained instruction settings, thereby probing whether an agent can follow detailed step-by-step guidance as well as interpret high-level goal descriptions.
\qwennav-8B achieves \textbf{63.75\%} SR and \textbf{57.93\%} SPL on the fine-grained split, surpassing NavFoM by {12.2\%} SR and {25.5\%} SPL, and Uni-NaVid by {18.0\%} SR and {31.0\%} SPL.
On the more challenging coarse-grained split, where instructions omit procedural detail and require the agent to infer intermediate waypoints, \qwennav-8B reaches \textbf{46.59\%} SR and \textbf{41.54\%} SPL, outperforming NavFoM by {8.6\%} SR and {18.4\%} SPL.
Notably, the SPL margins are substantially larger than the SR margins in both settings, indicating that \qwennav not only reaches the goal more often but follows markedly more efficient paths.
The 4B variant also surpasses all baselines, confirming that the performance advantage is not merely a consequence of model scale.

\begin{table}[t]
  \centering
  \small
  \caption{\textbf{VLN-PE results with flash controller (R2R Val-Unseen)~\citep{wang2025vlnpe}.} TL: Trajectory Length; NE: Navigation Error; FR: Fall Rate; OS: Oracle Success Rate; SR: Success Rate; SPL: Success weighted by Path Length. \textbf{Bold}: best; \underline{underline}: second best.}
  \label{tab:vlnpe}
  \begin{tabular}{lcccccc}
    \toprule
    \textbf{Method} & TL & NE$\downarrow$ & FR$\downarrow$ & OS$\uparrow$ & SR$\uparrow$ & SPL$\uparrow$ \\
    \midrule
    Seq2Seq~\citep{krantz2020beyond}
      & 19.24 & 8.27 & \textbf{0.22} & 43.05 & 15.74 & 9.70 \\
    CMA~\citep{krantz2020beyond}
      & 40.21 & 31.24 & \textbf{0.22} & 45.06 & 20.94 & 14.06 \\
    RDP~\citep{wang2025vlnpe}
      & 15.12 & 6.98 & \underline{0.30} & 42.54 & 24.94 & 17.54 \\
    InternVLA-N1~\citep{internvlan1}
      & 10.11 & \underline{4.13} & 0.45 & 67.63 & \underline{60.36} & 54.93 \\
    \midrule
    \textbf{\qwennav-4B}
      & 10.39 & 4.24 & 3.83 & \underline{72.70} & 60.28 & \underline{55.24} \\
    \textbf{\qwennav-8B}
      & 9.17 & \textbf{3.73} & 4.05 & \textbf{72.99} & \textbf{65.50} & \textbf{61.19} \\
    \bottomrule
  \end{tabular}
\end{table}

\noindent \textbf{VLN-PE.} Table~\ref{tab:vlnpe} further evaluates on VLN-PE~\citep{wang2025vlnpe} using the flash controller, which decouples high-level navigation planning from low-level locomotion control and thereby provides a more direct assessment of spatial reasoning quality.
\qwennav-8B achieves \textbf{65.50\%} SR and \textbf{61.19\%} SPL, surpassing InternVLA-N1 by {5.1\%} SR and {6.3\%} SPL, while also attaining the lowest navigation error (\textbf{3.73}\,m) and the highest oracle success rate (\textbf{72.99\%}) among all methods.
\qwennav-4B performs comparably to InternVLA-N1 in SR ({60.28\%} vs.\ {60.36\%}) but achieves higher oracle success ({72.70\%} vs.\ {67.63\%}), suggesting stronger goal-proximate planning even at smaller model capacity.

\subsubsection{Object-Goal Navigation}

\begin{table}[t]
  \centering
  \small
  \caption{\textbf{Closed-vocabulary object-goal navigation on MP3D and HM3D.} $^\dagger$: uses depth or odometry. $^\ddagger$: results on HM3D v1 as reported in original papers; \qwennav is evaluated on the harder HM3D v2 benchmark. \textbf{Bold}: best; \underline{underline}: second best.}
  \label{tab:mp3d_hm3d}
  \setlength{\tabcolsep}{4pt}
  \begin{tabular}{lcccc}
    \toprule
    \multirow{2}{*}{\textbf{Method}} &
      \multicolumn{2}{c}{\textbf{MP3D}} &
      \multicolumn{2}{c}{\textbf{HM3D}} \\
    \cmidrule(lr){2-3} \cmidrule(lr){4-5}
    & SR$\uparrow$ & SPL$\uparrow$
    & SR$\uparrow$ & SPL$\uparrow$ \\
    \midrule
    VLFM$^{\dagger\ddagger}$~\citep{yokoyama2024vlfm}
      & 36.4 & 17.5
      & 52.5 & 30.4 \\
    OpenFMNav$^{\dagger\ddagger}$~\citep{kuang2024openfmnav}
      & 37.2 & 15.7
      & 52.5 & 24.1 \\
    SG-Nav$^{\dagger\ddagger}$~\citep{yin2024sgnav}
      & 40.2 & 16.0
      & 54.0 & 24.9 \\
    TriHelper$^{\dagger\ddagger}$~\citep{zhang2024trihelperzeroshotobjectnavigation}
      & -- & --
      & 56.5 & 25.3 \\
    WMNav$^{\dagger\ddagger}$~\citep{nie2025wmnav}
      & 45.4 & 17.2
      & 58.1 & 31.2 \\
    CogNav$^{\dagger\ddagger}$~\citep{cao2024cognav}
      & \underline{46.6} & 16.1
      & 72.5 & 26.2 \\
    Uni-NaVid$^\ddagger$~\citep{zhang2024uni}
      & -- & --
      & \underline{73.7} & \textbf{37.1} \\
    \midrule
    \textbf{\qwennav-4B}
      & \textbf{52.2} & 16.0
      & \textbf{75.6} & 30.6 \\
    \textbf{\qwennav-8B}
      & 48.8 & \textbf{17.7}
      & 71.2 & \underline{33.0} \\
    \bottomrule
  \end{tabular}
\end{table}

\begin{table}[t]
  \centering
  \small
  \caption{\textbf{Open-vocabulary object-goal navigation on HM3D-OVON~\citep{yokoyama2024hm3d}.} $^\dagger$: uses depth or odometry. \textbf{Bold}: best; \underline{underline}: second best.}
  \label{tab:ovon}
  \setlength{\tabcolsep}{4pt}
  \begin{tabular}{lcccccc}
    \toprule
    \multirow{2}{*}{\textbf{Method}} &
      \multicolumn{2}{c}{\textbf{Seen}} &
      \multicolumn{2}{c}{\textbf{Synonyms}} &
      \multicolumn{2}{c}{\textbf{Unseen}} \\
    \cmidrule(lr){2-3} \cmidrule(lr){4-5} \cmidrule(lr){6-7}
    & SR$\uparrow$ & SPL$\uparrow$
    & SR$\uparrow$ & SPL$\uparrow$
    & SR$\uparrow$ & SPL$\uparrow$ \\
    \midrule
    VLFM$^\dagger$~\citep{yokoyama2024vlfm}
      & 35.2 & 18.6
      & 32.4 & 17.3
      & 35.2 & 19.6 \\
    DAgRL+OD$^\dagger$~\citep{yokoyama2024hm3d}
      & 38.5 & 21.1
      & 39.0 & 21.4
      & 37.1 & 19.8 \\
    MTU3D$^\dagger$~\citep{zhu2025move}
      & 55.0 & 23.6
      & 45.0 & 14.7
      & 40.8 & 12.1 \\
    Uni-NaVid~\citep{zhang2024uni}
      & 41.3 & 21.1
      & 43.9 & 21.8
      & 39.5 & 19.8 \\
    NavFoM~\citep{zhang2025embodied}
      & 40.1 & 27.1
      & 45.4 & \underline{32.6}
      & 45.2 & \textbf{31.9} \\
    ABot-N0~\citep{abotN0}
      & 55.3 & \textbf{32.1}
      & 55.4 & \textbf{33.2}
      & \textbf{54.0} & \underline{30.5} \\
    \midrule
    \textbf{\qwennav-4B}
      & \textbf{57.7} & 24.4
      & \textbf{60.1} & 25.1
      & \underline{53.1} & 20.9 \\
    \textbf{\qwennav-8B}
      & \underline{56.1} & \underline{28.5}
      & \underline{57.8} & 28.8
      & 51.2 & 24.0 \\
    \bottomrule
  \end{tabular}
\end{table}

\noindent \textbf{MP3D \& HM3D.} Table~\ref{tab:mp3d_hm3d} reports closed-vocabulary ObjectNav results on MP3D and HM3D.
Notably, most prior methods in this space rely on depth sensors or odometry (marked with $^\dagger$), whereas \qwennav operates from RGB observations alone.
Since prior works only report results on HM3D v1 (Semantics v0.1, used in the Habitat 2022 challenge), we list their published numbers as baselines. \qwennav is evaluated on HM3D v2 (Semantics v0.2, used in the Habitat 2023 challenge), which features updated semantic annotations and a larger scene pool (216 vs.\ 120 scenes).
Despite this disadvantage, \qwennav-4B achieves \textbf{52.2\%} SR on MP3D, surpassing all depth-based baselines including CogNav ({46.6\%}) and WMNav ({45.4\%}).
On HM3D v2, \qwennav-4B reaches \textbf{75.6\%} SR with a distance-to-goal of only \textbf{1.72\,m}, surpassing even the v1 results of the vision-only Uni-NaVid ({73.7\%}) and establishing a new state of the art on this benchmark.
The 8B variant achieves the best SPL on MP3D (\textbf{17.7\%}) and competitive SPL on HM3D v2 ({33.0\%}), indicating more efficient paths at larger model scale.

\noindent \textbf{HM3D-OVON.} Table~\ref{tab:ovon} reports open-vocabulary results on HM3D-OVON, where the agent must locate objects described by free-form category names rather than a fixed label set.
\qwennav-4B achieves \textbf{57.7\%} / \textbf{60.1\%} / {53.1\%} SR on the Seen / Synonyms / Unseen splits, attaining the best success rate on two of three splits and ranking a close second on Unseen ({53.1\%} vs.\ ABot-N0's {54.0\%}).
Importantly, \qwennav uses only a single forward-facing camera in this evaluation, whereas ABot-N0 consumes panoramic multi-view observations that provide $360^\circ$ scene coverage and substantially reduce the need for active exploration—yet \qwennav still surpasses it on both Seen and Synonyms by {2.4\%} and {4.7\%} SR respectively.
The 8B variant also outperforms ABot-N0 on these two splits while achieving higher SPL ({28.5\%} / {28.8\%}), reflecting more efficient goal approach at larger model capacity.
The lower SPL of \qwennav relative to NavFoM and ABot-N0 reflects a reach-first exploration behaviour: the skeleton-based training trajectories encourage thorough room-by-room search, which improves goal-finding rate but at the cost of longer paths.

\subsubsection{Active Visual Tracking}

\begin{table}[t]
  \centering
  \small
  \caption{\textbf{Active visual tracking on EVT-Bench~\citep{wang2025trackvla} (Single Target split, single-view).} SR: Success Rate; TR: Tracking Rate; CR: Collision Rate. $^\dagger$: uses GroundingDINO; $^\ddagger$: uses SoM+GPT-4o. \textbf{Bold}: best; \underline{underline}: second best.}
  \label{tab:evt}
  \setlength{\tabcolsep}{6pt}
  \begin{tabular}{lccc}
    \toprule
    \textbf{Method} & TR$\uparrow$ & CR$\downarrow$ & SR$\uparrow$ \\
    \midrule
    IBVS$^\dagger$~\citep{gupta2016novel}      & 56.2 & 3.75 & 42.9 \\
    PoliFormer$^\dagger$~\citep{zeng2024poliformer} & 15.5 & 40.1 & 4.67 \\
    EVT~\citep{zhong2024empowering}            & 39.1 & 42.5 & 24.4 \\
    EVT$^\ddagger$~\citep{zhong2024empowering} & 49.9 & 40.5 & 32.5 \\
    Uni-NaVid~\citep{zhang2024uni}             & 39.5 & 41.9 & 25.7 \\
    TrackVLA~\citep{wang2025trackvla}          & 78.6 & \textbf{1.65} & 85.1 \\
    TrackVLA++~\citep{liu2025trackvlapp}       & 81.0 & \underline{2.10} & \underline{86.0} \\
    \midrule
    NavFoM~\citep{zhang2025embodied}                      & 80.5 & -- & 85.0 \\
    ABot-N0~\citep{abotN0}                     & 87.6 & 8.54 & \textbf{86.9} \\
    \midrule
    \textbf{\qwennav-4B}                       & \textbf{90.0} & 6.40 & 77.4 \\
    \textbf{\qwennav-8B}                       & \underline{89.7} & 5.70 & 78.6 \\
    \bottomrule
  \end{tabular}
\end{table}

Table~\ref{tab:evt} reports single-target tracking performance on EVT-Bench. \qwennav achieves the highest tracking rates among all methods (\textbf{90.0\%} TR for 4B and {89.7\%} TR for 8B), surpassing ABot-N0 ({87.6\%}) by {2.4\%}, NavFoM ({80.5\%}) by {9.5\%}, and the dedicated tracker TrackVLA++ ({81.0\%}) by {9.0\%}. \qwennav-8B also achieves the lowest collision rate among generalist models ({5.70\%} CR), substantially below ABot-N0 ({8.54\%}).
However, the success rate of \qwennav ({77.4\%} / {78.6\%}) trails that of ABot-N0 ({86.9\%}) and TrackVLA++ ({86.0\%}), which are specifically optimised for tracking tasks. We hypothesise that the broader multi-task training of \qwennav introduces a trade-off where the model maintains tighter following behaviour (superior TR) while being more conservative in declaring episode success.

\subsection{Embodied Question Answering}
\begin{table*}[t]
\centering
\small
\caption{\textbf{Performance comparison on embodied question answering benchmarks.} Results are compared against prior state-of-the-art methods on HM-EQA, MT-HM3D, and EXPRESS-Bench. $^*$: result reported by prior work. $^\dagger$: results are on the full A-EQA split. \textbf{Bold}: best; \underline{underline}: second best.}

\label{tab:eqa_performance_comparison}
\begin{tabular}{l cc cc cc}
\toprule
\multirow{2}{*}{\textbf{Method}} 
& \multicolumn{2}{c}{\textbf{HM-EQA}} 
& \multicolumn{2}{c}{\textbf{MT-EQA}} 
& \multicolumn{2}{c}{\textbf{EXPRESS-Bench}} \\
\cmidrule(lr){2-3} 
\cmidrule(lr){4-5} 
\cmidrule(lr){6-7}
& Acc. ($\uparrow$) 
& Steps ($\downarrow$) 
& Acc. ($\uparrow$) 
& Steps ($\downarrow$) 
& LLM Score ($\uparrow$) 
& $E_{\text{path}}$ ($\uparrow$) \\
\midrule
Explore-EQA~\citep{exploreeqa}  
& 58.4 & 0.52 & 36.2$^*$ & 0.64 & -- & -- \\
Graph-EQA~\citep{grapheqa} 
& 63.5 & 0.20 & 45.6$^*$ & 0.45 & -- & -- \\
Memory-EQA~\citep{memoryeqa} 
& 61.4 & 0.40 & 43.1 & 0.41 & -- & -- \\
Fine-EQA~\citep{fineeqa} 
& 56.0 & 0.54 & -- & -- & 63.95 & 25.58 \\
3D-Mem~\citep{3dmem} 
& 50.4 & 0.63 & -- & -- & -- & -- \\
FAST-EQA~\citep{fasteqa} 
& 69.2 & 0.65 & 50.5 & 0.52 & 68.7 & 29.25 \\
\midrule
 Qwen3.5-Plus+QwenNav-8B & \underline{74.1}
& \underline{0.17}
& \underline{52.1}
& \underline{0.22}
& \underline{77.66}
& \underline{31.73} \\
Qwen3.6-Plus+QwenNav-8B 
& \textbf{76.7} 
& \textbf{0.15} 
& \textbf{54.4} 
& \textbf{0.19} 
& \textbf{79.27} 
& \textbf{33.96} \\
\bottomrule
\end{tabular}
\end{table*}

Table~\ref{tab:eqa_performance_comparison} summarises embodied question answering results on HM-EQA, MT-HM3D, and EXPRESS-Bench. Our \qwennav-based variants outperform prior methods across all three benchmarks. In particular, \textbf{Qwen3.6-Plus+\qwennav} achieves the best overall performance, reaching \textbf{76.7} SR on HM-EQA, \textbf{54.4} SR on MT-HM3D, and \textbf{79.27} LLM Score on EXPRESS-Bench, consistently surpassing recent strong baselines such as FAST-EQA and Memory-EQA. Compared with FAST-EQA, Qwen3.6-Plus+\qwennav improves absolute performance by {7.5} points on HM-EQA, {3.9} points on MT-HM3D, and {10.57} points on EXPRESS-Bench. It also reduces normalized
equivalent steps on HM-EQA and MT-EQA and improves
EXPRESS-Bench $E_{path}$ by 4.71 points, indicating that the
learned executor supports more targeted physical exploration. These results suggest that the multimodal reasoning and navigation capabilities of \qwennav transfer effectively to embodied question answering, enabling both stronger scene exploration and more accurate answer prediction.
Furthermore, the substantial improvement in navigation efficiency achieved by \qwennav enables our system to reach target locations using significantly fewer equivalent steps. This further demonstrates its viability as an effective physical-world tool for general-purpose multimodal agents.

\subsection{Autonomous Driving}

\begin{table}[t]
  \centering
  \small
  \caption{\textbf{Performance comparison on NAVSIM navtest using closed-loop metrics.} NC: Navigation Compliance; DAC: Drivable Area Compliance; TTC: Time-to-Collision; Comf.: Comfort; EP: Ego Progress; PDMS: PDM Score. $^\dagger$: uses additional LiDAR information. $^\ddag$: our model evaluated without historical ego-status prior. \textbf{Bold}: best; \underline{underline}: second best.}
  \label{tab:navsim}
  \setlength{\tabcolsep}{5pt}
  \begin{tabular}{lcccccc}
    \toprule
    \textbf{Method} & NC$\uparrow$ & DAC$\uparrow$ & TTC$\uparrow$ & Comf.$\uparrow$ & EP$\uparrow$ & PDMS$\uparrow$ \\
    \midrule
    Human                         & 100  & 100  & 100  & 99.9 & 87.5 & 94.8 \\
    Constant Velocity             & 68.0 & 57.8 & 50.0 & 100  & 19.4 & 20.6 \\
    Ego Status MLP                & 93.0 & 77.3 & 83.6 & 100  & 62.8 & 65.6 \\
    \midrule
    UniAD~\citep{hu2023planning}                         & 97.8 & 91.9 & 92.9 & 100  & 78.8 & 83.4 \\
    TransFuser$^\dagger$~\citep{chitta2022transfuser}    & 97.7 & 92.8 & 92.8 & 100  & 79.2 & 84.0 \\
    PARA-Drive~\citep{weng2024drive}                     & 97.9 & 92.4 & 93.0 & 99.8 & 79.3 & 84.0 \\
    LAW~\citep{li2024enhancing}                          & 96.4 & 95.4 & 88.7 & 99.9 & 81.7 & 84.6 \\
    DRAMA$^\dagger$~\citep{yuan2024drama}                & 98.0 & 93.1 & 94.8 & 100  & 80.1 & 85.5 \\
    Hydra-MDP++~\citep{li2025hydra}                      & 97.6 & 96.0 & 93.1 & 100  & 80.4 & 86.6 \\
    DiffusionDrive$^\dagger$                             & 98.2 & 96.2 & 94.7 & 100  & 82.2 & 88.1 \\
    WoTE$^\dagger$~\citep{li2025end}                     & 98.5 & 96.8 & 94.9 & 99.9 & 81.9 & 88.3 \\
    Hydra-NeXt$^\dagger$~\citep{li2025hydranext}         & 98.1 & 97.7 & 94.6 & 100  & 81.8 & 88.6 \\
    VADv2~\citep{jiang2024vadv2}                         & 98.3 & 97.4 & 95.7 & 100  & 82.3 & 89.3 \\
    GoalFlow$^\dagger$~\citep{xing2025goalflow}          & 98.4 & \underline{98.3} & 94.6 & 100  & 85.0 & 90.3 \\
    \midrule
    DrivingGPT~\citep{chen2025drivinggpt}                & \underline{98.9} & 90.7 & 94.9 & 95.6 & 79.7 & 82.4 \\
    NavFoM~\citep{zhang2025embodied}                     & 97.7 & 93.5 & 92.3 & 100  & 79.6 & 84.3 \\
    AutoVLA~\citep{zhou2025autovla}                      & 98.4 & 95.6 & 98.0 & 99.9 & 81.9 & 89.1 \\
    ReCogDrive~\citep{li2025recogdrive}                  & 97.9 & 97.3 & 94.9 & 100  & \textbf{87.3} & 90.8 \\
    ReflectDrive~\citep{li2026discrete}                  & 97.7 & \textbf{99.3} & 93.5 & 100  & \underline{86.9} & \underline{91.1} \\
    \midrule
    \textbf{\qwennav-4B}$^\ddag$   & 96.4 & 90.9 & 89.0 & 99.9 & 75.2 & 79.5 \\
    \textbf{\qwennav-8B}$^\ddag$   & 95.9 & 91.3 & 88.4 & 100  & 75.5 & 79.5 \\
    \textbf{\qwennav-4B}           & \textbf{99.8} & 97.5 & \textbf{98.5} & 99.9 & 84.4 & \textbf{91.4} \\
    \textbf{\qwennav-8B}           & \textbf{99.8} & 96.9 & \underline{98.2} & 99.9 & 84.2 & 90.9 \\
    \bottomrule
  \end{tabular}
\end{table}

\begin{figure*}[t]
    \centering
    \includegraphics[width=0.95\linewidth]{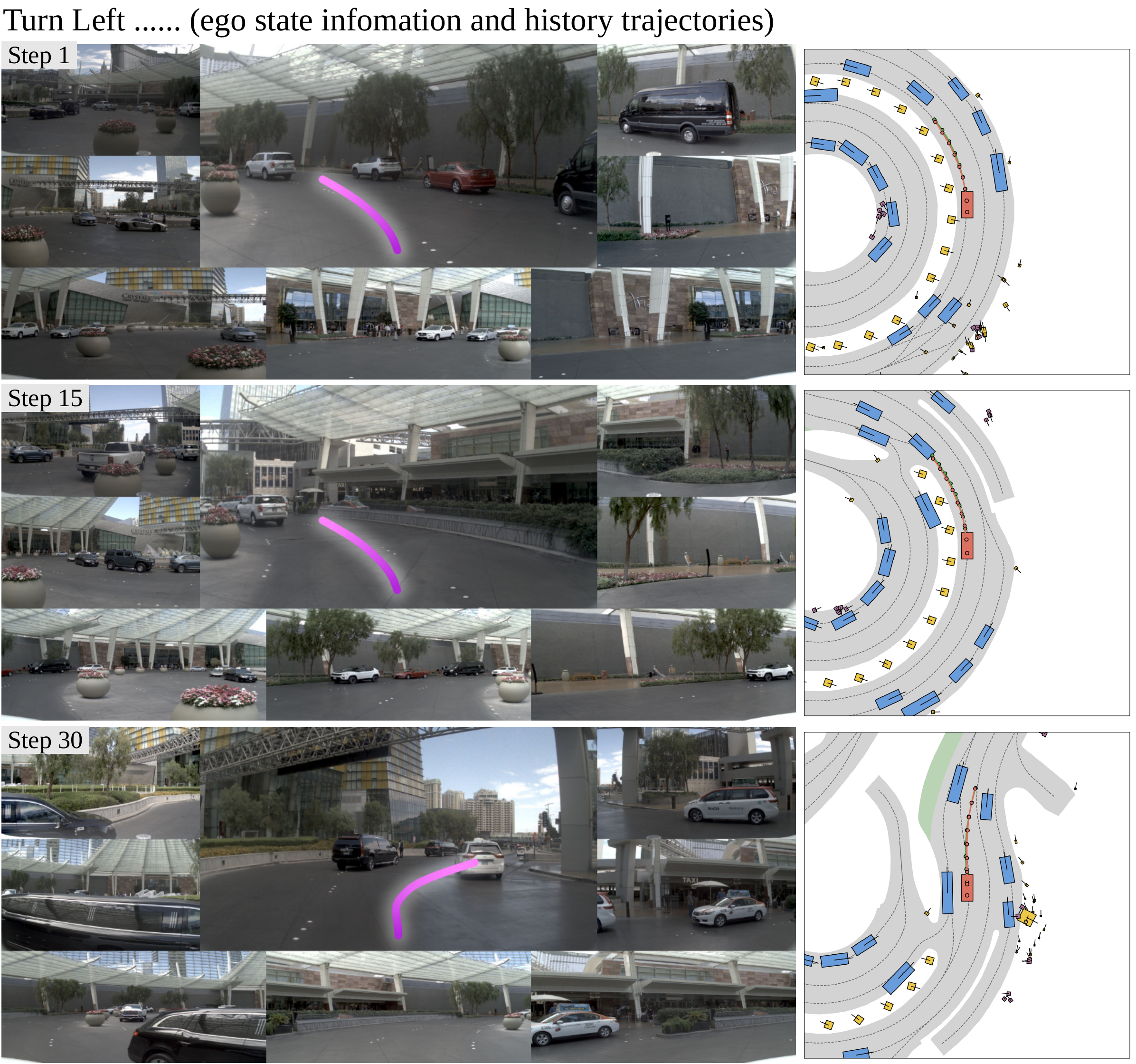}
    \caption{\textbf{Qualitative closed-loop planning visualization on NAVSIM.} We visualize a representative left-turn case in an annular road scene. For each timestep, the figure shows the multi-view camera observations, the predicted future trajectory overlaid on the front-view image, and the corresponding BEV scene. \qwennav produces temporally consistent curved trajectories from Step~1 to Step~30, progressively completing the left turn while remaining aligned with the drivable lane.}
    \label{fig:navsim_vis}
\end{figure*}

\noindent \textbf{NAVSIM.}
Table~\ref{tab:navsim} reports trajectory planning results with closed-loop metrics on NAVSIM navtest. During evaluation, we provide the ground-truth trajectories of the previous three frames in the prompt as historical priors, allowing the model to condition its prediction on short-term ego-motion context. \qwennav achieves strong performance among vision-language driving models. In particular, \qwennav-4B reaches \textbf{91.4} PDMS, outperforming NavFoM by 7.1 points, AutoVLA by 2.3 points, ReCogDrive by 0.6 points, and ReflectDrive by 0.3 points. It also achieves highly competitive safety-related scores, with \textbf{99.8} NC and \textbf{98.5} TTC, while maintaining strong DAC and EP performance. \qwennav-8B obtains similarly strong results, reaching 90.9 PDMS with \textbf{99.8} NC and \underline{98.2} TTC. Compared with the variants evaluated without historical ego-status priors, both models obtain a substantial gain of more than 11 points in PDMS, highlighting the importance of short-term trajectory history for closed-loop driving prediction.

As shown in \Cref{fig:navsim_vis}, the qualitative NAVSIM visualization further illustrates the closed-loop behavior of \qwennav in a left-turn scenario. The model leverages multi-view observations together with ego-state and historical trajectory information to predict a smooth turning trajectory. Across different timesteps, the predicted path adapts to the changing scene context, follows the curvature of the annular road, and gradually transitions from entering the turn to aligning with the outgoing lane. This demonstrates that \qwennav can maintain temporally coherent planning behavior under multi-view driving observations.

\begin{table}[t]
  \centering
  \small
\caption{\textbf{AlpaSim closed-loop evaluation on the PhysicalAI-AV NuRec dataset.} Results are evaluated on 920 scenarios using at-fault closed-loop metrics, where close encounters are counted only when the ego vehicle is deemed responsible, excluding rear-end close encounters. \textbf{Bold}: best; \underline{underline}: second best.}
  \label{tab:alpasim}
  \setlength{\tabcolsep}{6pt}
  \begin{tabular}{lccc}
    \toprule
    \textbf{Method} & Close Encounter Rate$\downarrow$ (\%) & Off-Road Rate$\downarrow$ (\%) & AlpaSim Score$\uparrow$ \\
    \midrule
    Alpamayo-R1-0.5B      & \underline{9.0} & \underline{19.0} & \underline{0.35} \\
    Alpamayo-R1-10B       & \textbf{4.0} & \textbf{16.0} & \textbf{0.72} \\
    \midrule
    \textbf{\qwennav-4B}                     & 22.0 & 34.0 & 0.15 \\
    \textbf{\qwennav-8B}                     & 22.0 & 27.0 & 0.17 \\
    \bottomrule
  \end{tabular}
\end{table}

\begin{figure*}[t]
\centering
\includegraphics[width=0.95\linewidth]{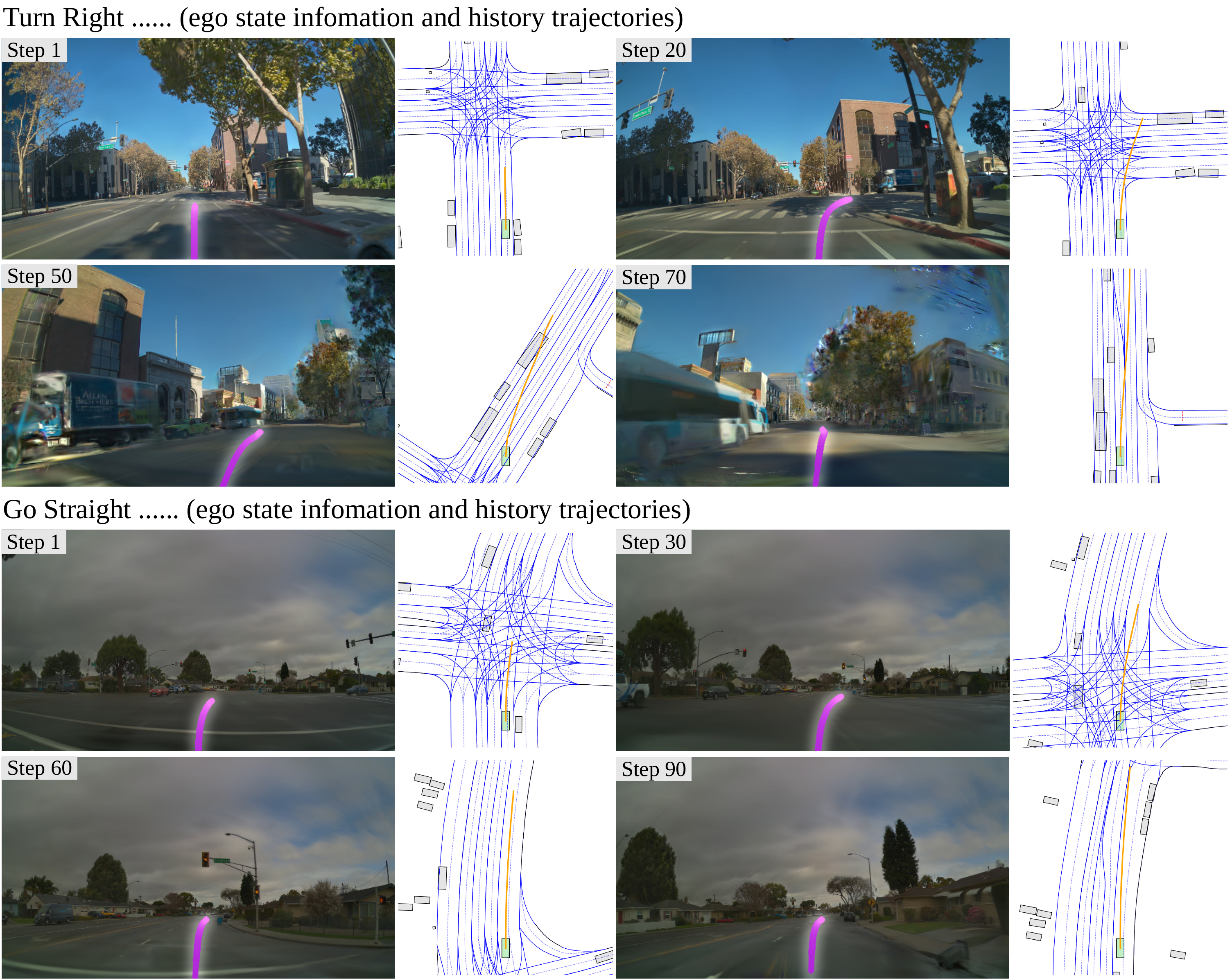}
\caption{\textbf{Qualitative zero-shot closed-loop simulation visualization on AlpaSim.} We show two representative cases from the PhysicalAI-AV NuRec dataset. In the first right-turn case, \qwennav slows down and proceeds straight after approaching an intersection, then performs a right turn while keeping away from the road boundary, and finally continues forward along the outgoing lane. In the second straight-driving case, the ego vehicle starts from a stopped state after the traffic light turns green and safely passes through a complex intersection while avoiding surrounding traffic participants.}
\label{fig:alpasim_vis}
\end{figure*}

\noindent \textbf{AlpaSim.}
Table~\ref{tab:alpasim} presents closed-loop evaluation on the PhysicalAI-AV NuRec dataset using AlpaSim. This benchmark evaluates driving performance with at-fault metrics, where close encounters are counted only when the ego vehicle is deemed responsible. We include AlpaSim as an additional autonomous driving benchmark to examine closed-loop safety and long-horizon driving robustness beyond NAVSIM.

Importantly, \qwennav is evaluated in a zero-shot setting on AlpaSim: the model is directly rolled out in the simulator without AlpaSim-specific training or closed-loop adaptation on the PhysicalAI-AV NuRec scenarios. Therefore, this benchmark mainly measures the out-of-domain transfer ability of a general vision-language navigation model under long-horizon autonomous driving simulation. Although \qwennav still lags behind the Alpamayo-R1 models that are designed for this evaluation setting, it achieves non-trivial zero-shot closed-loop performance. Moreover, increasing the model scale from 4B to 8B improves the off-road rate from 34.0\% to 27.0\% and the AlpaSim score from 0.15 to 0.17, suggesting that larger backbones provide better closed-loop stability and scene-level generalization under this challenging transfer setting.

\Cref{fig:alpasim_vis} provides qualitative examples of \qwennav in zero-shot AlpaSim closed-loop simulation. In the right-turn scenario, the ego vehicle first approaches the intersection cautiously and proceeds straight at a reduced speed, then executes a right turn while avoiding the road boundary, and finally aligns with the outgoing lane to continue driving forward. In the intersection scenario, the vehicle correctly reacts to the green traffic light, starts from a stopped state, and proceeds through a complex multi-agent intersection without colliding with other vehicles. These cases suggest that, despite the remaining quantitative gap on aggregate AlpaSim metrics, \qwennav can still exhibit meaningful reactive planning behaviors when transferred zero-shot to challenging closed-loop driving scenes.

\subsection{Ablation Study}
\label{sec:ablation}

\begin{figure}[t]
    \centering
    \includegraphics[width=0.7\linewidth]{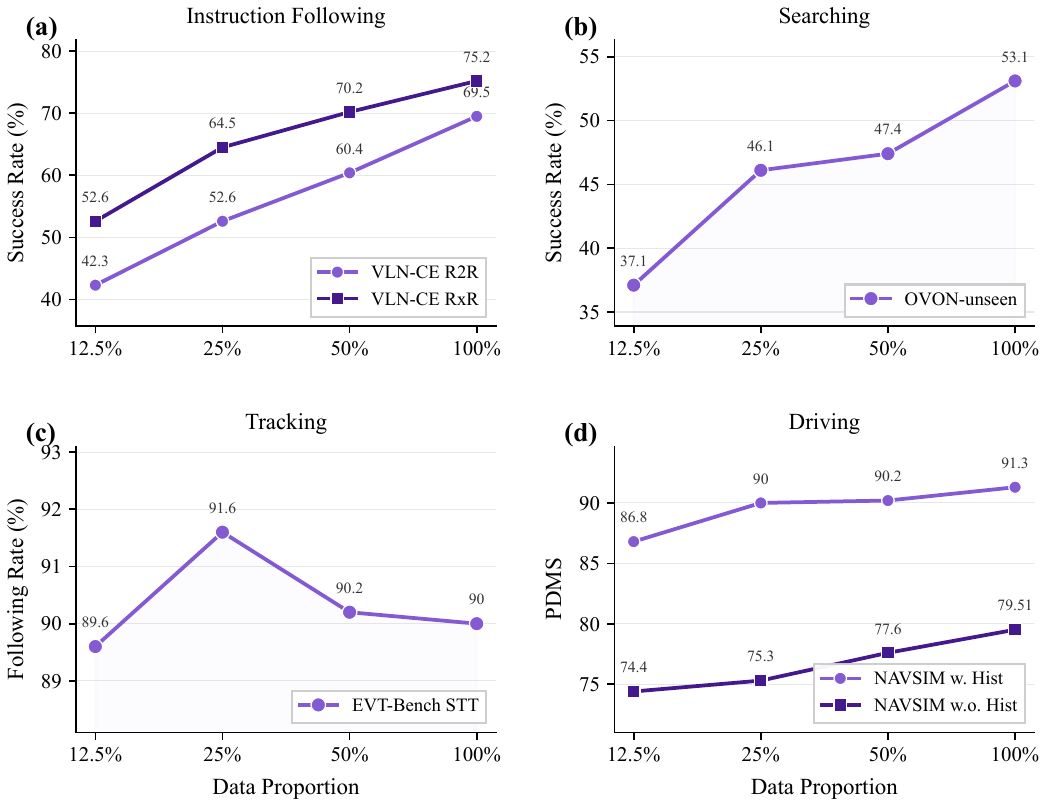}
    \caption{\textbf{Data scaling behavior of \qwennav.} Performance on representative navigation benchmarks as a function of the training data fraction. Increasing the amount of navigation training data yields clear gains on most tasks, with especially strong improvements on long-horizon tasks such as VLN-CE RxR, while short-horizon tracking saturates earlier and exhibits mild non-monotonicity.}
    \label{fig:scaling}
\end{figure}

\noindent \textbf{Effect of training data scale.}
\Cref{fig:scaling} shows how \qwennav performance scales with the fraction of navigation training data used during training. Across representative benchmarks, increasing the data fraction from 12.5\% to 100\% leads to clear overall gains, most notably on instruction-following and driving tasks. The gains are particularly pronounced on long-horizon instruction-following tasks such as VLN-CE RxR, suggesting that broader trajectory coverage improves grounding between linguistic instructions and extended visual histories. On shorter-horizon tasks such as target tracking (EVT-Bench), performance improves rapidly with moderate data and then saturates with mild fluctuations, indicating that these tasks are less sensitive to additional training data once basic tracking behavior has been learned. Importantly, the full-data model remains competitive while preserving the stronger gains obtained on long-horizon and cross-embodiment tasks.

\begin{figure}[t]
    \centering
    \includegraphics[width=\linewidth]{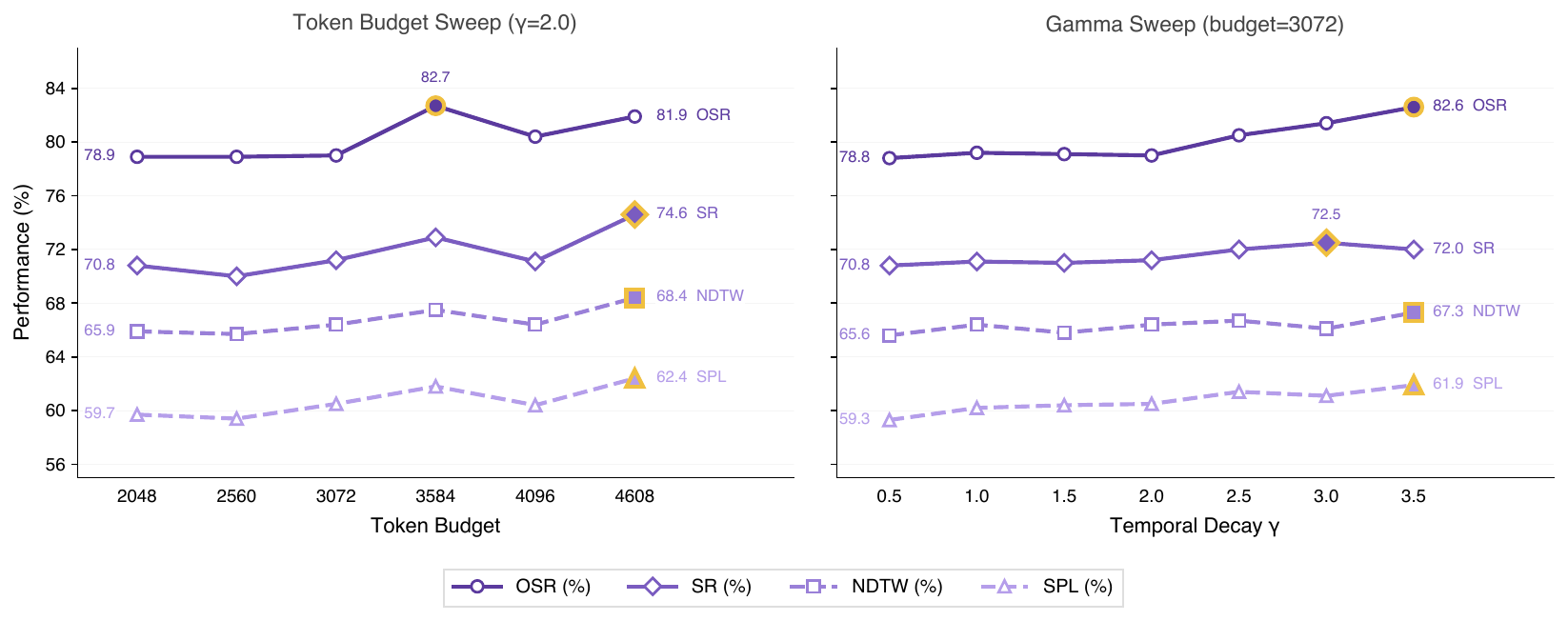}
    \caption{\textbf{Ablation on token budget $B$ and temporal decay $\gamma$.} We evaluate \qwennav-4B on 500 VLN-CE R2R Val-Unseen episodes under varying configurations. \textit{Left:} Sweeping the token budget from 2048 to 4608 at fixed $\gamma{=}2.0$. \textit{Right:} Sweeping the temporal decay from 0.5 to 3.5 at fixed $B{=}3072$.}
    \label{fig:ablation_token_gamma}
\end{figure}

\noindent \textbf{Effect of token budget and temporal decay.}
\Cref{fig:ablation_token_gamma} examines the two primary control parameters of the task-adaptive observation encoding: the total visual token budget $B$ and the temporal decay factor $\gamma$. We evaluate \qwennav-4B on 500 VLN-CE R2R Val-Unseen episodes, reporting SR, SPL, nDTW, and OSR.
In the token budget sweep (left panel, $\gamma{=}2.0$), increasing $B$ substantially improves performance over the low-budget setting, though the benefit is not strictly monotonic once the budget becomes large. SR improves from 70.8\% at $B{=}2048$ to 74.6\% at $B{=}4608$, while OSR rises from 78.9\% and peaks at 82.7\% when $B{=}3584$ before slightly decreasing at the largest budget. This pattern indicates that retaining more visual tokens generally provides richer spatial context and improves goal-reaching ability, but excessive or poorly allocated visual context can yield diminishing returns.
In the gamma sweep (right panel, $B{=}3072$), OSR shows a clear overall improvement from 78.8\% ($\gamma{=}0.5$) to 82.6\% ($\gamma{=}3.5$), while SR peaks at 72.5\% ($\gamma{=}3.0$) before slightly declining. Larger $\gamma$ concentrates more tokens on recent frames, enhancing the model's ability to resolve the current scene at the expense of early-history context. For instruction-following tasks such as VLN-CE, where the agent must react to its immediate surroundings while still respecting earlier landmarks, this recency bias is beneficial but exhibits a trade-off: oracle success continues to improve at high $\gamma$, whereas strict success and path-quality metrics saturate or fluctuate slightly.

\subsection{Real-World Deployment Results}
\label{sec:real_world}

\begin{figure*}[t]
  \centering
  \includegraphics[width=0.95\linewidth]{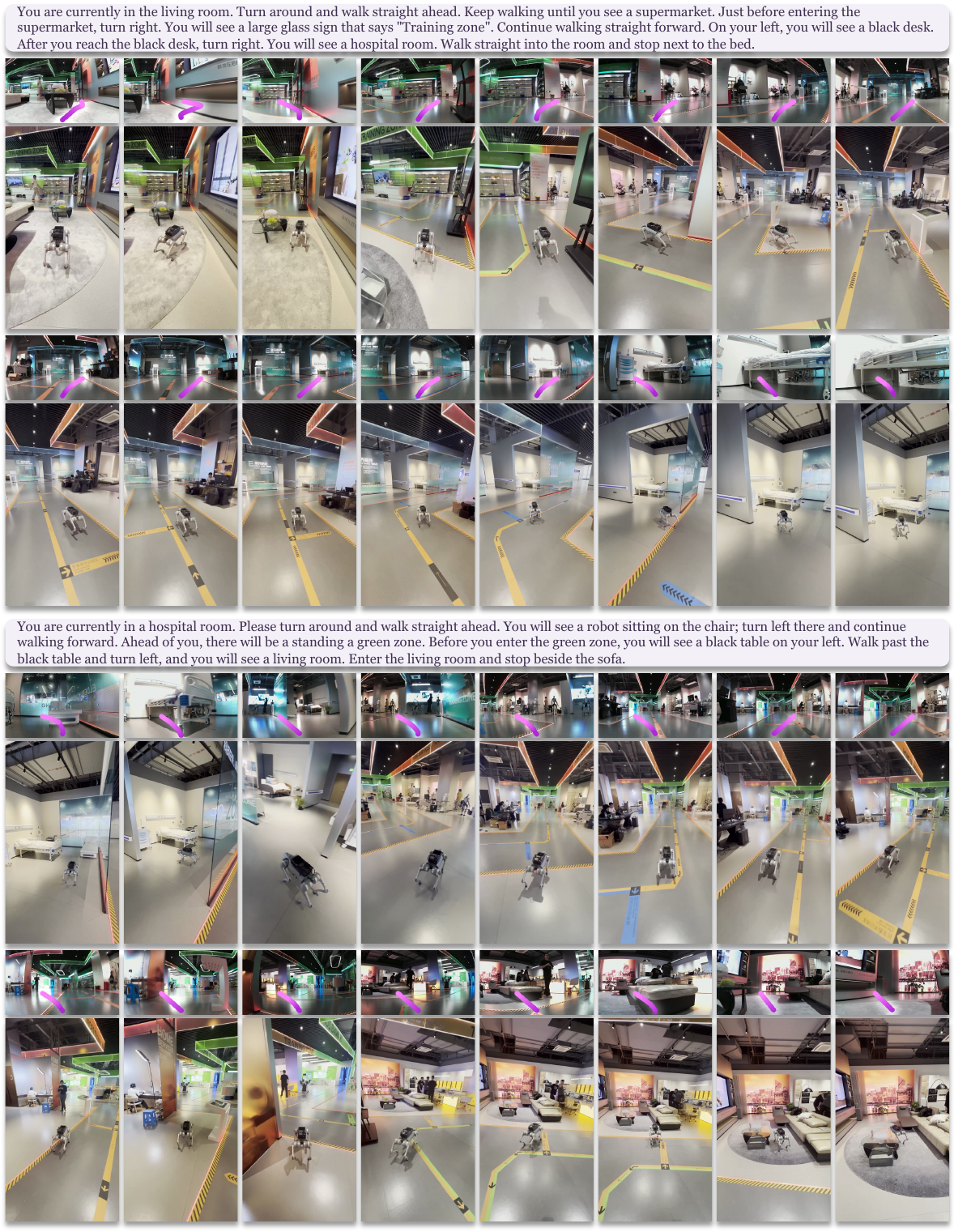}
  \caption{\textbf{Real-world VLN deployment in an unseen exhibition hall.} The robot dog navigates 21.78\,m from a living room to a medical room following pure language instructions, leveraging different visual landmarks along the route. Upon receiving a reverse language command, the robot precisely walks backward to its original starting position.}
  \label{fig:deployment_vln}
\end{figure*}

\begin{figure*}[t]
  \centering
  \includegraphics[width=0.95\linewidth]{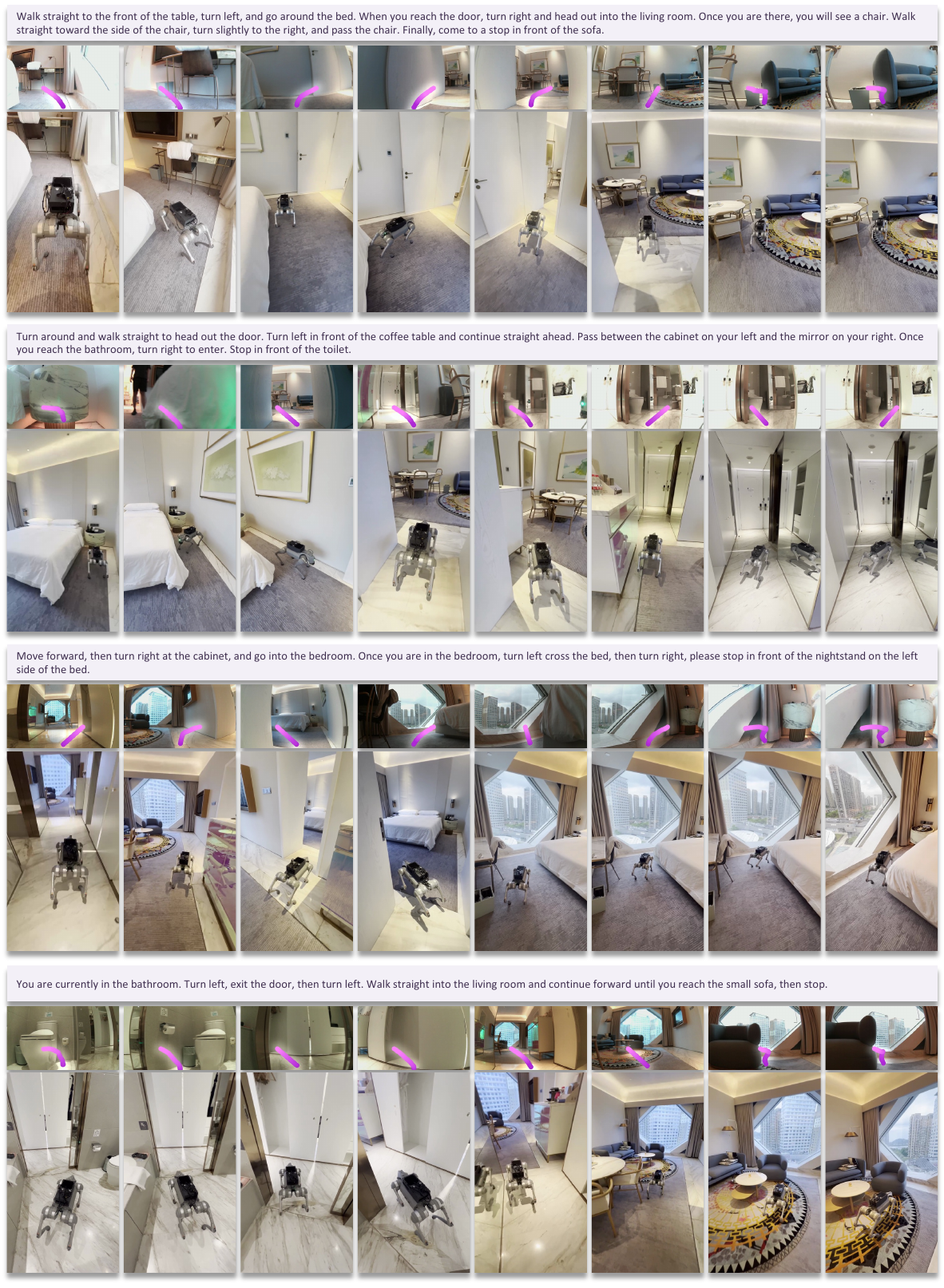}
  \caption{\textbf{Indoor deployment with verbal commands.} The robot executes navigation tasks in an apartment setting using step-by-step verbal instructions, traversing between the bedroom, living room, and bathroom while responding to fine-grained spatial directives.}
  \label{fig:deployment_vln_indoor}
\end{figure*}

To further validate the practical applicability of \qwennav, we deploy it on a quadruped robot in a previously unseen, real-world exhibition hall. This environment is highly out-of-distribution, as the hall encompasses various types of environments such as living rooms, medical rooms, and open corridors, presenting a rich mixture of heterogeneous scenes and objects that are absent from the training distribution.

As shown in \Cref{fig:deployment_vln}, we evaluate a vision-and-language navigation task using \emph{pure language instructions} without any goal images or coordinates. The robot is commanded to navigate from a living room area to a medical room located 21.78\,m away. Throughout the trajectory, the model leverages different visual landmarks, such as furniture, doorways, and signage, to ground language instructions into spatial decisions, successfully traversing multiple visually distinct zones to reach the designated target area.

We further demonstrate fine-grained language-based control by issuing a \emph{reverse language command} that instructs the robot to walk backward. Upon receiving this command, the model switches its locomotion mode and accurately executes the reverse trajectory, retracing the entire route until it returns to a position closely matching its initial starting pose. This result highlights three key capabilities: (1)~\qwennav can faithfully interpret and execute diverse motion primitives, including non-standard behaviors such as backward locomotion, purely from natural language; (2)~the model effectively utilizes visual landmarks encountered during forward navigation to maintain spatial awareness along the return path; and (3)~when directed to a previously visited location, the model can accurately return to that exact position, demonstrating precise spatial grounding in a cluttered, previously unseen environment.

We also evaluate \qwennav in an indoor apartment setting, as shown in \Cref{fig:deployment_vln_indoor}. The key finding is that the model's behavior can be precisely controlled through natural language commands. For instance, when instructed to stop at the nightstand on the left side of the bed, the robot navigates into the bedroom and halts at the exact specified side; when instructed to turn around before exiting to the living room, the robot faithfully completes the detour rather than taking a direct path. In the four representative examples we showcase, the robot consistently translates fine-grained verbal directives into accurate navigation behaviors across multiple rooms, confirming that \qwennav provides reliable and precise language-based control in complex indoor environments.
\begin{figure*}[t]
  \centering
  \includegraphics[width=0.98\linewidth]{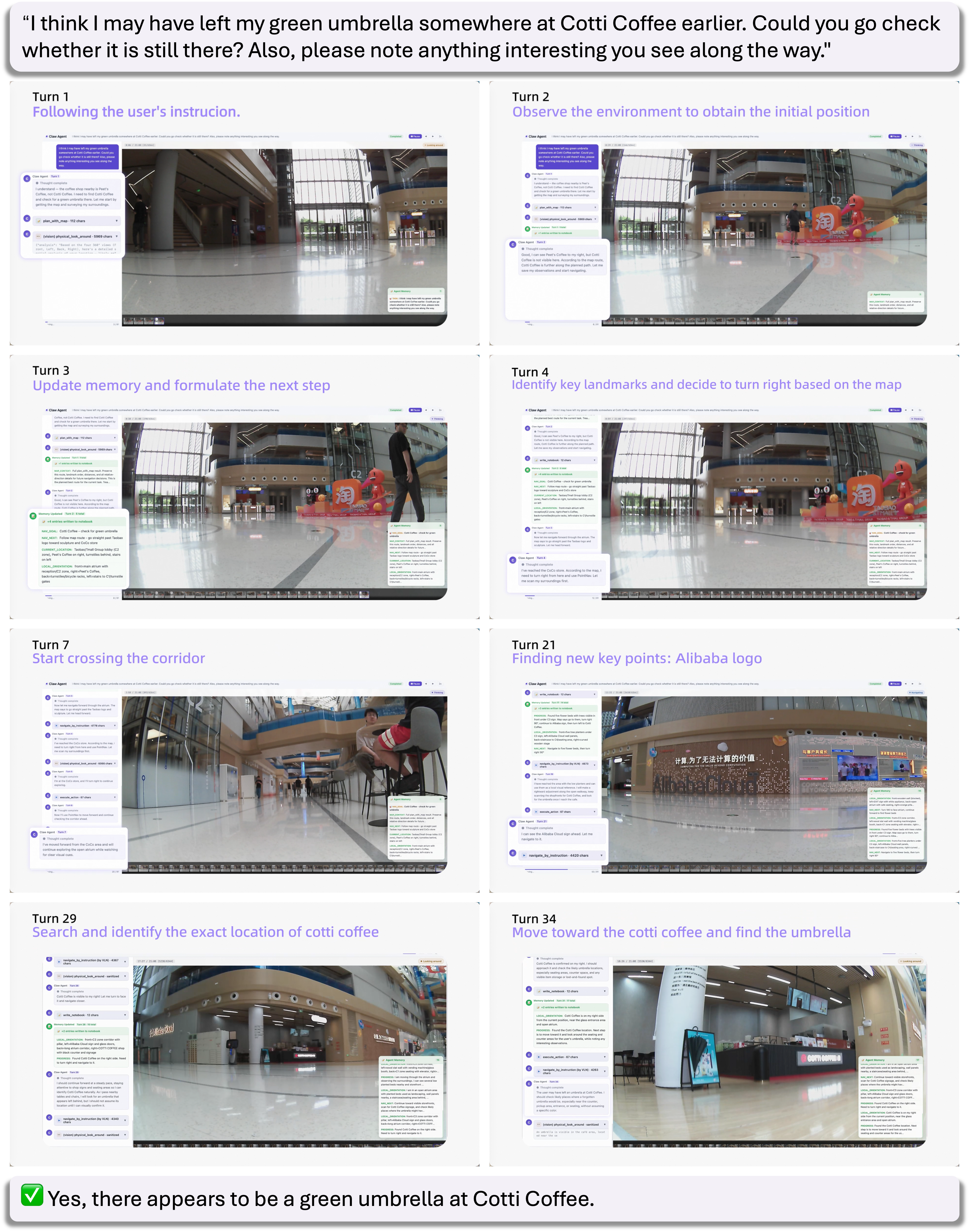}
\caption{\textbf{Real-world long-horizon navigation with agentic \qwennav.}
On a real robot, the agent answers an open-ended request by decomposing the task into sub-goals, following landmarks to Cotti Coffee, and verifying the green umbrella from visual evidence.
Selected turns show the loop of upper-level planning, \qwennav execution, memory updates, and final response generation.
}
  \label{fig:deployment_vl_agent}
\end{figure*}
We further evaluate the agentic navigation interface in a real indoor environment with a mobile robot.
As shown in \Cref{fig:deployment_vl_agent}, the user gives an open-ended instruction: the robot should check whether a green umbrella was left at Cotti Coffee and also report salient observations along the way.
Unlike a short route-following command, this task requires the system to infer the destination from language, maintain task progress over many decision turns, use visual landmarks for localization, and update its plan as new evidence is collected.
The selected turns in \Cref{fig:deployment_vl_agent} summarize the high-level agent loop rather than every low-level control step.
The agent first parses the user request and observes the initial scene to establish its starting position.
It then updates its memory, identifies useful landmarks, and decides to follow the corridor toward the likely store area.
During navigation, the agent records intermediate cues such as the Alibaba logo, uses them to refine its localization, and continues searching for the exact Cotti Coffee location.
After reaching the store area, the robot inspects the scene and observes a green umbrella, allowing the agent to produce the final response that the umbrella appears to still be there.

This example illustrates the intended division of labor in the agentic \qwennav system.
The upper-level agent decomposes the open-ended user request into successive sub-goals and maintains an evidence notebook across turns, while \qwennav executes grounded navigation segments and returns trajectory evidence for subsequent planning.
The real-world episode shows that the same interface can support long-horizon navigation, landmark-based reasoning, memory update, and final evidence-grounded response generation outside simulated benchmark settings.

\section{Conclusion}

We have presented \qwennav, a unified navigation model built on \qwenvl that reframes the central challenge of multi-task navigation as a \emph{context modeling} problem rather than an architecture or task-head problem.
Our key insight is that diverse navigation tasks, from instruction following to target tracking to autonomous driving, share the same perception and planning backbone but demand fundamentally different strategies for consuming the observation stream.
Rather than committing to a single fixed strategy, \qwennav exposes a parameterised observation encoding interface that allows an outer agent or human operator to dynamically select the appropriate context modeling strategy for each sub-task at inference time, with training-time randomisation ensuring robustness to any configuration without retraining.

This design principle yields two broader implications.
First, it transforms a navigation model from a fixed policy into a \emph{reconfigurable navigation primitive}: the same model seamlessly switches between global-history mode for exploration and recency-focused mode for precise local manoeuvring, enabling complex long-horizon behaviours to emerge from the composition of simple, well-defined working modes.
Second, natural-language viewpoint and temporal identification demonstrate that architectural simplicity can be a strength: by communicating structure through ordinary vocabulary tokens rather than learned positional embeddings, the model preserves its open-world language grounding at zero additional cost.

Extensive experiments across navigation, tracking, autonomous driving, and embodied question answering validate these principles, and zero-shot transfer to real-world robots confirms that the unified design generalises beyond simulation.
We believe the task-adaptive observation encoding introduced here, treating observation context as a first-class, externally controllable degree of freedom, offers a promising direction for building navigation foundation models that are both broadly capable and practically deployable within agentic navigation systems.

\section{Contributions and Acknowledgments}

\bgroup
\renewcommand{\thefootnote}{\fnsymbol{footnote}}
\setcounter{footnote}{0} 

\textbf{Core Contributors:} 
Jiazhao Zhang\footnote{Equal contribution.}\footnote{Project lead.}, 
Gengze Zhou\footnotemark[1]\footnotemark[2], 
Hale Yin\footnotemark[1], 
Yiyang Huang\footnotemark[1], 
Zixing Lei\footnotemark[1], 
Qihang Peng\footnotemark[1], 
Haoqi Yuan, 
Jie Zhang, 
Xudong Guo,
Xiaoyue Chen,
An Yang, 
Fei Huang, 
Zhibo Yang,
Junyang Lin, 
Dayiheng Liu, 
Jingren Zhou, 
Chenxu Lv\footnote{Corresponding author.}, 
Chenfei Wu\footnotemark[3], 
Xiong-Hui Chen\footnotemark[2]\footnotemark[3]

\vspace{0.5em}
\textbf{Contributors:} Zhuoyuan Yu, Jingyang Fan, Zhixuan Liang, Pei Lin, Ye Wang, Haoyang Li, Anzhe Chen, Kun Yan, Xiao Xu, Jiahao Li, Lulu Hu, Minying Zhang, Shurui Li, Wenhu Xiao, Shuai Bai, Xuancheng Ren

\textbf{Acknowledgments:} We acknowledge the National Pilot Base for Embodied Intelligence Applications for providing the real-robot experimental environment and equipment. We appreciate that Unitree provides their Unitree Go 2 for our experiments.

\egroup

\bibliography{colm2024_conference}

@inproceedings{goyal2017making,
  title={Making the v in vqa matter: Elevating the role of image understanding in visual question answering},
  author={Goyal, Yash and Khot, Tejas and Summers-Stay, Douglas and Batra, Dhruv and Parikh, Devi},
  booktitle={CVPR},
  year={2017}
}

@inproceedings{lin2014microsoft,
  title={Microsoft coco: Common objects in context},
  author={Lin, Tsung-Yi and Maire, Michael and Belongie, Serge and Hays, James and Perona, Pietro and Ramanan, Deva and Doll{\'a}r, Piotr and Zitnick, C Lawrence},
  booktitle={ECCV},
  year={2014},
}

@inproceedings{dvqa,
  title={Dvqa: Understanding data visualizations via question answering},
  author={Kafle, Kushal and Price, Brian and Cohen, Scott and Kanan, Christopher},
  booktitle={CVPR},
  year={2018}
}

@article{figureqa,
  title={FigureQA: An Annotated Figure Dataset for Visual Reasoning},
  author={Kahou, Samira Ebrahimi and Michalski, Vincent and Atkinson, Adam and K{\'a}d{\'a}r, {\'A}kos and Trischler, Adam and Bengio, Yoshua},
  journal={arXiv preprint arXiv:1710.07300},
  year={2018}
}

@inproceedings{johnson2017clevr,
  title={CLEVR: A Diagnostic Dataset for Compositional Language and Elementary Visual Reasoning},
  author={Johnson, Justin and Hariharan, Bharath and van der Maaten, Laurens and Fei-Fei, Li and Zitnick, C Lawrence and Girshick, Ross},
  booktitle={CVPR},
  year={2017}
}

@misc{qwen3,
      title={Qwen3 Technical Report}, 
      author={An Yang and Anfeng Li and Baosong Yang and Beichen Zhang and Binyuan Hui and Bo Zheng and Bowen Yu and others},
      journal = {arXiv:2505.09388},
      year={2025},
}

@inproceedings{anderson2018vision,
  title={Vision-and-language navigation: Interpreting visually-grounded navigation instructions in real environments},
  author={Anderson, Peter and Wu, Qi and Teney, Damien and Bruce, Jake and Johnson, Mark and S{\"u}nderhauf, Niko and Reid, Ian and Gould, Stephen and Van Den Hengel, Anton},
  booktitle={CVPR},
  year={2018}
}

@inproceedings{qi2020reverie,
  title={Reverie: Remote embodied visual referring expression in real indoor environments},
  author={Qi, Yuankai and Wu, Qi and Anderson, Peter and Wang, Xin and Wang, William Yang and Shen, Chunhua and Hengel, Anton van den},
  booktitle={CVPR},
  year={2020}
}

@inproceedings{meng2024deepstack,
  title={Deepstack: Deeply stacking visual tokens is surprisingly simple and effective for lmms},
  author={Meng, Lingchen and Yang, Jianwei and Tian, Rui and Dai, Xiyang and Wu, Zuxuan and Gao, Jianfeng and Jiang, Yu-Gang},
  booktitle={Advances in Neural Information Processing Systems},
  volume={37},
  pages={23464--23487},
  year={2024}
}

@article{tschannen2025siglip,
  title={Siglip 2: Multilingual vision-language encoders with improved semantic understanding, localization, and dense features},
  author={Tschannen, Michael and Gritsenko, Alexey and Wang, Xiao and Naeem, Muhammad Ferjad and Alabdulmohsin, Ibrahim and Parthasarathy, Nikhil and Evans, Talfan and Beyer, Lucas and Xia, Ye and Mustafa, Basil and others},
  journal={arXiv preprint arXiv:2502.14786},
  year={2025}
}

@article{chen2025comp,
  title={Comp: Continual multimodal pre-training for vision foundation models},
  author={Chen, Yitong and Meng, Lingchen and Peng, Wujian and Wu, Zuxuan and Jiang, Yu-Gang},
  journal={arXiv preprint arXiv:2503.18931},
  year={2025}
}

@inproceedings{shao2019objects365,
  title={Objects365: A large-scale, high-quality dataset for object detection},
  author={Shao, Shuai and Li, Zeming and Zhang, Tianyuan and Peng, Chao and Yu, Gang and Zhang, Xiangyu and Li, Jing and Sun, Jian},
  booktitle={Proceedings of the IEEE/CVF international conference on computer vision},
  pages={8430--8439},
  year={2019}
}

@article{abotN0,
  title={{ABot-N0}: Technical Report on the VLA Foundation Model for Versatile Embodied Navigation},
  author={{AMAP CV Lab}},
  journal={arXiv preprint},
  year={2025},
}

@inproceedings{zhou2025same,
  title={Same: Learning generic language-guided visual navigation with state-adaptive mixture of experts},
  author={Zhou, Gengze and Hong, Yicong and Wang, Zun and Zhao, Chongyang and Bansal, Mohit and Wu, Qi},
  booktitle={Proceedings of the IEEE/CVF International Conference on Computer Vision},
  pages={7794--7807},
  year={2025}
}

@inproceedings{zhou2024navgpt,
  title={Navgpt: Explicit reasoning in vision-and-language navigation with large language models},
  author={Zhou, Gengze and Hong, Yicong and Wu, Qi},
  booktitle={Proceedings of the AAAI Conference on Artificial Intelligence},
  volume={38},
  number={7},
  pages={7641--7649},
  year={2024}
}

@article{zhang1984fast,
  title={A fast parallel algorithm for thinning digital patterns},
  author={Zhang, Tongjie Y and Suen, Ching Y.},
  journal={Communications of the ACM},
  volume={27},
  number={3},
  pages={236--239},
  year={1984},
  publisher={ACM New York, NY, USA}
}

@inproceedings{zhou2024navgpt2,
  title={Navgpt-2: Unleashing navigational reasoning capability for large vision-language models},
  author={Zhou, Gengze and Hong, Yicong and Wang, Zun and Wang, Xin Eric and Wu, Qi},
  booktitle={European Conference on Computer Vision},
  pages={260--278},
  year={2024},
  organization={Springer}
}

@inproceedings{long2024instructnav,
  author       = {Yuxing Long and
                  Wenzhe Cai and
                  Hongcheng Wang and
                  Guanqi Zhan and
                  Hao Dong},
  editor       = {Pulkit Agrawal and
                  Oliver Kroemer and
                  Wolfram Burgard},
  title        = {InstructNav: Zero-shot System for Generic Instruction Navigation in
                  Unexplored Environment},
  booktitle    = {Conference on Robot Learning, 6-9 November 2024, Munich, Germany},
  series       = {Proceedings of Machine Learning Research},
  pages        = {2049--2060},
  publisher    = {{PMLR}},
  year         = {2024},}

@inproceedings{das2018embodied,
  title={Embodied question answering},
  author={Das, Abhishek and Datta, Samyak and Gkioxari, Georgia and Lee, Stefan and Parikh, Devi and Batra, Dhruv},
  booktitle={Proceedings of the IEEE conference on computer vision and pattern recognition},
  pages={1--10},
  year={2018}
}

@inproceedings{krantz2020beyond,
  title={Beyond the Nav-Graph: Vision-and-Language Navigation in Continuous Environments},
  author={Krantz, Jacob and Wijmans, Erik and Majumdar, Arjun and Batra, Dhruv and Lee, Stefan},
  booktitle={European Conference on Computer Vision (ECCV)},
  year={2020},
}

@inproceedings{wijmans2020ddppo,
  title={{DD-PPO}: Learning Near-Perfect PointGoal Navigators from 2.5 Billion Frames},
  author={Wijmans, Erik and Kadian, Abhishek and Morcos, Ari and Lee, Stefan and Essa, Irfan and Parikh, Devi and Savva, Manolis and Batra, Dhruv},
  booktitle={International Conference on Learning Representations (ICLR)},
  year={2020},
}

@inproceedings{batra2020objectnav,
  title={Object{N}av Revisited: On Evaluation of Embodied Agents Navigating to Objects},
  author={Batra, Dhruv and Gokaslan, Aaron and Kembhavi, Aniruddha and Maksymets, Oleksandr and Mottaghi, Roozbeh and Savva, Manolis and Toshev, Alexander and Wijmans, Erik},
  booktitle={arXiv preprint arXiv:2006.13171},
  year={2020},
}

@inproceedings{anderson2020rxr,
  title={Room-Across-Room: Multilingual Vision-and-Language Navigation with Dense Spatiotemporal Grounding},
  author={Ku, Alexander and Anderson, Peter and Patel, Roma and Ie, Eugene and Baldridge, Jason},
  booktitle={Proceedings of the 2020 Conference on Empirical Methods in Natural Language Processing (EMNLP)},
  pages={4392--4412},
  year={2020},
}

@article{yokoyama2024hm3d,
  title={{HM3D-OVON}: A Dataset and Benchmark for Open-Vocabulary Object Goal Navigation},
  author={Yokoyama, Naoki and Ramrakhya, Ram and Das, Abhishek and Batra, Dhruv and Ha, Sehoon},
  journal={arXiv preprint arXiv:2409.14296},
  year={2024},
}

@article{wang2025trackvla,
  title={{TrackVLA}: Embodied Visual Tracking in the Wild},
  author={Wang, Shaoan and Zhang, Jiazhao and Li, Minghan and Liu, Jiahang and Li, Anqi and Wu, Kui and Zhong, Fangwei and Yu, Junzhi and Zhang, Zhizheng and Wang, He},
  journal={arXiv preprint arXiv:2505.23189},
  year={2025},
}

@inproceedings{cheng2024navila,
  title={{NaVILA}: Legged Robot Vision-Language-Action Model for Navigation},
  author={Cheng, An-Chieh and Ji, Yandong and Yang, Zhaojing and Zou, Xueyan and Kautz, Jan and Biyik, Erdem and Yin, Hongxu and Liu, Sifei and Wang, Xiaolong},
  booktitle={Robotics: Science and Systems (RSS)},
  year={2025},
}

@article{li2026gn0,
  title={GN0: Toward a Unified Paradigm for Generation, Evaluation, and Policy Learning in Visual-Language Navigation},
  author={Li, Xinhai and Zhang, Xiaotao and Huang, Yuehao and Dong, Jiankun and Wang, Tianhang and Zhou, Sunyao and Wu, Yunzi and Sun, Chengnuo and Ge, Yunfei and Weng, Qizhen and others},
  journal={arXiv preprint arXiv:2606.03682},
  year={2026}
}

@article{omninav,
  title={{OmniNav}: A Unified Framework for Prospective Exploration and Visual-Language Navigation},
  author={Hu, Weijing and Wang, Jun and Hu, Teng and Chen, Jiteng and Xue, Siwen and Yue, Yufeng and Xie, Haoran and Zhang, Weixun and Lu, Huchuan and Lu, Zongqing and He, Haibin and Wang, Bolei},
  journal={arXiv preprint arXiv:2510.06436},
  year={2025},
}

@article{astranavworld,
  title={{AstraNav-World}: World Model for Foresight Control and Consistency},
  author={Hu, Weijing and Wang, Jun and Hu, Teng and Chen, Jiteng and Xue, Siwen and Yue, Yufeng and Wu, Yanyun and He, Haibin and Wang, Bolei and Lu, Huchuan and Lu, Zongqing},
  journal={arXiv preprint arXiv:2603.23745},
  year={2026},
}

@article{zhang2024uni,
  title={{Uni-NaVid}: A Video-based Vision-Language-Action Model for Unifying Embodied Navigation Tasks},
  author={Zhang, Jiazhao and Wang, Kunyu and Wang, Shaoan and Li, Minghan and Liu, Haoran and Wei, Songlin and Wang, Zhongyuan and Zhang, Zhizheng and Wang, He},
  journal={Robotics: Science and Systems},
  year={2025},
}

@inproceedings{gupta2016novel,
  title={A Novel Policy for Multi-robot Navigation Using Image-Based Visual Servoing},
  author={Gupta, Monika and Tiwari, Ritu and Raja, G Laxmidhar},
  booktitle={Proceedings of the International Conference on Informatics and Analytics},
  year={2016},
}

@article{an2024etpnav,
  title={Etpnav: Evolving topological planning for vision-language navigation in continuous environments},
  author={An, Dong and Wang, Hanqing and Wang, Wenguan and Wang, Zun and Huang, Yan and He, Keji and Wang, Liang},
  journal={IEEE Transactions on Pattern Analysis and Machine Intelligence},
  year={2024},
  publisher={IEEE}
}

@inproceedings{wang2024lookahead,
  title={Lookahead exploration with neural radiance representation for continuous vision-language navigation},
  author={Wang, Zihan and Li, Xiangyang and Yang, Jiahao and Liu, Yeqi and Hu, Junjie and Jiang, Ming and Jiang, Shuqiang},
  booktitle={Proceedings of the IEEE/CVF Conference on Computer Vision and Pattern Recognition},
  pages={13753--13762},
  year={2024}
}

@article{zhang2024navid,
  title={Navid: Video-based vlm plans the next step for vision-and-language navigation},
  author={Zhang, Jiazhao and Wang, Kunyu and Xu, Rongtao and Zhou, Gengze and Hong, Yicong and Fang, Xiaomeng and Wu, Qi and Zhang, Zhizheng and Wang, He},
  journal={arXiv preprint arXiv:2402.15852},
  year={2024}
}

@article{wei2025streamvln,
  title={Streamvln: Streaming vision-and-language navigation via slowfast context modeling},
  author={Wei, Meng and Wan, Chenyang and Yu, Xiqian and Wang, Tai and Yang, Yuqiang and Mao, Xiaohan and Zhu, Chenming and Cai, Wenzhe and Wang, Hanqing and Chen, Yilun and others},
  journal={arXiv preprint arXiv:2507.05240},
  year={2025}
}

@article{wei2025ground,
  title={Ground slow, move fast: A dual-system foundation model for generalizable vision-and-language navigation},
  author={Wei, Meng and Wan, Chenyang and Peng, Jiaqi and Yu, Xiqian and Yang, Yuqiang and Feng, Delin and Cai, Wenzhe and Zhu, Chenming and Wang, Tai and Pang, Jiangmiao and others},
  journal={arXiv preprint arXiv:2512.08186},
  year={2025}
}

@inproceedings{yokoyama2024vlfm,
  title={Vlfm: Vision-language frontier maps for zero-shot semantic navigation},
  author={Yokoyama, Naoki and Ha, Sehoon and Batra, Dhruv and Wang, Jiuguang and Bucher, Bernadette},
  booktitle={2024 IEEE International Conference on Robotics and Automation (ICRA)},
  pages={42--48},
  year={2024},
  organization={IEEE}
}

@inproceedings{zhu2025move,
  title={Move to understand a 3d scene: Bridging visual grounding and exploration for efficient and versatile embodied navigation},
  author={Zhu, Ziyu and Wang, Xilin and Li, Yixuan and Zhang, Zhuofan and Ma, Xiaojian and Chen, Yixin and Jia, Baoxiong and Liang, Wei and Yu, Qian and Deng, Zhidong and others},
  booktitle={Proceedings of the IEEE/CVF International Conference on Computer Vision},
  pages={8120--8132},
  year={2025}
}

@article{zeng2024poliformer,
  title={Poliformer: Scaling on-policy rl with transformers results in masterful navigators},
  author={Zeng, Kuo-Hao and Zhang, Zichen and Ehsani, Kiana and Hendrix, Rose and Salvador, Jordi and Herrasti, Alvaro and Girshick, Ross and Kembhavi, Aniruddha and Weihs, Luca},
  journal={arXiv preprint arXiv:2406.20083},
  year={2024}
}

@inproceedings{zhong2024empowering,
  title={Empowering embodied visual tracking with visual foundation models and offline rl},
  author={Zhong, Fangwei and Wu, Kui and Ci, Hai and Wang, Churan and Chen, Hao},
  booktitle={European Conference on Computer Vision},
  pages={139--155},
  year={2024},
  organization={Springer}
}

@article{liu2025trackvlapp,
  title={Trackvla++: Unleashing reasoning and memory capabilities in vla models for embodied visual tracking},
  author={Liu, Jiahang and Qi, Yunpeng and Zhang, Jiazhao and Li, Minghan and Wang, Shaoan and Wu, Kui and Ye, Hanjing and Zhang, Hong and Chen, Zhibo and Zhong, Fangwei and others},
  journal={arXiv preprint arXiv:2510.07134},
  year={2025}
}

@inproceedings{openeqa,
  title={OpenEQA: Embodied Question Answering in the Era of Foundation Models},
  author={Majumdar, Arjun and Ajay, Anurag and Zhang, Xiaohan and Putta, Pranav and Yenamandra, Sriram and Henaff, Mikael and Silwal, Sneha and Mcvay, Paul and Maksymets, Oleksandr and Arnaud, Sergio and Yadav, Karmesh and Li, Qiyang and Newman, Ben and Sharma, Mohit and Berges, Vincent and Zhang, Shiqi and Agrawal, Pulkit and Bisk, Yonatan and Batra, Dhruv and Kalakrishnan, Mrinal and Meier, Franziska and Paxton, Chris and Sax, Alexander and Rajeswaran, Aravind},
  booktitle={Proceedings of the IEEE/CVF Conference on Computer Vision and Pattern Recognition (CVPR)},
  pages={16488--16498},
  year={2024},
}

@inproceedings{exploreeqa,
  author       = {Allen Z. Ren and
                  Jaden Clark and
                  Anushri Dixit and
                  Masha Itkina and
                  Anirudha Majumdar and
                  Dorsa Sadigh},
  editor       = {Dana Kulic and
                  Gentiane Venture and
                  Kostas E. Bekris and
                  Enrique Coronado},
  title        = {Explore until Confident: Efficient Exploration for Embodied Question
                  Answering},
  booktitle    = {Robotics: Science and Systems XX, Delft, The Netherlands, July 15-19, 2024},
  year         = {2024},
}

@article{grapheqa,
  author       = {Saumya Saxena and
                  Blake Buchanan and
                  Chris Paxton and
                  Bingqing Chen and
                  Narunas Vaskevicius and
                  Luigi Palmieri and
                  Jonathan Francis and
                  Oliver Kroemer},
  title        = {GraphEQA: Using 3D Semantic Scene Graphs for Real-time Embodied Question
                  Answering},
  journal      = {CoRR},
  volume       = {abs/2412.14480},
  year         = {2024},
}

@article{memoryeqa,
  author       = {Chengyang Li and
                  Shuai Wang and
                  Kejiang Ye and
                  Weijie Yuan and
                  Boyu Zhou and
                  Yik{-}Chung Wu and
                  Cheng{-}Zhong Xu and
                  Huseyin Arslan},
  title        = {Memory Centric Power Allocation for Multi-Agent Embodied Question
                  Answering},
  journal      = {CoRR},
  volume       = {abs/2604.17810},
  year         = {2026}
}

@inproceedings{fineeqa,
  author       = {Kaixuan Jiang and
                  Yang Liu and
                  Weixing Chen and
                  Jingzhou Luo and
                  Ziliang Chen and
                  Ling Pan and
                  Guanbin Li and
                  Liang Lin},
  title        = {Beyond the Destination: {A} Novel Benchmark for Exploration-Aware
                  Embodied Question Answering},
  booktitle    = {{IEEE/CVF} International Conference on Computer Vision, {ICCV} 2025,
                  Honolulu, HI, USA, October 19-25, 2025},
  pages        = {9091--9101},
  publisher    = {{IEEE}},
  year         = {2025},
}

@inproceedings{3dmem,
  author       = {Yuncong Yang and
                  Han Yang and
                  Jiachen Zhou and
                  Peihao Chen and
                  Hongxin Zhang and
                  Yilun Du and
                  Chuang Gan},
  title        = {3D-Mem: 3D Scene Memory for Embodied Exploration and Reasoning},
  booktitle    = {{IEEE/CVF} Conference on Computer Vision and Pattern Recognition,
                  {CVPR} 2025, Nashville, TN, USA, June 11-15, 2025},
  pages        = {17294--17303},
  publisher    = {Computer Vision Foundation / {IEEE}},
  year         = {2025}
}

@inproceedings{fasteqa,
  author       = {Haochen Zhang and
                  Nirav Savaliya and
                  Faizan Siddiqui and
                  Enna Sachdeva},
  title        = {{FAST-EQA:} Efficient Embodied Question Answering with Global and
                  Local Region Relevancy},
  booktitle    = {{IEEE/CVF} Winter Conference on Applications of Computer Vision, {WACV}
                  2026, Tucson, AZ, USA, March 6-10, 2026},
  pages        = {1664--1673},
  publisher    = {{IEEE}},
  year         = {2026}
}

@inproceedings{caesar2020nuscenes,
  title={nuscenes: A multimodal dataset for autonomous driving},
  author={Caesar, Holger and Bankiti, Varun and Lang, Alex H and Vora, Sourabh and Liong, Venice Erin and Xu, Qiang and Krishnan, Anush and Pan, Yu and Baldan, Giancarlo and Beijbom, Oscar},
  booktitle={Proceedings of the IEEE/CVF conference on computer vision and pattern recognition},
  pages={11621--11631},
  year={2020}
}

@inproceedings{sun2020scalability,
  title={Scalability in Perception for Autonomous Driving: Waymo Open Dataset},
  author={Sun, Pei and Kretzschmar, Henrik and Dotiwalla, Xerxes and Chouard, Aurelien and Patnaik, Vijaysai and Tsui, Paul and Guo, James and Zhou, Yin and Chai, Yuning and Caine, Benjamin and Vasudevan, Vijay and Han, Wei and Ngiam, Jiquan and Zhao, Hang and Timofeev, Aleksei and Ettinger, Scott and Krivokon, Maxim and Gao, Amy and Joshi, Aditya and Zhang, Yu and Shlens, Jonathon and Chen, Zhifeng and Anguelov, Dragomir},
  booktitle={Proceedings of the IEEE/CVF Conference on Computer Vision and Pattern Recognition (CVPR)},
  pages={2446--2454},
  year={2020}
}

@misc{openscene2023,
  title={OpenScene: The Largest Up-to-Date 3D Occupancy Prediction Benchmark in Autonomous Driving},
  author={OpenScene Contributors},
  howpublished={\url{https://github.com/OpenDriveLab/OpenScene}},
  year={2023}
}

@inproceedings{hu2023planning,
  title={Planning-oriented autonomous driving},
  author={Hu, Yihan and Yang, Jiazhi and Chen, Li and Li, Keyu and Sima, Chonghao and Zhu, Xizhou and Chai, Siqi and Du, Senyao and Lin, Tianwei and Wang, Wenhai and others},
  booktitle={Proceedings of the IEEE/CVF conference on computer vision and pattern recognition},
  pages={17853--17862},
  year={2023}
}

@InProceedings{Peng_2026_CVPR,
    author    = {Peng, Qihang and Chen, Xuesong and Yang, Chenye and Shi, Shaoshuai and Li, Hongsheng},
    title     = {ColaVLA: Leveraging Cognitive Latent Reasoning for Hierarchical Parallel Trajectory Planning in Autonomous Driving},
    booktitle = {Proceedings of the IEEE/CVF Conference on Computer Vision and Pattern Recognition (CVPR)},
    month     = {June},
    year      = {2026},
    pages     = {17809-17819}
}

@article{chitta2022transfuser,
  title={Transfuser: Imitation with transformer-based sensor fusion for autonomous driving},
  author={Chitta, Kashyap and Prakash, Aditya and Jaeger, Bernhard and Yu, Zehao and Renz, Katrin and Geiger, Andreas},
  journal={IEEE transactions on pattern analysis and machine intelligence},
  volume={45},
  number={11},
  pages={12878--12895},
  year={2022},
  publisher={IEEE}
}

@inproceedings{weng2024drive,
  title={Para-drive: Parallelized architecture for real-time autonomous driving},
  author={Weng, Xinshuo and Ivanovic, Boris and Wang, Yan and Wang, Yue and Pavone, Marco},
  booktitle={Proceedings of the IEEE/CVF Conference on Computer Vision and Pattern Recognition},
  pages={15449--15458},
  year={2024}
}

@article{yuan2024drama,
  title={Drama: An efficient end-to-end motion planner for autonomous driving with mamba},
  author={Yuan, Chengran and Zhang, Zhanqi and Sun, Jiawei and Sun, Shuo and Huang, Zefan and Lee, Christina Dao Wen and Li, Dongen and Han, Yuhang and Wong, Anthony and Tee, Keng Peng and others},
  journal={arXiv preprint arXiv:2408.03601},
  year={2024}
}

@article{li2025hydra,
  title={Hydra-mdp++: Advancing end-to-end driving via expert-guided hydra-distillation},
  author={Li, Kailin and Li, Zhenxin and Lan, Shiyi and Xie, Yuan and Zhang, Zhizhong and Liu, Jiayi and Wu, Zuxuan and Yu, Zhiding and Alvarez, Jose M},
  journal={arXiv preprint arXiv:2503.12820},
  year={2025}
}

@inproceedings{li2025hydranext,
  title={Hydra-next: Robust closed-loop driving with open-loop training},
  author={Li, Zhenxin and Wang, Shihao and Lan, Shiyi and Yu, Zhiding and Wu, Zuxuan and Alvarez, Jose M},
  booktitle={Proceedings of the IEEE/CVF International Conference on Computer Vision},
  pages={27305--27314},
  year={2025}
}

@inproceedings{xing2025goalflow,
  title={Goalflow: Goal-driven flow matching for multimodal trajectories generation in end-to-end autonomous driving},
  author={Xing, Zebin and Zhang, Xingyu and Hu, Yang and Jiang, Bo and He, Tong and Zhang, Qian and Long, Xiaoxiao and Yin, Wei},
  booktitle={Proceedings of the Computer Vision and Pattern Recognition Conference},
  pages={1602--1611},
  year={2025}
}

@inproceedings{li2025end,
  title={End-to-end driving with online trajectory evaluation via bev world model},
  author={Li, Yingyan and Wang, Yuqi and Liu, Yang and He, Jiawei and Fan, Lue and Zhang, Zhaoxiang},
  booktitle={Proceedings of the IEEE/CVF International Conference on Computer Vision},
  pages={27137--27146},
  year={2025}
}

@inproceedings{jiang2024vadv2,
  title={VADv2: End-to-end vectorized autonomous driving via probabilistic planning},
  author={Jiang, Bo and Chen, Shaoyu and Gao, Hao and Liao, Bencheng and Zhang, Qian and Liu, Wenyu and Wang, Xinggang},
  booktitle={The Fourteenth International Conference on Learning Representations},
  year={2024}
}

@article{zhou2025autovla,
  title={Autovla: A vision-language-action model for end-to-end autonomous driving with adaptive reasoning and reinforcement fine-tuning},
  author={Zhou, Zewei and Cai, Tianhui and Zhao, Seth Z and Zhang, Yun and Huang, Zhiyu and Zhou, Bolei and Ma, Jiaqi},
  journal={arXiv preprint arXiv:2506.13757},
  year={2025}
}

@article{li2025recogdrive,
  title={Recogdrive: A reinforced cognitive framework for end-to-end autonomous driving},
  author={Li, Yongkang and Xiong, Kaixin and Guo, Xiangyu and Li, Fang and Yan, Sixu and Xu, Gangwei and Zhou, Lijun and Chen, Long and Sun, Haiyang and Wang, Bing and others},
  journal={arXiv preprint arXiv:2506.08052},
  year={2025}
}

@inproceedings{chen2025drivinggpt,
  title={Drivinggpt: Unifying driving world modeling and planning with multi-modal autoregressive transformers},
  author={Chen, Yuntao and Wang, Yuqi and Zhang, Zhaoxiang},
  booktitle={Proceedings of the IEEE/CVF International Conference on Computer Vision},
  pages={26890--26900},
  year={2025}
}

@inproceedings{
li2026discrete,
title={Discrete Diffusion for Reflective Vision-Language-Action Models in Autonomous Driving},
author={Pengxiang Li and Yinan Zheng and Yue Wang and Huimin Wang and Hang Zhao and Jingjing Liu and Xianyuan Zhan and Kun Zhan and XianPeng Lang},
booktitle={The Fourteenth International Conference on Learning Representations},
year={2026},
url={https://openreview.net/forum?id=XJxXSMLDoZ}
}

@article{li2024enhancing,
  title={Enhancing end-to-end autonomous driving with latent world model},
  author={Li, Yingyan and Fan, Lue and He, Jiawei and Wang, Yuqi and Chen, Yuntao and Zhang, Zhaoxiang and Tan, Tieniu},
  journal={arXiv preprint arXiv:2406.08481},
  year={2024}
}

@article{zhang2025embodied,
  title={Embodied navigation foundation model},
  author={Zhang, Jiazhao and Li, Anqi and Qi, Yunpeng and Li, Minghan and Liu, Jiahang and Wang, Shaoan and Liu, Haoran and Zhou, Gengze and Wu, Yuze and Li, Xingxing and others},
  journal={arXiv preprint arXiv:2509.12129},
  year={2025}
}

@inproceedings{chang2017matterport3d,
  author       = {Angel X. Chang and
                  Angela Dai and
                  Thomas A. Funkhouser and
                  Maciej Halber and
                  Matthias Nie{\ss}ner and
                  Manolis Savva and
                  Shuran Song and
                  Andy Zeng and
                  Yinda Zhang},
  title        = {Matterport3D: Learning from {RGB-D} Data in Indoor Environments},
  booktitle    = {2017 International Conference on 3D Vision, 3DV 2017, Qingdao, China,
                  October 10-12, 2017},
  pages        = {667--676},
  publisher    = {{IEEE} Computer Society},
  year         = {2017},
  url          = {https://doi.org/10.1109/3DV.2017.00081},
  doi          = {10.1109/3DV.2017.00081},
}

@inproceedings{loshchilov2019decoupled,
  author       = {Ilya Loshchilov and
                  Frank Hutter},
  title        = {Decoupled Weight Decay Regularization},
  booktitle    = {7th International Conference on Learning Representations, {ICLR} 2019,
                  New Orleans, LA, USA, May 6-9, 2019},
  publisher    = {OpenReview.net},
  year         = {2019},
  url          = {https://openreview.net/forum?id=Bkg6RiCqY7},
}

@inproceedings{savva2019habitat,
  author       = {Manolis Savva and
                  Jitendra Malik and
                  Devi Parikh and
                  Dhruv Batra and
                  Abhishek Kadian and
                  Oleksandr Maksymets and
                  Yili Zhao and
                  Erik Wijmans and
                  Bhavana Jain and
                  Julian Straub and
                  Jia Liu and
                  Vladlen Koltun},
  title        = {Habitat: {A} Platform for Embodied {AI} Research},
  booktitle    = {2019 {IEEE/CVF} International Conference on Computer Vision, {ICCV}
                  2019, Seoul, Korea (South), October 27 - November 2, 2019},
  pages        = {9338--9346},
  publisher    = {{IEEE}},
  year         = {2019},
  url          = {https://doi.org/10.1109/ICCV.2019.00943},
  doi          = {10.1109/ICCV.2019.00943},
}

@inproceedings{ramakrishnan2021hm3d,
  author       = {Santhosh Kumar Ramakrishnan and
                  Aaron Gokaslan and
                  Erik Wijmans and
                  Oleksandr Maksymets and
                  Alexander Clegg and
                  John M. Turner and
                  Eric Undersander and
                  Wojciech Galuba and
                  Andrew Westbury and
                  Angel X. Chang and
                  Manolis Savva and
                  Yili Zhao and
                  Dhruv Batra},
  title        = {Habitat-Matterport 3D Dataset {(HM3D):} 1000 Large-scale 3D Environments
                  for Embodied {AI}},
  booktitle    = {Proceedings of the Neural Information Processing Systems Track on
                  Datasets and Benchmarks 1, NeurIPS Datasets and Benchmarks 2021, December
                  2021, virtual},
  year         = {2021},
  url          = {https://datasets-benchmarks-proceedings.neurips.cc/paper/2021/hash/34173cb38f07f89ddbebc2ac9128303f-Abstract-round2.html},
}

@inproceedings{kazemzadeh2014referitgame,
  author       = {Sahar Kazemzadeh and
                  Vicente Ordonez and
                  Mark Matten and
                  Tamara L. Berg},
  title        = {ReferItGame: Referring to Objects in Photographs of Natural Scenes},
  booktitle    = {Proceedings of the 2014 Conference on Empirical Methods in Natural
                  Language Processing, {EMNLP} 2014, October 25-29, 2014, Doha, Qatar,
                  {A} meeting of SIGDAT, a Special Interest Group of the {ACL}},
  pages        = {787--798},
  publisher    = {{ACL}},
  year         = {2014},
  url          = {https://doi.org/10.3115/v1/d14-1086},
  doi          = {10.3115/V1/D14-1086},
}

@article{wu2025qwenimage,
  author       = {Chenfei Wu and
                  Jiahao Li and
                  Jingren Zhou and
                  Junyang Lin and
                  Kaiyuan Gao and
                  Kun Yan and
                  Shengming Yin and
                  Shuai Bai and
                  Xiao Xu and
                  Yilei Chen and
                  Yuxiang Chen and
                  Zecheng Tang and
                  Zekai Zhang and
                  Zhengyi Wang and
                  An Yang and
                  Bowen Yu and
                  Chen Cheng and
                  Dayiheng Liu and
                  Deqing Li and
                  Hang Zhang and
                  Hao Meng and
                  Hu Wei and
                  Jingyuan Ni and
                  Kai Chen and
                  Kuan Cao and
                  Liang Peng and
                  Lin Qu and
                  Minggang Wu and
                  Peng Wang and
                  Shuting Yu and
                  Tingkun Wen and
                  Wensen Feng and
                  Xiaoxiao Xu and
                  Yi Wang and
                  Yichang Zhang and
                  Yongqiang Zhu and
                  Yujia Wu and
                  Yuxuan Cai and
                  Zenan Liu},
  title        = {Qwen-Image Technical Report},
  journal      = {CoRR},
  volume       = {abs/2508.02324},
  year         = {2025},
  url          = {https://doi.org/10.48550/arXiv.2508.02324},
  doi          = {10.48550/ARXIV.2508.02324},
  eprinttype   = {arXiv},
  eprint       = {2508.02324},
}

@article{vlnverse,
  author       = {Sihao Lin and
                  Zerui Li and
                  Xunyi Zhao and
                  Gengze Zhou and
                  Liuyi Wang and
                  Rong Wei and
                  Rui Tang and
                  Juncheng Li and
                  Hanqing Wang and
                  Jiangmiao Pang and
                  Anton van den Hengel and
                  Jiajun Liu and
                  Qi Wu},
  title        = {VLNVerse: {A} Benchmark for Vision-Language Navigation with Versatile,
                  Embodied, Realistic Simulation and Evaluation},
  journal      = {CoRR},
  volume       = {abs/2512.19021},
  year         = {2025},
  url          = {https://doi.org/10.48550/arXiv.2512.19021},
  doi          = {10.48550/ARXIV.2512.19021},
  eprinttype   = {arXiv},
  eprint       = {2512.19021},
}

@article{wang2025vlnpe,
  author       = {Liuyi Wang and
                  Xinyuan Xia and
                  Hui Zhao and
                  Hanqing Wang and
                  Tai Wang and
                  Yilun Chen and
                  Chengju Liu and
                  Qijun Chen and
                  Jiangmiao Pang},
  title        = {Rethinking the Embodied Gap in Vision-and-Language Navigation: {A}
                  Holistic Study of Physical and Visual Disparities},
  journal      = {CoRR},
  volume       = {abs/2507.13019},
  year         = {2025},
  url          = {https://doi.org/10.48550/arXiv.2507.13019},
  doi          = {10.48550/ARXIV.2507.13019},
  eprinttype   = {arXiv},
  eprint       = {2507.13019},
}

@techreport{internvlan1,
  title={InternVLA-N1: An Open Dual-System Vision-Language Navigation Foundation Model with Learned Latent Plans},
  author={Cai, Wenzhe and Feng, Delin and Liu, Yu and Pang, Jiangmiao and Peng, Jiaqi and Wan, Chenyang and Wang, Hanqing and Wang, Liuyi and Wang, Tai and Wei, Meng and Yang, Yuqiang and Yu, Xiqian and Zhu, Chenming},
  year={2025},
  institution={Shanghai AI Laboratory},
  url={https://internrobotics.github.io/internvla-n1.github.io/}
}

@article{yin2024sgnav,
      title={SG-Nav: Online 3D Scene Graph Prompting for LLM-based Zero-shot Object Navigation},
      author={Hang Yin and Xiuwei Xu and Zhenyu Wu and Jie Zhou and Jiwen Lu},
      journal={arXiv preprint arXiv:2410.08189},
      year={2024}
}

@article{kuang2024openfmnav,
  title={OpenFMNav: Towards Open-Set Zero-Shot Object Navigation via Vision-Language Foundation Models},
  author={Kuang, Yuxuan and Lin, Hai and Jiang, Meng},
  journal={arXiv preprint arXiv:2402.10670},
  year={2024}
}

@inproceedings{thomason2020cvdn,
  title={Vision-and-Dialog Navigation},
  author={Thomason, Jesse and Murray, Michael and Cakmak, Maya and Zettlemoyer, Luke},
  booktitle={Proceedings of the Conference on Robot Learning},
  pages={394--406},
  year={2020},
  volume={100},
  series={Proceedings of Machine Learning Research},
  publisher={PMLR}
}

@inproceedings{zhu2021soon,
  title={SOON: Scenario Oriented Object Navigation With Graph-Based Exploration},
  author={Zhu, Fengda and Liang, Xiwen and Zhu, Yi and Yu, Qizhi and Chang, Xiaojun and Liang, Xiaodan},
  booktitle={Proceedings of the IEEE/CVF Conference on Computer Vision and Pattern Recognition (CVPR)},
  pages={12689--12699},
  month={June},
  year={2021}
}

@inproceedings{wang2024srdf,
  title={Bootstrapping Language-Guided Navigation Learning with Self-Refining Data Flywheel},
  author={Wang, Zun and Li, Jialu and Hong, Yicong and Li, Songze and Li, Kunchang and Yu, Shoubin and Wang, Yi and Qiao, Yu and Wang, Yali and Bansal, Mohit and Wang, Limin},
  booktitle={International Conference on Learning Representations},
  volume={2025},
  pages={23542--23568},
  year={2025}
}

@article{zhao2025vlnmme,
  title={VLN-MME: Diagnosing MLLMs as Language-guided Visual Navigation Agents},
  author={Zhao, Xunyi and Zhou, Gengze and Wu, Qi},
  journal={arXiv preprint arXiv:2512.24851},
  year={2025}
}

@inproceedings{zhong2025robotrom,
  title={RoboTrom-Nav: A Unified Framework for Embodied Navigation Integrating Perception, Planning, and Prediction},
  author={Zhong, Yufeng and Feng, Chengjian and Yan, Feng and Liu, Fanfan and Zheng, Liming and Ma, Lin},
  booktitle={Proceedings of the IEEE/CVF International Conference on Computer Vision},
  pages={6416--6425},
  year={2025}
}

@article{cao2024cognav,
  title={CogNav: Cognitive Process Modeling for Object Goal Navigation with LLMs},
  author={Cao, Yihan and Zhang, Jiazhao and Yu, Zhinan and Liu, Shuzhen and Qin, Zheng and Zou, Qin and Du, Bo and Xu, Kai},
  journal={arXiv preprint arXiv:2412.10439},
  year={2024}
}

@article{nie2025wmnav,
  title={WMNav: Integrating Vision-Language Models into World Models for Object Goal Navigation},
  author={Nie, Dujun and Guo, Xianda and Duan, Yiqun and Zhang, Ruijun and Chen, Long},
  journal={arXiv preprint arXiv:2503.02247},
  year={2025}
}

@misc{zhang2024trihelperzeroshotobjectnavigation,
      title={TriHelper: Zero-Shot Object Navigation with Dynamic Assistance}, 
      author={Lingfeng Zhang and Qiang Zhang and Hao Wang and Erjia Xiao and Zixuan Jiang and Honglei Chen and Renjing Xu},
      year={2024},
      eprint={2403.15223},
      archivePrefix={arXiv},
      primaryClass={cs.RO},
      url={https://arxiv.org/abs/2403.15223}, 
}

@article{liang2026planning,
  title={Planning-aligned Token Compression for Long-Context Autonomous Driving},
  author={Liang, Zhixuan and Chen, Yuxiao and You, Yurong and Karkus, Peter and Ding, Wenhao and Li, Boyi and Popov, Alexander and Wang, Yan and Igl, Maximilian and Li, Yiming and others},
  journal={arXiv preprint arXiv:2606.07464},
  year={2026}
}

@article{xia2026habitat,
  title={Habitat-GS: A High-Fidelity Navigation Simulator with Dynamic Gaussian Splatting},
  author={Xia, Ziyuan and Xu, Jingyi and Cui, Chong and Yu, Yuanhong and Zhang, Jiazhao and Yan, Qingsong and Ni, Tao and Chen, Junbo and Zhou, Xiaowei and Bao, Hujun and others},
  journal={arXiv preprint arXiv:2604.12626},
  year={2026}
}

@inproceedings{
zheng2025densegrounding,
title={DenseGrounding: Improving Dense Language-Vision Semantics for Ego-centric 3D Visual Grounding},
author={Henry Zheng and Hao Shi and Qihang Peng and Yong Xien Chng and Rui Huang and Yepeng Weng and zhongchao shi and Gao Huang},
booktitle={The Thirteenth International Conference on Learning Representations},
year={2025},
url={https://openreview.net/forum?id=iGafR0hSln}
}

@inproceedings{zheng2024denseg,
  title={Denseg: Alleviating vision-language feature sparsity in multi-view 3d visual grounding},
  author={Zheng, Henry and Shi, Hao and Chng, Yong Xien and Huang, Rui and Ni, Zanlin and Tan, Tianyi and Peng, Qihang and Weng, Yepeng and Shi, Zhongchao and Huang, Gao},
  booktitle={Autonomous Grand Challenge CVPR 2024 Workshop},
  volume={2},
  pages={6},
  year={2024}
}

@inproceedings{peng2025proxytransformation,
  title={ProxyTransformation: Preshaping Point Cloud Manifold With Proxy Attention For 3D Visual Grounding},
  author={Peng, Qihang and Zheng, Henry and Huang, Gao},
  booktitle={Proceedings of the Computer Vision and Pattern Recognition Conference},
  pages={24582--24592},
  year={2025}
}

@article{yin2025unigoal, 
      title={UniGoal: Towards Universal Zero-shot Goal-oriented Navigation}, 
      author={Hang Yin and Xiuwei Xu and Linqing Zhao and Ziwei Wang and Jie Zhou and Jiwen Lu},
      journal={arXiv preprint arXiv:2503.10630},
      year={2025}
}

@article{yin2025gcvln,
      title={GC-VLN: Instruction as Graph Constraints for Training-free Vision-and-Language Navigation},
      author={Hang Yin and Haoyu Wei and Xiuwei Xu and Wenxuan Guo and Jie Zhou and Jiwen Lu},
      journal={arXiv preprint arXiv:2509.10454},
      year={2025}
}

@article{yin2026alldaynav,
  title={AllDayNav: Lifelong Navigation via Real-World Reinforcement Learning},
  author={Yin, Hang and Zhang, Jiazhao and Liang, Yinan and Liu, Jiahang and Li, Minghan and Wang, He},
  journal={arXiv preprint arXiv:2606.10927},
  year={2026}
}
\bibliographystyle{colm2024_conference}

\appendix
\clearpage

\end{document}